%% file: main.tex
\documentclass[lettersize,journal]{IEEEtran}
\usepackage{amsmath,amsfonts}
\usepackage{array}
\usepackage{textcomp}
\usepackage{stfloats}
\usepackage{url}
\usepackage{verbatim}
\usepackage{cite}

\usepackage{graphicx}
\usepackage{algorithm}
\usepackage{algorithmic}
\usepackage{subcaption}
\usepackage{makecell}
\usepackage{tabularx}
\usepackage{subfig}

\newtheorem{theorem}{Theorem}
\newtheorem{definition}{Definition}

\begin{document}

\title{Foundations of a Developmental Design Paradigm for Integrated Continual Learning, Deliberative Behavior, and Comprehensibility}


\author{Zeki Doruk Erden and Boi Faltings,~\IEEEmembership{Member,~IEEE,}
\thanks{Z. D. Erden and B. Faltings are with École Polytechnique Fédérale de Lausanne, 1015 Switzerland, and Z. D. Erden is also with Sabancı University, 34956 İstanbul. \textit{(e-mail: doruk.erden@sabanciuniv.edu; boi.faltings@epfl.ch. Corresponding author: Zeki Doruk Erden)}}
\thanks{© 2025 IEEE.  Personal use of this material is permitted.  Permission from IEEE must be obtained for all other uses, in any current or future media, including reprinting/republishing this material for advertising or promotional purposes, creating new collective works, for resale or redistribution to servers or lists, or reuse of any copyrighted component of this work in other works.}}

\markboth{IEEE Transactions on Emerging Topics in Computational Intelligence, Vol. X, No. X, XXX 20XX}%
{Shell \MakeLowercase{\textit{et al.}}: A Sample Article Using IEEEtran.cls for IEEE Journals}


\maketitle

\begin{abstract}

Inherent limitations of contemporary machine learning systems in crucial areas—importantly in continual learning, information reuse, comprehensibility, and integration with deliberate behavior—are receiving increasing attention. To address these challenges, we introduce a system design, fueled by a novel learning approach conceptually grounded in principles of evolutionary developmental biology, that overcomes key limitations of current methods. Our design comprises three core components: The Modeller, a gradient-free learning mechanism inherently capable of continual learning and structural adaptation; a planner for goal-directed action over learned models; and a behavior encapsulation mechanism that can decompose complex behaviors into a hierarchical structure. We demonstrate proof-of-principle operation in a simple test environment. Additionally, we extend our modeling framework to higher-dimensional network-structured spaces, using MNIST for a shape detection task. Our framework shows promise in overcoming multiple major limitations of contemporary machine learning systems simultaneously and in an organic manner.

\end{abstract}


\section{Introduction}
\label{sec:intro}

The current machine learning (ML) paradigm uses continuous representations to approximate environmental structures through fixed internal architectures like neural networks (NNs). This approach has effectively addressed numerous challenges once considered among the toughest in AI. However, as these problems are solved, important limitations related to the methods of solving them and their practical integration into larger systems start to receive more attention \cite{clune2019ai,zador2019critique,marcus2018deep,lecun2022path}—most notably, the incapability of continual learning and information reuse, incomprehensibility and non-designability of internal structure, and difficulty of integrating learned information with deliberative behavior. These limitations force contemporary methods into frequent, resource-intensive retrainings, render scalability impractical, and obstruct human intervention or validation in the process.

An alternative exists: We can retain the power of advanced learning systems while prioritizing structure, comprehensibility, control, and preservation of knowledge as is essential for building adaptable, trustworthy systems suited for real-world complexities. These issues originate from the shared limitation of approximating environmental structures with fixed models, rather than learning them topologically. They can be addressed collectively and without limitations of individual subfields tackling them separately, through a different design philosophy that tackles the problem from the ground up.

To that end, we present the initial design of a system consisting of three components \cite{erden2025foundations}: (1) \textit{Modeller}, an alternative learning mechanism that captures the structure of the environment topologically in a discrete network without using gradients, able to learn continually without destroying existing knowledge, (2) A \textit{planner} that executes goal-directed actions based on a model generated by Modeller, (3) A \textit{behavior encapsulation mechanism}, currently demonstrated independently of agent operation, that decomposes behavior patterns produced by the planner into arbitrary hierarchical structures with autonomously detected subgoals.

We provide a detailed analysis of these components, explaining how they address several key limitations of contemporary ML, and demonstrate their proof-of-principle operation in a simple test environment. Additionally, we extend our modeling framework to higher-dimensional observation spaces, represented as networks, and demonstrate its application in a shape detection task involving 2D objects from the MNIST dataset. Our design, grounded in strong conceptual foundations from evolutionary theory, along with the supporting experimental results, highlights the potential of this approach and the opportunities it creates for simultaneously addressing multiple core limitations of AI systems in an organic manner.

\section{Background}

\subsection{Limitations of Machine Learning}
\label{sec:background}

The learning method introduced in this work—along with the general framework built through extensions upon it—is intended to lay the foundations for AI systems that \textit{inherently} avoid some of the major limitations of contemporary ML. Below, we provide a brief overview of these limitations and examine how existing design approaches fall short.

\subsubsection{Destructive adaptation in continual learning}
\label{sec:ml_limitations_continual_learning}

Modern ML methods are incapable of effectively integrating new knowledge into existing knowledge without destroying old information, which is a capability crucial for learning continually \cite{vandeven2024continuallearningcatastrophicforgetting, hadsell2020embracing}. Several \textit{ad-hoc} fixes exist for this problem of \textit{destructive adaptation}.\footnote{We prefer this term over "catastrophic forgetting," which misleadingly implies an unintended loss, whereas the process in NNs involves an active overwriting of past information.} The most straightforward involves storing and replaying past examples \cite{rolnick2019experience, buzzega2021rethinking, buzzega2020dark}. While simple, this approach merely circumvents the issue: As a system continues to learn, it must retain more past data, leading to explosive memory, computational demands and dilution of previous knowledge—impractical for real-world agents which can be expected to face millions of distinct situations over their lifetime. Other approaches try to mitigate the issue at a higher level but depend on restrictive assumptions, like clear task boundaries \cite{masse2018alleviating, jacobson2022task, kirkpatrick2017overcoming, iyer2022avoiding}, external signals indicating the active task \cite{rusu2016progressive}, access to bulk of expert-specific data at once \cite{erden2024directed, lee2020neural, iyer2022avoiding, hafez2023continual} or applying only to tasks with high orthogonality among them \cite{iyer2022avoiding, damleactive}. These assumptions clash with the structure of real-world learning, where agents encounter information sequentially, integrating new knowledge with past experiences without the need to store all prior data or rely on neatly defined 'tasks.' These issues also affect visual problems \cite{qu2021recent, galashov2023continually, wang2022continual}. The constraints of such solutions, along with a lack of guarantees beyond intuition or statistical bounds, position them more as mitigations than true solutions to the problem. One of the main contributions of this paper is to provide a mechanism of learning continuallly (1) without the aforementioned limiting assumptions, and (2) with a formal guarantee of preservation of information.

\subsubsection{Lack of internal structural organization and consequences} Learned structure in NNs is inherently intertwined, nondecomposable and non-organized, owing to their reliance on overparameterization \cite{du2018power}, weights fine-tuned to a few significant digits, dense connectivity, and nonlinear responses. Although there is a trend in the literature that attempts to incorporate high-level structural organization into learned models, most notably modularity and hierarchy; little practical advancement has been made in this regard. Many such methods impose rigid, predefined substructures rather than generating structured organization themselves \cite{su2024focuslearn, goyal2020object, pateria2021hierarchical}. Meanwhile, approaches that attempt to circumvent this constraint often rely on unstable foundations that restrict their expressive capacity due to inherent limitations in contemporary ML systems, such as dynamic task-specific substructures in continual learning being fundamentally constrained by the task-dependency issues of NNs (pt. (1) above) \cite{andreas2017modular, devin2017learning, sahni2017learning, goyal2019reinforcement, yang2020multi, pateria2021hierarchical, iyer2022avoiding}. Moreover, no existing approach achieves structural organization across all levels of organization in a fractal-like manner, limiting benefits to a layer atop the underlying neural networks while accepting a monolithic representation below this level. This absence of an \textit{internal structural organization}—which our approach aims to overcome by adaptively generating a topological model of the environment—is closely linked to, and arguably lies at the core of, several significant downstream limitations: \begin{itemize}
    \item \textit{Incomprehensibility and non-engineerability:} Overparameterization, fine-tuning, nonlinearity, and dense connectivity render a neural network’s internal structure incomprehensible and not "engineerable"—understood as designer modifiability or the ability to decouple internal components for use in other processes. These limitations hinder their widespread adoption and poses significant risks in sensitive domains that require high degrees of understanding and control \cite{marcus2018deep}. Research on explaining internal operations mainly focuses on post-hoc analyses of responses and parameter statistics rather than addressing the models' intrinsic incomprehensibility \cite{xu2019explainable, kashefi2023explainability}. This sharply contrasts with nearly all human-engineered systems, hampering our understanding of how ML models function, inhibiting decomposability \& modifiability, and making AI inherently uncontrollable \& unreliable beyond statistical guarantees. Below, we demonstrate that the design presented in this paper, through its structural learning operations, yields a model with an internal structure whose representations are decomposable as needed (see Section \ref{sec:planner} for their use in planning) and reasonably comprehensible upon inspection (see discussions of sample models throughout the paper).
    \item \textit{Deliberative behavior:} Planning is a well-established area of AI research \cite{ghallab2016automated}, offering advantages over approaches like reinforcement learning (RL) \cite{ccalicsir2019model}, such as precision and no need for relearning new goals in a known environment. Due to the nondecomposable and internally unstructured representation in NNs, however, the combination of these algorithms for deliberative behavior and flexible learning systems poses major problems. Traditional planning methods generally do not include environment model learning, and even those that do so to some degree \cite{jimenez2012review} operate under very restrictive assumptions, such as predefined variables or modelling only simple, linear relations \cite{mordoch2023learning,verma2021asking, stern2017efficient}. While model-based RL (MBRL) \cite{moerland2023model,moerland2020framework} partially merges deliberative behavior with learning, it suffers from limitations due to its non-structured representation of environments, making it challenging to represent alternative pathways to goals and conduct goal-oriented backward searches, often relying on random forward-sampling \cite{hammersley2013monte, mcmahon2022survey, otte2015survey} instead of a precise goal-directed search. Our modelling approach is not limited by the assumptions underlying environment-model learning approaches in planning literature, and our planner enables precise goal-directed behavior without the need for next-state sampling unlike MBRL methods.
    \item \textit{Behavior encapsulation:} A reflection for the desire for structured representations is the goal of automatically breaking down behavior into subunits, motivating fields like Hierarchical RL \cite{pateria2021hierarchical}. However, these approaches share the same limitations as the other methods that aim to incorporate a structural organization into learned models discussed above. Among our contributions is an initial demonstration of a behavior encapsulation mechanism (currently independent of the agent's operation) that can generate arbitrary hierarchical decompositions of behaviors designed by the planner. This mechanism can identify relevant subpolicies, along with their internal preconditions and subgoals, without any prior definitions, thus achieving this goal in a different context.
\end{itemize}

In addition to the shortcomings we described above, existing literature typically addresses these issues in isolation. However, these two limitations stem from a deeper, shared matter: the approximation of environmental structures using fixed models rather than learning them structurally, as detailed in the rest of this paper.\footnote{Several previous works have pursued structurally flexible learning throughout network development \cite{dai2019nest, evci2022gradmax, mitchell2023self, maile2022and, maile2022structural}, but none have directly targeted the specific challenges addressed in this work, and as a result, they lack the corresponding properties of our design (see next section). As a partial exception, some approaches have highlighted the potential of developmental systems for continual learning by triggering growth in response to information saturation—albeit within constrained, single-layer architectures and narrowly scoped learning processes limited to input-similarity-based clustering \cite{ding2023structural}.} By overcoming this fundamental challenge, these limitations can be addressed collectively, as is the central objective of our work. \footnote{Our approach to modelling is also possibly applicable to Bayesian structure learning \cite{kitson2023survey}; although this is not our primary motivation.}

\subsection{Evolutionary developmental biology}
\label{sec:evolution_and_ai}


The conceptual foundations of this work are grounded in an analogy between the Modern Synthesis understanding of evolutionary theory and the contemporary statistical machine learning paradigm, as well as in recent conceptual advances in evolutionary theory (particularly in \textit{evolutionary developmental biology}) which address explanatory gaps in the Modern Synthesis that mirror the aforementioned capability limitations of ML systems \cite{erden2025parallels, erden2025evolutionary}. For readers interested in this theoretical basis, we elaborate it in Appendix \ref{sec:app_evol_ai}, omitted from the main text for brevity. To briefly summarize that discussion, our design is grounded in the following principles of adaptation observed in biological systems:

\textit{Adaptation via local variation and selection}: A particular category of developmental processes leverage the \textit{variation and selection} principles underlying biological evolution \textit{locally}, where variants of a substructure are generated and then refined through a selective signal, forming the primary means for an organism to adapt to needs of the environment where no evolutionary preprogramming is possible \cite{marc2005plausibility, west2003developmental, gerhart2007theory}. At the heart of the learning mechanism introduced below lies a process of local variation and selection, which not only enables structural adaptation in the model but also underpins its continual learning capability (Theorem 1).


\textit{Regulation of weakly-linked core processes:} Discoveries in developmental mechanisms have highlighted the dominant role of regulatory changes to existing processes (such as their reuse in new contexts or recombination into novel functions) in the evolution of complex organisms, as opposed to the creation or modification of fundamental processes themselves \cite{marc2005plausibility, gerhart2007theory, holstein2012evolution, lemons2006genomic}. This mechanism also underlies universal structural features of biological systems like modularity, hierarchy, and repetition. A key enabler of such flexibility is the widely observed phenomenon of \textit{weak linkage} \cite{marc2005plausibility}, which refers to the interconnectivity of biological processes through simple, low-dimensional signals that carry minimal information—making them easily reusable or recombinable across different contexts. Our design enforces weak linkage through a discrete internal representation (see Def. 1), which supports straightforward modifiability during learning and facilitates usage by other processes such as planning. Furthermore, it enables the regulation of learned connections through the properties of fundamental computational units (see Def. 3), allowing for directed and on-demand model complexification.

\section{The Modeller: Learning Basic Environment Dynamics}
\label{sec:Modeller}

\subsection{Description of The Modeller}

The Modeller \cite{Erden2024Modelleyen} is designed to model sequential observations from an environment but can be applied to any prediction task. It learns the environment's structure with minimal exposure and can continually learn while maintaining consistency with past experiences. At the core of our method is a \textit{local variation-selection process} and \textit{weak linkage} between input sources and computational units, which, as will become evident, are critical for its continual learning guarantees and structured environment modeling. Below, we outline The Modeller's core mechanism, which relies on the immediate succession of activities in discrete state variables to model simple environmental dynamics. Section \ref{sec:mnr} extends this mechanism to observation spaces that can be represented as networks. Due to space constraints, we only provide key definitions, the basic learning process, and core continual learning properties. For a full description, see Appendix \ref{sec:Modeller_details} and Alg. \ref{alg:algorithm_adaptationloop} and \ref{alg:algorithm_csvstate}.

The Modeller is built upon, and seeks to learn, a discrete topological representation to enable weak linkage between components. Its fundamental representational unit is a construct termed the \textit{state variable (SV)}.

\begin{definition}
    \textit{A \textbf{state variable (SV)} $X$ is a unit whose state, $S_X$, can take values 1 (\textit{active}), -1 (\textit{inactive}), or 0 (\textit{unobserved/undefined}).} 
\end{definition}

SVs can be interpreted as Boolean variables with additional possibility to take an \textit{unobserved} value. This \textit{unobserved} state in addition to the usual Boolean active and inactive states, is the key to both the representational and the learning capability of this design, as will be explained below. The integers assigned for states are only for notation and not for algebraic operation. The following are subtypes of SVs (illustrated in Fig. \ref{fig:svs} in the Appendix):

\begin{definition}
    \textit{A \textbf{Base SV (BSV)} $B$ is an SV whose values are provided externally each timestep and whose state is limited by $S_B\in\{-1,1\}$. Each BSV comes with two \textbf{Dynamics SVs (DSVs)}, $B_A$ and $B_D$, that represent its activation and deactivation at current step ($t$) compared to previous timestep respectively; where $S_{B_A}$ and $S_{B_D}$ are $1$ if and only if $(S_B(t-1),\ S_B(t))=(-1,1)$ and $(1,-1)$ respectively, and persisting as long as no new event in BSVs are observed.}
\end{definition}

Intuitively, BSVs are environment observations, while DSVs represent their changes.\footnote{In our implementation we also use BSVs to represent actions taken by the agent in the previous step. Differently from environment observations, the actions do not have associated DSVs, since their activation and deactivation is in agent's control.}

\begin{definition}
    \textit{A \textbf{Conditioning SV (CSV)} $C$ is a type of SV with mutable sets of positive sources $X_P$, negative sources $X_N$, and conditioning targets $Y$. Positive and negative sources are BSVs and DSVs, while targets can be DSVs or other CSVs. The sources of $C$ are considered "satisfied" if all positive sources are active and all negative sources are not active. $S_C=1$ if sources are satisfied and $S_Y \in \{0,1\},\ \forall x \in Y$ (targets are active); $S_C=-1$ if sources are satisfied and $S_Y \in \{0,-1\},\ \forall x \in Y$ (targets are inactive), and $S_C=0$ otherwise. Additionally, each CSV has a "unconditionality" flag, which indicates if the CSV has, in the past, been always observed active when sources were satisfied ("unconditional"), was never observed active without a predictive explanation ("conditional"), or was sometimes observed active without a predictive explanation ("possibly conditional"), the latter representing uncertainty in a qualitative manner.}
\end{definition}

CSVs, the core computational units in this method and the \textit{loci of learning}, model the presence or absence of \textit{a relationship} between a learned condition (sources) and its effect (active targets), indicated by the CSV being active (1) or inactive (-1). A CSV being active means that the particular modelled relationship is seen (e.g. a change, as represented by a DSV, is observed after the observation of some sources), and it being inactive means that this relationship is not seen (e.g. the sources are not active, or DSV already in the target state). The CSV being undefined or unobserved corresponds to the case in which the conditions for the observation of the relationship are not satisfied in the first place (e.g. sources are not observed, or a SV is already active in case of an activation-DSV target). The distinction between a true "inactive" state and an "unobserved/undefined" state is important, as detailed below, to properly identify the presence of a "nonactivity" as a trigger for a learning process (\textit{inactive}), as opposed to a "nonactivity" that creates no conflict with the existing information and hence requires no change in the model (\textit{unobserved}). Notice that as such, CSVs are \textit{not} feedforward computational units; they \textit{represent} the presence or absence of the \textit{relationship between sources and targets} – states of their targets are set independently of the CSV, unlike feedforward units that determine target states based on source states. CSVs partially function as feedforward units only when they are used for the prediction of alternative outcomes (e.g. Sec. \ref{sec:planner}).



\begin{figure*}
     \centering
     \begin{subfigure}[t]{0.19\textwidth}
         \centering
         \includegraphics[width=\textwidth]{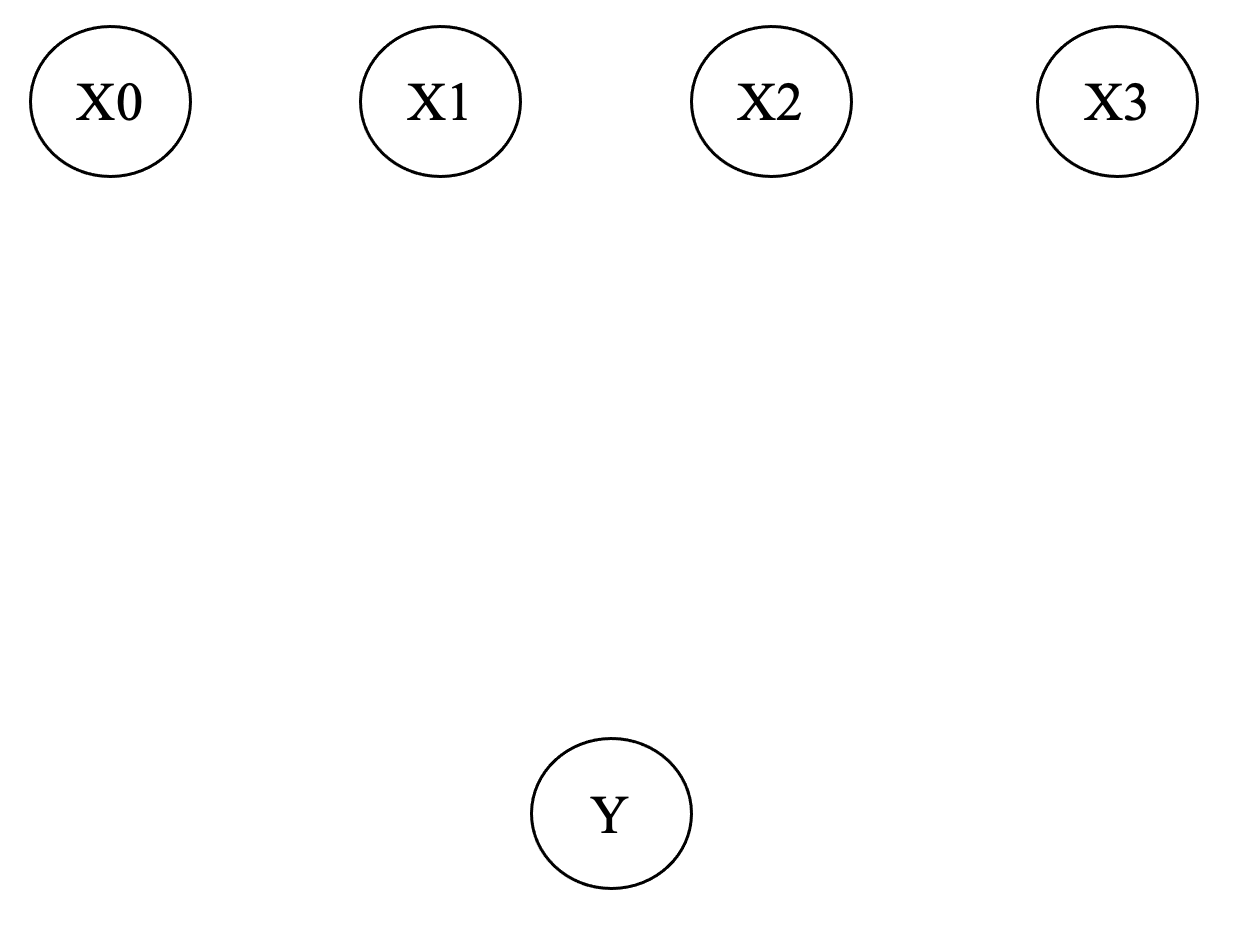}
         \caption{}
         \label{fig:csvform_1}
     \end{subfigure}
     \hfill
     \begin{subfigure}[t]{0.19\textwidth}
         \centering
         \includegraphics[width=\textwidth]{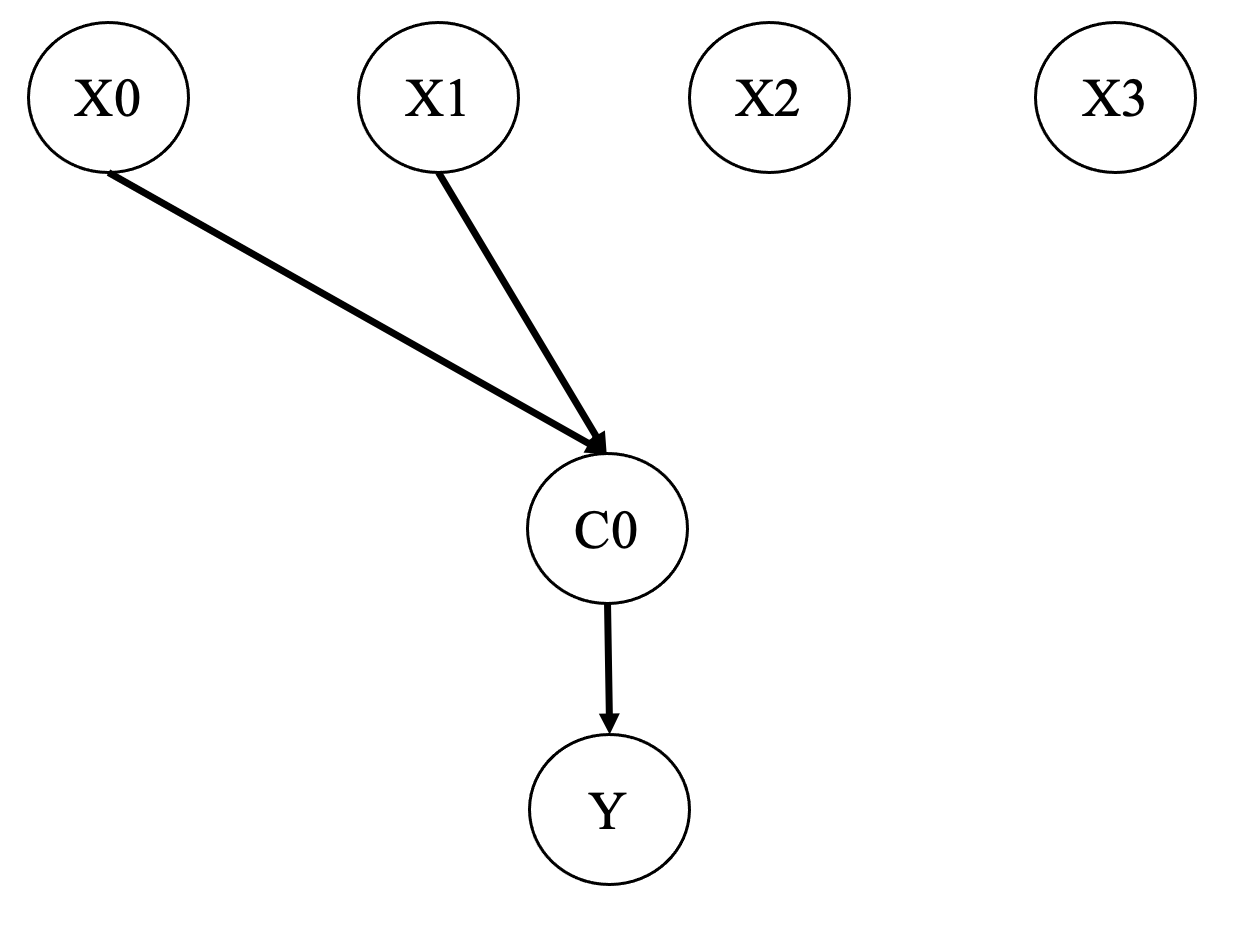}
         \caption{}
         \label{fig:csvform_2}
     \end{subfigure}
     \hfill
     \begin{subfigure}[t]{0.19\textwidth}
         \centering
         \includegraphics[width=\textwidth]{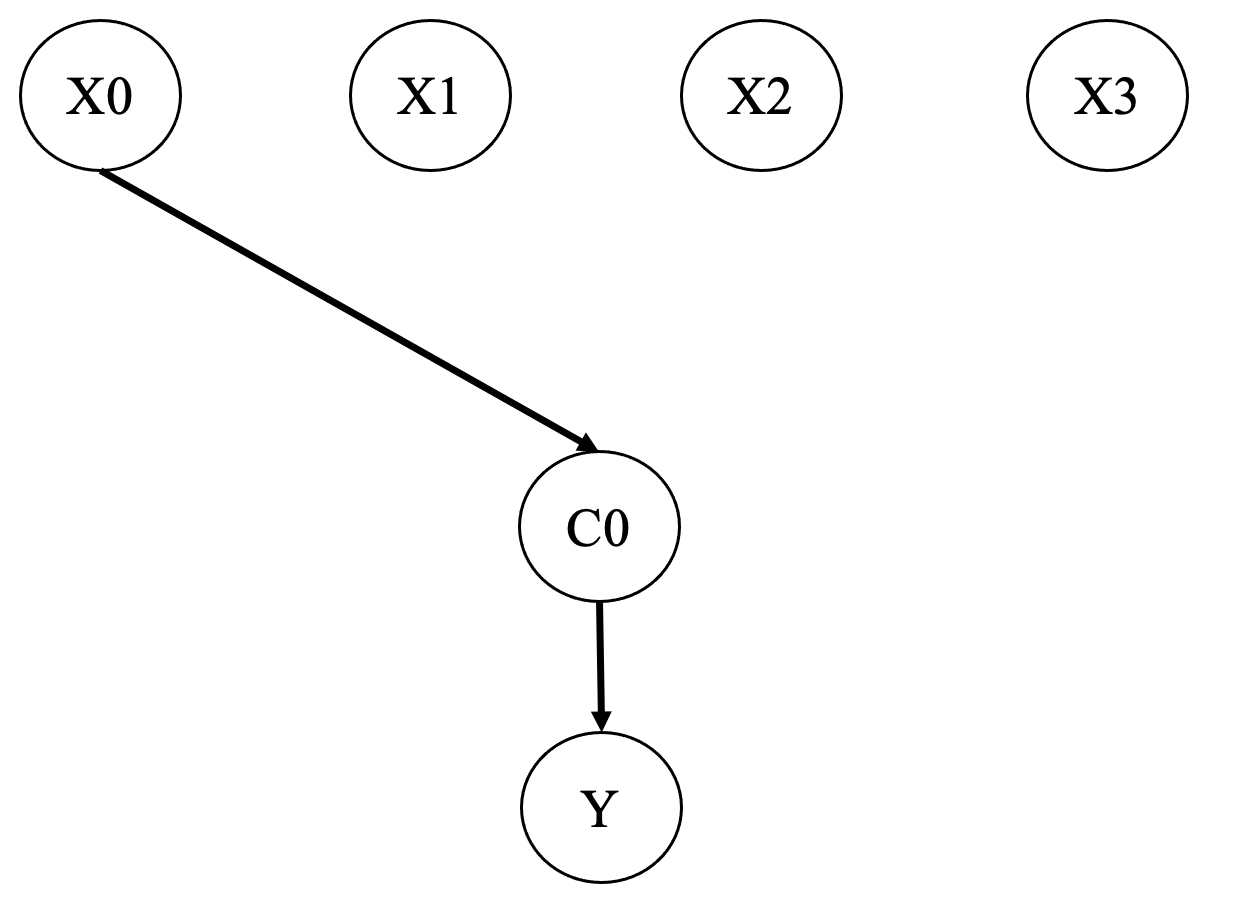}
         \caption{}
         \label{fig:csvform_3}
     \end{subfigure}
     \hfill
     \begin{subfigure}[t]{0.19\textwidth}
         \centering
         \includegraphics[width=\textwidth]{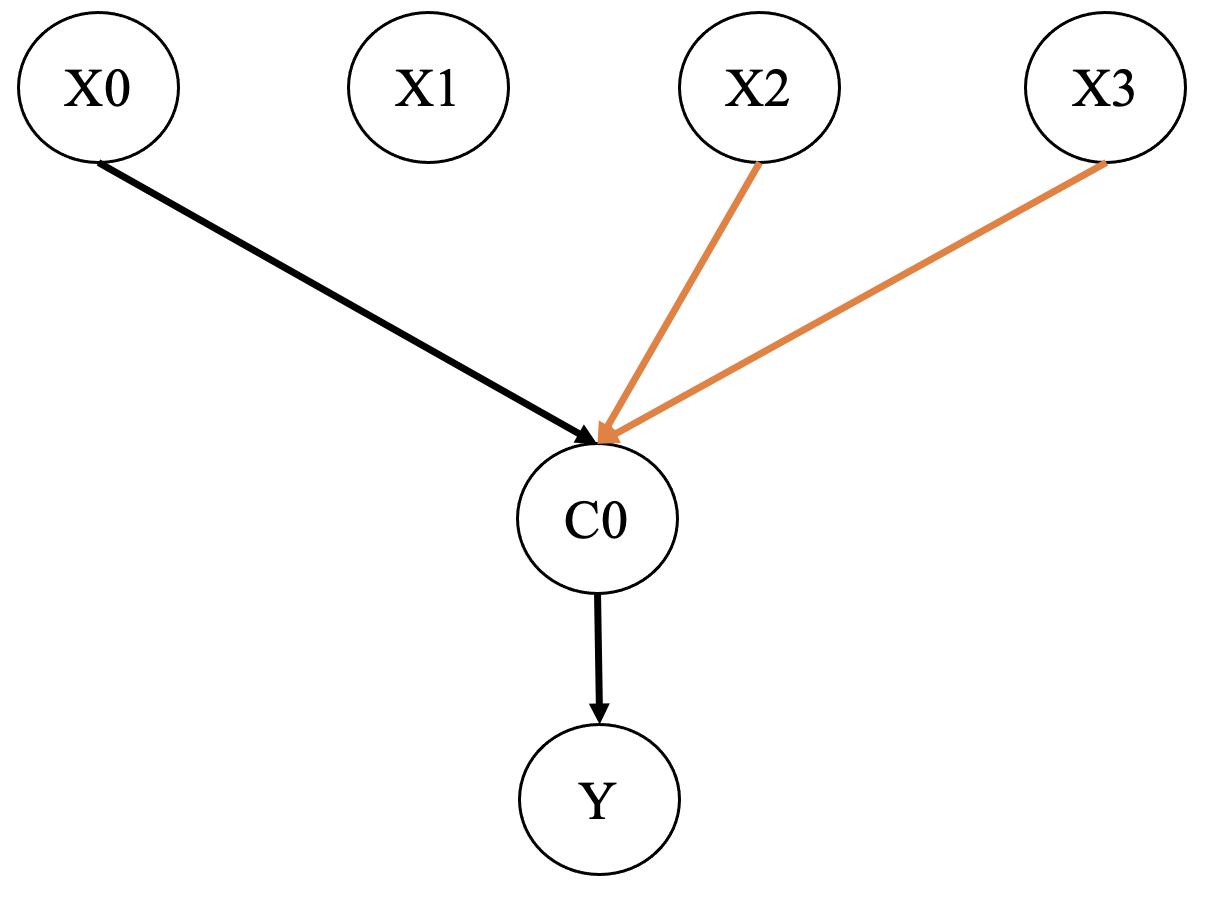}
         \caption{}
         \label{fig:csvform_4}
     \end{subfigure}
     \hfill
     \begin{subfigure}[t]{0.19\textwidth}
         \centering
         \includegraphics[width=\textwidth]{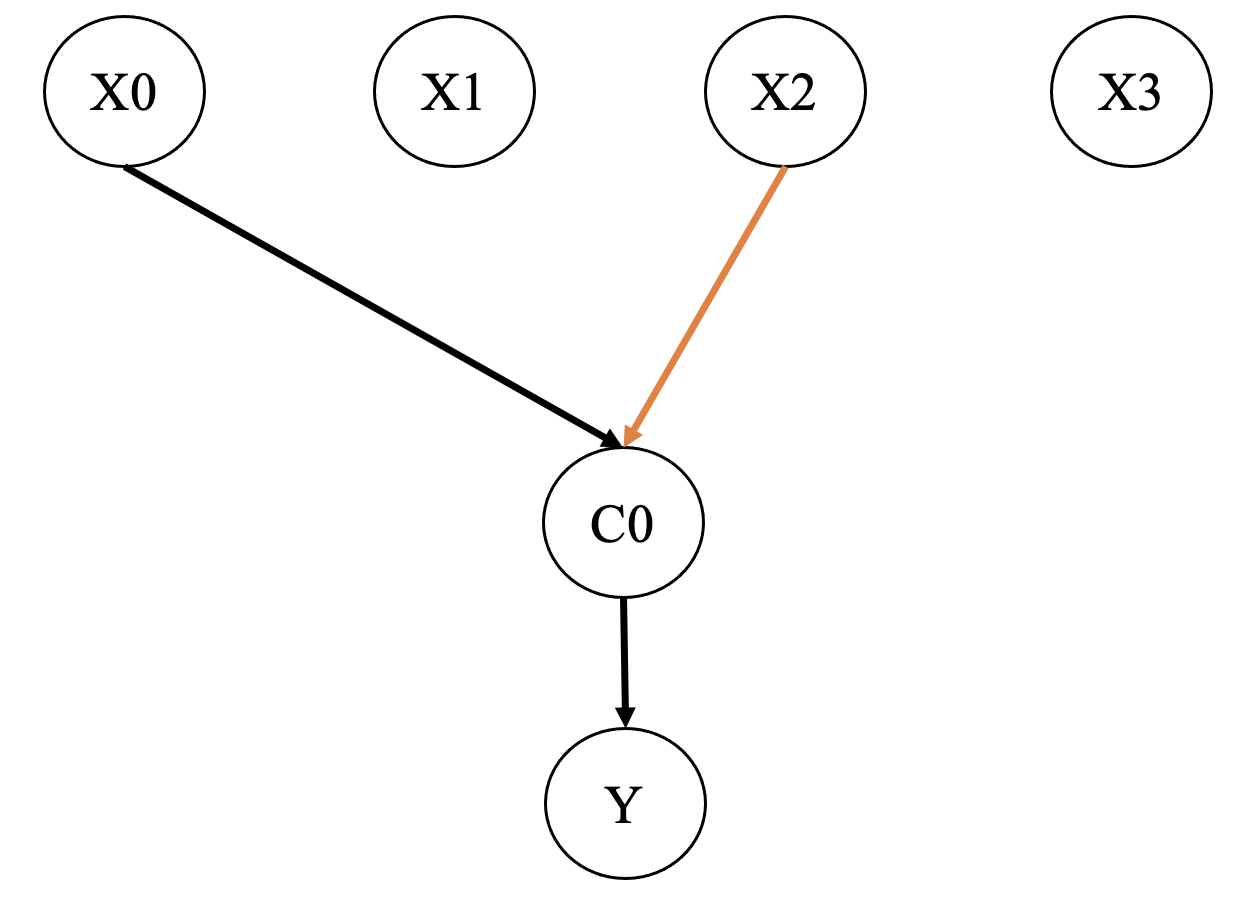}
         \caption{}
         \label{fig:csvform_5}
     \end{subfigure}
    \caption{Sample formation of a CSV in a continual manner. The relationship to be modelled is $Y = X0\ and\ !X2$ ("!" denotes "not"). Black and orange arrows represent positive and negative sources for CSV $C0$ respctively. $Xi$ can be interpreted either as single or grouped SVs. (a) Initial state with no relation formed between $X0-3$ and $Y$. (b) $X0, X1 \rightarrow Y$ observed. Positive connections hypothesizing both $X0$ \& $X1$ are required for Y are formed. (c) $X0 \rightarrow Y$ is observed. $X1$ is deduced unnecessary for $Y$. (d) $X0, X2, X3 \rightarrow !Y$ observed. $Y$ is hypothesized to be suppressed by $X2$ and $X3$. (e) $X0, X2 \rightarrow !Y$ observed. $X3$, seen unnecessary for suppression of $Y$, refined. Correct structure learned and is stable from now on.}
    \label{fig:csvform}
\end{figure*}

\begin{figure}
     \centering
     \begin{subfigure}[t]{0.15\textwidth}
         \centering
         \includegraphics[width=\textwidth]{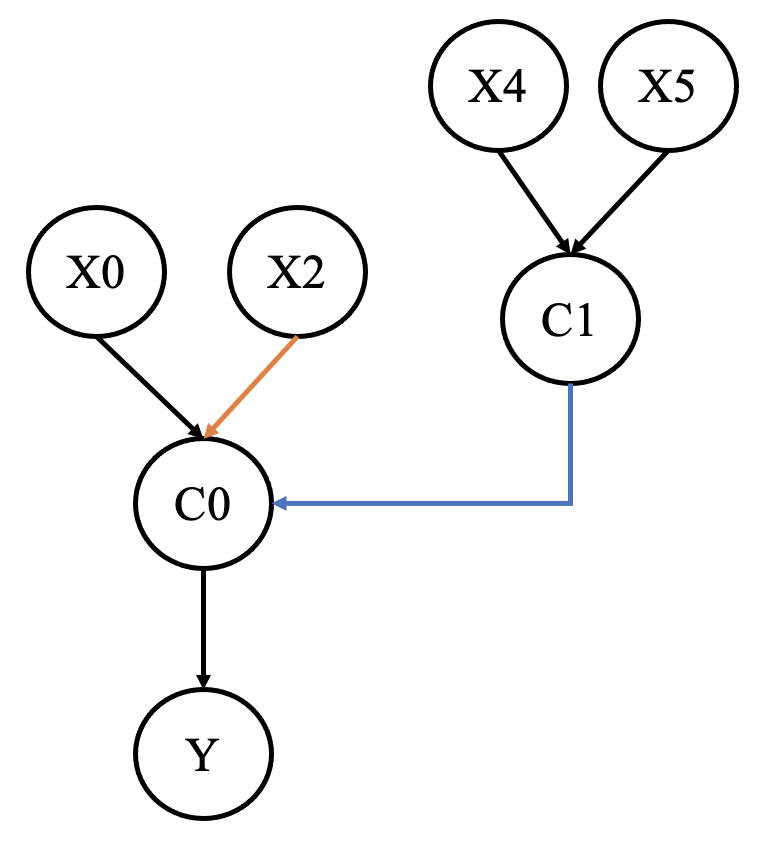}
         \caption{}
         \label{fig:csvformupstream_1}
     \end{subfigure}
     \vline
     \begin{subfigure}[t]{0.3\textwidth}
         \centering
         \includegraphics[width=\textwidth]{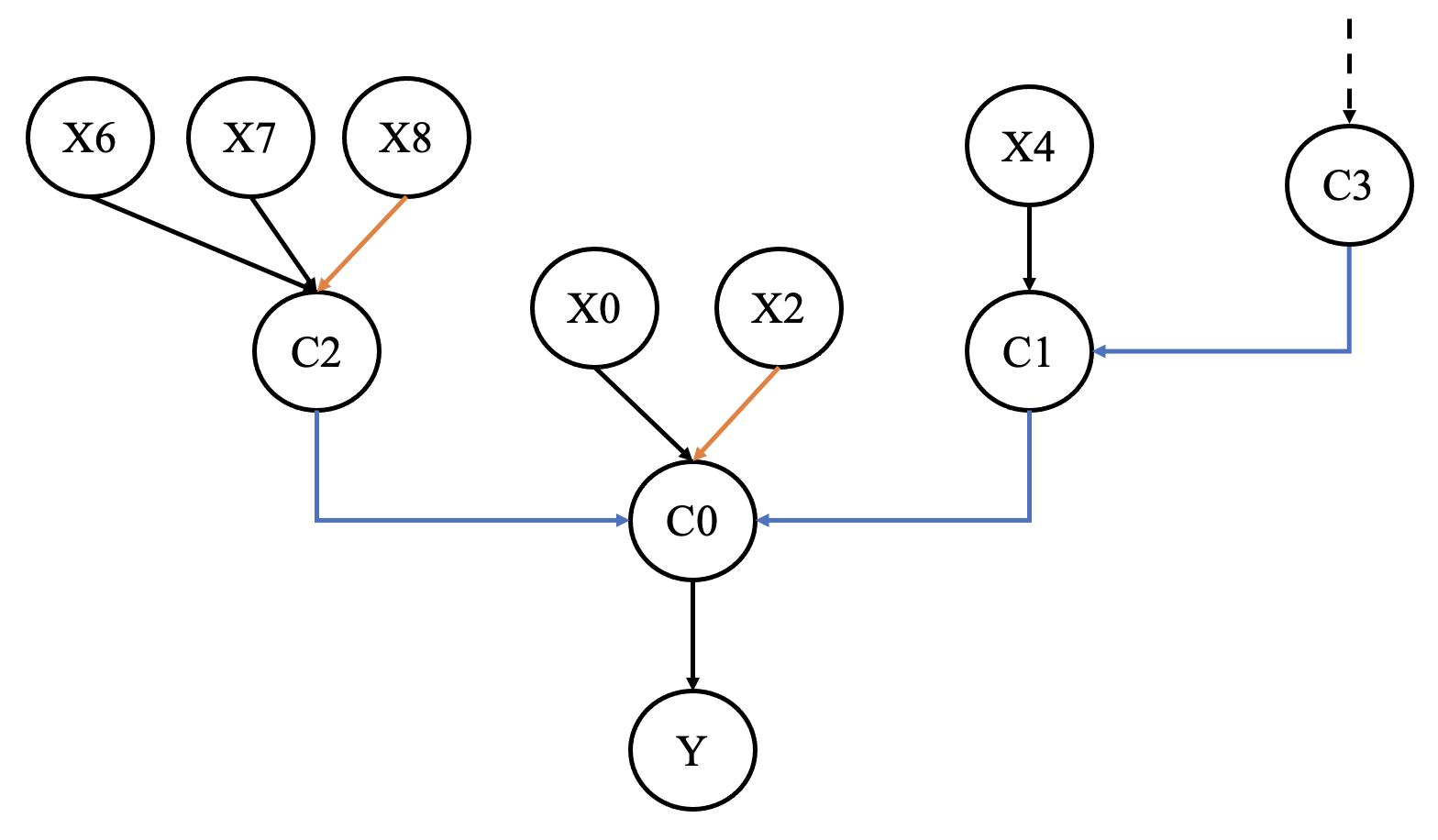}
         \caption{}
         \label{fig:csvformupstream_2}
     \end{subfigure}
    \caption{Example of upstream conditioning, continuing from Figure \ref{fig:csvform}. Assume that the unconditionality flag of $C0$ is set following an observation that $(X0,\ !X2)$ did not result in its activation (see main text). (a) $X0, !X2, X4, X5 \rightarrow Y$ observed. $C0$ is observed to be active. A new CSV $C1$ conditioning $C0$ formed. Note that $(X4, X5)$ alone will not predict activation of $C0$ if $C0$’s sources are not also active. (b) New conditioners are also subject to the CSV processes: Here, the source $X5$ of $C1$ has been refined, and new conditioners $C2$ and $C3$ are formed. Multiple conditioners represent alternative paths: In this case, $C0$ is expected to be active when sources of either $C1$ or $C2$ is active. Any logical function can hence be incorporated in a conditioning pathway in a minimal and ongoing manner without destroying past knowledge.}
    \label{fig:csvformupstream}
\end{figure}

Learning in The Modeller process proceeds \textit{step-by-step}, without incorporating an aggregate evaluation of multiple observational samples gathered from the environment, nor relying on iterative, repeated passes over a batch of data—distinct from traditional approaches. Initially, the model includes only BSVs and their DSVs, with no CSVs. At each step, The Modeller seeks to \textit{explain} the observed states of CSVs and DSVs in the previous timestep (modeling BSVs indirectly via DSVs). It does so by creating new CSVs to account for \textit{unexplained} DSVs and CSVs (those that are \textit{active} without an active conditioner – see Appendix \ref{sec:Modeller_details} for details), hence creating an adaptively-growing, flexible topology directed acyclic graph (see Fig.s \ref{fig:csvform}, \ref{fig:csvformupstream} for illustrations we will discuss shortly and Fig. \ref{fig:sample_model} for an experimental example). These \textit{retrospective explanations} for the observed events captured by CSVs become predictions for \textit{potential outcomes} in the next step. Notice that this design does not regard learning as an "iterative optimization" problem, but instead as a task of \textit{generating and identifying explanations}. As such, it represents a considerable shift from how the problem is conceptualized in the domain of contemporary machine learning.

The learning process of The Modeller is carried throughout the \textit{operations} on CSVs that apply at each step – their formation, and the modification of their positive and negative sources. Intuitively, learning is realized by the following cycle: A CSV, when formed to explain an event (an active DSV or another CSV without any valid standing explanation) starts by being connected to \textit{all} active SVs at formation, representing a \textit{comprehensive hypothesis of relationships} (Fig. \ref{fig:csvform_2}). These relationships are then \textit{refined} based on observations where some connections are deemed unnecessary, ensuring the CSV remains consistent with past observations locally (Fig. \ref{fig:csvform_3}). This exemplifies an iterative mechanism of \textit{local variation and selection,} while (as was mentioned above) the simple, unweighted connections with discrete input values (active/inactive/unobserved) reflect \textit{weak linkage}—two developmental principles detailed in Sec. \ref{sec:evolution_and_ai}.

More formally, these operations are defined as follows (detailed further on Section \ref{sec:Modeller_details} and Alg \ref{alg:algorithm_csvstate}):

\begin{itemize}
    \item \textit{Initial formation:} Fig. \ref{fig:csvform_2}. At each step, if there are active DSVs or CSVs without an explanation (an active conditioner or an unconditionality flag, see Appendix), a new CSV is generated to explain them. Initially, the CSV has no negative sources ($X_N = \emptyset$) and includes \textit{all} active BSVs and DSVs at that time as positive sources ($X_P$), representing the \textit{local variation} stage. No additional positive sources can be added to the CSV.
    \item \textit{Negative connections formation:} Fig. \ref{fig:csvform_4}. At the first instance where a CSV's sources are satisfied but its state is inactive, the CSV receives all active DSVs and BSVs at that time as negative sources ($X_N$), similar to previous step. No additional negative sources are added thereafter.\footnote{This process is separate from initial sources' formation to avoid creating exhaustive negative connections where unnecessary. Otherwise, a negative connection would be made with everything inactive during CSV creation, which, while accurate, would be overly exhaustive and unnecessary for most negative sources.}
    \item \textit{Refinements:} Fig.s \ref{fig:csvform_3} and \ref{fig:csvform_5}. When a CSV's state is determined as 1 with at least one active positive source and active targets, we remove nonactive positive sources (${x \in X_P : S_X \neq 1}$) from $X_P$ and active negative sources (${x \in X_N : S_X = 1}$) from $X_N$. When the state is 0, with at least one active positive source, inactive targets, and at least one active negative source, we remove nonactive negative sources (${x \in X_N : S_X \neq 1}$) from $X_N$. These refinements represent the \textit{local selection} stage.
\end{itemize}

With these operations in mind, we can summarize one learning step in The Modeller by being formed of the following stages (detailed further on Appendix \ref{sec:Modeller_details_learning} and Alg. \ref{alg:algorithm_adaptationloop}):

\begin{enumerate}
    \item The effective network created by DSVs and CSVs are traversed in the reverse order of computation (similar to backpropagation algorithm); starting from DSVs, then the CSVs that condition these BSVs, then the conditioners of these CSVs, and so on. Throughout the traversal, the states of all the model's state variables are computed according to their respective definitions above.
    \item For CSVs, sources and targets are modified to align the current states with the predictions/explanations provided by the CSVs, ensuring that the model remains consistent with the environment at each step (i.e. "refinement" and possibly "negative connections formation" operations) .
    \item New CSVs are generated for the DSVs and CSVs that currently lack an explanation. Each new CSV is formed by exhaustively assigning all currently active and eligible SVs as its positive sources (i.e. "initial formation" operations).
    \item  Finally, the model undergoes a refinement stage in which unnecessary state variables—specifically, those that are either duplicates of one another or who have no remaining sources or targets—are pruned to reduce representational complexity.
\end{enumerate}

These operations are central to, and realize, The Modeller's continual learning capability starting from the lowest level of organization, formalized by the following theorem:

\begin{theorem}
    \textit{Let $y_i$ be an \textit{instance} that includes the previous states of all the positive and negative sources of a CSV $C$ and the current states of all its conditioning targets. Then, if $C$ undergoes any modification as a result of encounter with an instance $y_1$, its state in reponse to any past instance $y_0$ is not altered by this modification; as long as its set of targets remain identical and $C$ does not undergo negative sources formation (either because inactive state is not observed or because it has already undergone it).\footnote{The requirement for identicality of targets in this theorem is only to account for the fact that heterogeneous targets result in duplication of CSVs - see the Appendix for details of this mechanism. The theorem holds when one considers the response of the duplicated CSVs with respect to the targets assigned to each duplicate as well.} For the proof, see Appendix \ref{sec:app_proof}.}
\end{theorem}

Theorem 1 is exemplified in Figure \ref{fig:csvform}: In \ref{fig:csvform_2}, after elimination of $X1$ as a positive source, the earlier exposure of $X0, X1 \rightarrow Y$ still results in a state of activity in $C0$, and likewise for $X2$ \& $X3$. With this property, we know that the state of a CSV in response to any past encounter is not altered except possibly for initial negative sources formation (happening only once per CSV), hence realizing continual learning without destructive adaptation in The Modeller inherently and from the lowest level of organization throughout the local preservation of past responses, with system-wide continual learning arising (as validated with our experiments below as well) as a sort of \textit{fractal property}. Importantly, \textit{none} of the standard assumptions made by existing continual learning methods—such as the use of replay buffers, predefined task boundaries, or access to large data batches—are required for this property. As such, The Modeller achieves continual learning intrinsically, free from these constraining assumptions.\footnote{For an alternative perspective on Theorem 1, we refer the reader to Appendix \ref{sec:app_lppr}, which examines the same property (local preservation of past responses) mathematically through the lens of the conditions required for it to hold in neural networks, which turn out to be generally not satisfied, and shows how The Modeller satisfies them.}

Additionally, we quantify the statistical significance of relationships between each CSV and their targets, which helps prevent excessively large models and instability in environments with spurious relationships—an aspect expected to be particularly important when scaling to higher-dimensional environments. Details of this mechanism are omitted from the main text for brevity, and can be found in Appendix \ref{sec:statistical_significance}.

\textit{\textbf{Upstream conditioning:}} A CSV can condition/predict not only direct environmental dynamics (DSVs) but also \textit{other CSVs}, enabling upstream complexity and the representation of arbitrary logical relations with minimal structure—without requiring a priori knowledge of their existence. Consequently, it is not constrained by assumptions like those in \cite{mordoch2023learning,verma2021asking, stern2017efficient} discussed in Sec. \ref{sec:background}. This formation of upstream conditioning pathways is illustrated in Fig. \ref{fig:csvformupstream}, extending the example from Fig. \ref{fig:csvform}. The processes of refinement, negative source formation, and further upstream conditioning remain identical regardless of the CSV’s target. Notice that in this way, a CSV acts as a potential \textit{point of regulation} between its sources and targets, with upstream generation following the same mechanisms. This allows previously learned representations to serve as foundations for new upstream ones as needed—exemplifying the principle of \textit{adaptation by regulatory modifications} prevalent in biological systems discussed in Sec. \ref{sec:evolution_and_ai}.

\textit{\textbf{Discussion:}} This learning approach is fundamentally different from methods like NNs. In The Modeller, the agent updates its model instantly with new information at each step, unlike methods that make incremental adjustments over many steps. This process can be seen as the agent initially "overfitting" to observations (fully accounting for them) while gradually refining the model to be as structurally and explanatorily minimal as possible without contradicting past experiences. At every stage, the model is as general as necessary based on prior exposures, but no more. The more specific representation (e.g., more sources per CSV) allows for precise generalization when new observations arise, increasing likelihood of consistency as sources are refined, reflecting the fundamental process local variation and selection in biology. Unlike conventional methods that start with underfitting and progressively adjust while avoiding overfitting, this concern is irrelevant in The Modeller, as the necessary level of generalization is inherently built into the model based on all previous observations.

\section{Integrating The Modeller with Behavior}

As discussed in Section \ref{sec:background}, contemporary ML lacks effective methods for integrating goal-oriented behavior with learning. The Modeller enables this integration through inherently structured representations with weakly linked components, allowing for both a goal-oriented planner and a behavior encapsulation mechanism to be incorporated into the system operation in a straightforward manner. In this section, we detail these two key components pertaining to behavioral agents.

\subsection{Planner}
\label{sec:planner}

Our planner operates on a model learned by The Modeller, demonstrating goal-directed planning by tracing backward from desired goal states to current states—as opposed to the forward sampling approach typical in methods like MBRL.

\textbf{\textit{Preprocessing the model and Group SVs:}} We first briefly preprocess a learned model to reduce the number of connections. To this end, we group the sets of BSVs in our that either (1) collectively act as positive or negative source of a CSV, or (2) have an event that is collectively predicted by a CSV. Each such grouping becomes a \textit{constituent} of a \textit{Group SV (GSV)}. For example, if a CSV $C0$ has positive sources $(B0, B1, B2)$ and predicts deactivation of $(B3, B4)$; then two GSVs are created: $G0 = (B0, B1, B2)$, $G2 = (B3, B4)$. This preprocessing stage, while not strictly required for planner operation, is crucial for ensuring the long-term scalability of The Modeller’s learned representations. Notice that this provides a simple example of a mechanism for transforming components of a base representation into abstract entities—a task that is ill-defined in architectures like neural networks but naturally straightforward in a learning framework grounded in developmental principles like The Modeller.

\textbf{\textit{Main process of the planner:}} (Alg. \ref{alg:planner} in the Appendix.) Our planner constructs an \textit{action network (AN)} based on a model generated by The Modeller, incorporating alternative outcomes. An AN is a dependency graph with root nodes representing the current environmental states (current BSV, GSV, and DSVs), along with possible alternative connections (shown by multiple conditioning links from CSVs) needed to achieve a specified goal state variable (see Fig. \ref{fig:fullAN} for an example from experiments).

To build this, we use a simple recursive function that generates the upstream action network for a given node (see Figure \ref{fig:angen} for an illustration and Alg. \ref{alg:planner} in Appendix for the full description). At each call, the function operates as follows:

\begin{itemize}
    \item \textit{Predecessors} of the argument node are added to the action network based on the current topology of the model. These predecessors to be "expanded" vary by the state variable type of the current node (see also Figure \ref{fig:angen2}):\begin{itemize}
        \item For CSVs, their upstream conditioners and sources;
        \item For GSVs, their constituents, constituencies and precondition events;
        \item For DSVs, their precondition states and conditioners.
    \end{itemize}
    Note that for CSVs and DSVs, different predecessors represent \textit{alternative pathways} they can be activated.
    \item The function recalls itself on all these "opened" predecessors until it reaches the root nodes that represent current environmental states.
\end{itemize}

\begin{figure}
     \centering
     \begin{subfigure}[t]{0.15\textwidth}
         \centering
         \includegraphics[width=\textwidth]{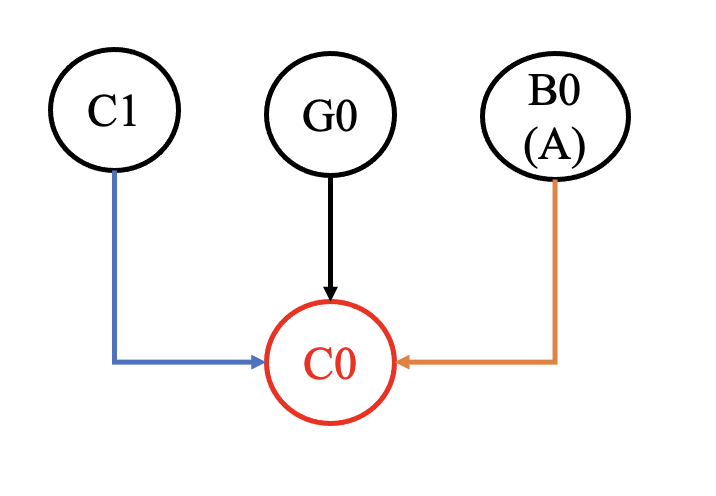}
         \caption{}
         \label{fig:angen1}
     \end{subfigure}
     \vline
     \begin{subfigure}[t]{0.33\textwidth}
         \centering
         \includegraphics[width=\textwidth]{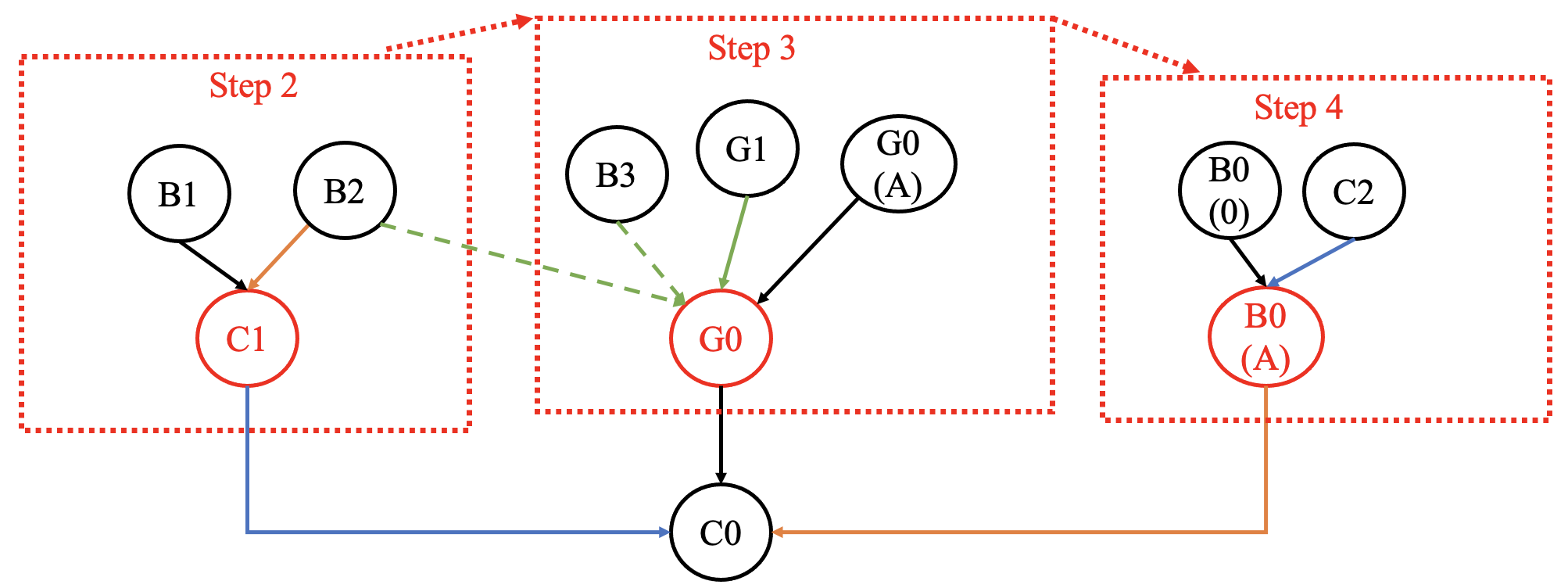}
         \caption{}
         \label{fig:angen2}
     \end{subfigure}
    \caption{Illustration step-by-step upstream generation of action network, operating on different SV types (see the text for details). $BX$, $CX$ and $GX$ stand for BSV, CSV and GSV nodes respectively, (A) for activation, (0) for nonactive state. Black arrows are positive sources and precondition targets, green arrows are constituent (dashed) and constituency (solid) relations. The node that is extended at each step is highlighted in red. (a) Step 1. CSV $C0$ is opened. (b) Steps 2-4. Each step opens up one of the sources of previous step. Possible interrelations (e.g. $B2$ for $C1$, $G0$) do not need reopening if they already exist.}
    \label{fig:angen}
\end{figure}

\textbf{\textit{Action choice:}} The agent generates an action network each time it needs to select an action. (While this is computationally unnecessary—since an AN can be reused until the agent reaches the goal—we maintain this approach for simplicity.) From the generated AN, the agent identifies actions that can immediately activate any CSV in the action model, specifically those whose sources and sources of their downstream targets do not involve any unactualized BSV states. The agent then randomly selects one of these actions for the current step. Since only one action is chosen, the agent can consider the entire AN including alternative pathways.

Our planner is explicitly goal-directed, identifying a path from initial states to the goal without needing rewards, although rewards can help prioritize the search. Unlike methods like MBRL (Sec. \ref{sec:background}), which typically search from initial states to goals via forward-sampling, our planner considers both initial and goal states, focusing on steps derived from the environment model. The planning algorithm is a simple search method that unfolds upstream action networks from the model, as our main aim is to demonstrate the interface between The Modeller’s modeling components and general deliberative behavior without going into extensive detail. Planning is a well-established field with efficient methods and useful heuristics \cite{ghallab2016automated}, and once the interface between The Modeller and planning is established, implementing more advanced algorithms is straightforward. We note, however, two visible limitations of the current version of the planner: First, the generated action networks are exhaustive, including every possible path to initial states. Second, the current version does not account for the precise timing of multiple events.\footnote{In our experiments, for instance, the RS environment subtype (see Figure \ref{fig:environment}) typically takes longer due to the BSV \textit{DO} having two pathways for deactivation, the correct one being the one that deactivates BSV \textit{W} as well at the same time. The planner fails to distinguish between these pathways, leading to some unnecessary loops.} These limitations are not addressed in current framework to keep its simplicity, since they do not affect our demonstrative use of the planner to a major degree. They are discussed in Sec. \ref{sec:conclusion}.

In summary, the operation of an agent utilizing The Modeller and the planner is formed of the following steps, repeated continuously as the agent interacts with the environment in an online manner, without the need for episode division or learning iterations over collected bulk data:
\begin{enumerate}
    \item Execute actions and gather observations.
    \item Process observations and update the model (Modeller)
    \item Generate a plan based on the current model and goals, select an action from the resulting plan (Planner)
\end{enumerate}

\subsection{Behavior Encapsulation}
\label{sec:behavior_encaps}

\begin{figure}[t]{}
     \centering
     \includegraphics[width=0.4\textwidth]{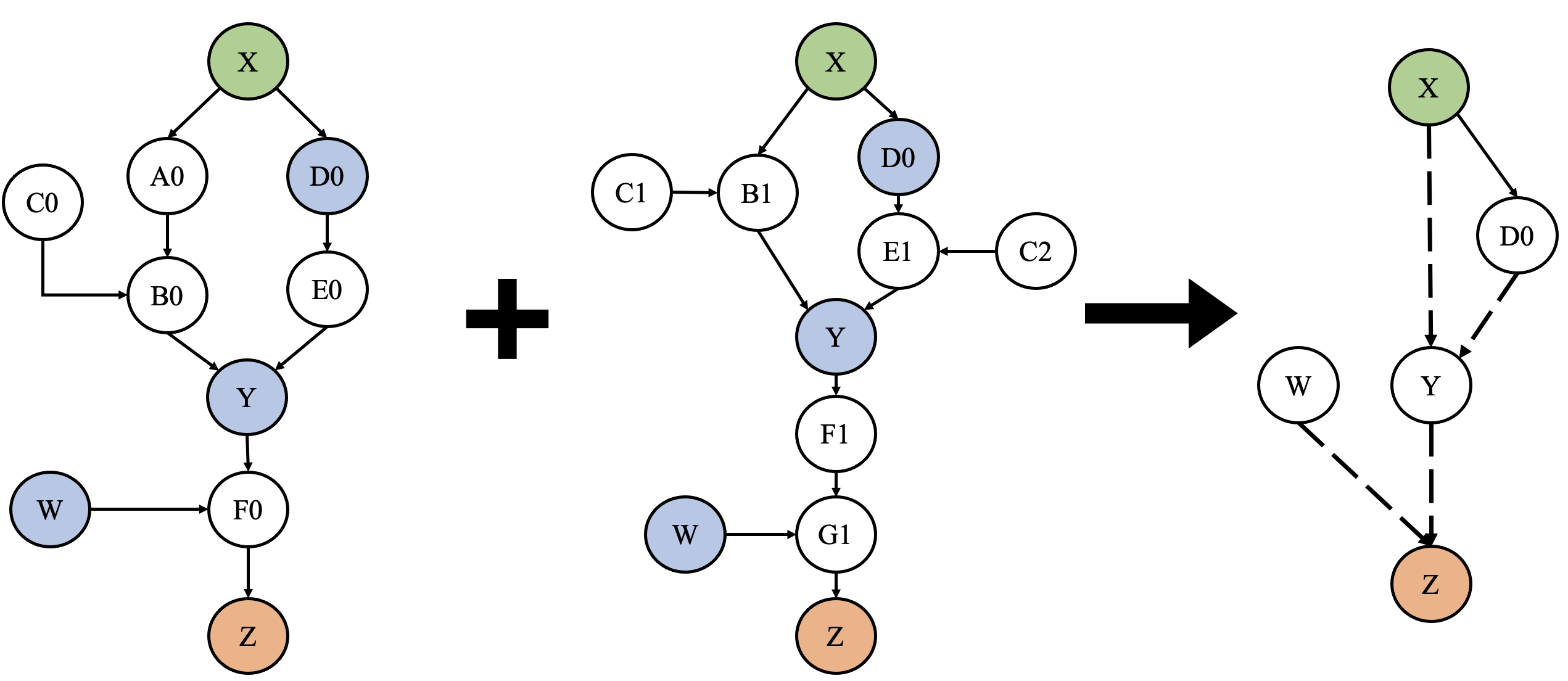}
     \caption{Illustrative example for the aim of behavior encapsulation process. To the left are two action networks that represent two alternative pathways, split from the unified AN generated by the planner (node names are placeholders and can be any SV type or target effect). We want to encapsulate the pathways between X and Z. For that; all paths that are reliably present in both networks are identified and a new \textit{encapsulated AN (EAN)} is formed with them (right). Each encapsulated edge (dashed) in EAN includes copies of subnetworks that corresponded to this path in the original AN variants; which can be further encapsulated in subgroups via a recursive call (e.g. edge (D0,Y) would include two pathways; first one formed only of E0, the second of C2 and E1). The EAN on right can be regarded as the \textit{subpolicy} for realization of Z from X.}
     \label{fig:encapsulation_example}
 \end{figure}

The Modeller and our planner design together create a complete system capable of continual learning and structured goal-directed behavior. However, the exhaustive action networks produced by the planner do not exemplify a comprehensible representation, which is one of our key goals. Additionally, the planner does not fully leverage this structured representation to address a long-standing challenge in behavioral AI: the decomposition of behavior into subunits defined by automatically determined preconditions and consequences in an arbitrarily hierarchical manner. To address this, we introduce a \textit{behavior encapsulation} mechanism that operates on the action networks generated by the planner, transforming flat, exhaustive ANs into a hierarchically structured and comprehensible format. This is carried out by a process composed of the following steps (detailed in Alg. \ref{alg:behenc} in the Appendix):

\begin{itemize}
    \item As detailed above, the action network produced by the planner contains multiple alternative pathways via the predecessors of DSVs and CSVs. Our first step is to isolate these pathways into distinct alternative action networks. This is done by duplicating the original network, with each copy retaining only one conditioning alternative per CSV and DSV—along with the upstream predecessors connected to the selected alternatives.
    
    \item We then construct a reduced, high-level network (\textit{encapsulated action network}) that captures the consistently observed pathways across all alternative ANs (see Figure \ref{fig:encapsulation_example} for an abstract example, and Figure~\ref{fig:encapsulatedAN} for a specific case from our experiments). Nodes in this graph represent subgoals necessary for the current objective, while encapsulated edges denote subpolicies linking their respective start and end states. This is achieved through a straightforward edge-centric procedure: starting from one AN, we iteratively remove edges whose endpoints aren't preserved across the other ANs, while linking their predecessors and successors so that the paths between them are retained. The process repeats until no further changes occur, yielding a minimal structured form (function \textit{constructEncapsulatedAN} in Algorithm \ref{alg:behenc}).

    \item Once the high-level encapsulated network is built, we extract subgraphs that connect subgoal nodes present in this network—representing alternative subpolicy pathways. This involves collecting all nodes and edges linking these subgoal nodes across the original action networks, along with their upstream predecessors, and inducing a subnetwork from the combined set. Each resulting subnetwork collected from one of the original action networks corresponds to one such alternative pathway (see function \textit{getConnectingSubnetwork} in Algorithm~\ref{alg:behenc}, and Fig.~\ref{fig:encapsulatedAN} for an example).
    

    \item Finally, we recursively apply the entire process to the encapsulated subnetwork groups that connect two subgoal nodes, as generated in the previous step, grouping networks that share at least one common node. This continues until no further grouping is possible (see outer function \textit{EncapsulateBehavior} in Algorithm~\ref{alg:behenc}).
    
\end{itemize}

These operations result in a behavior representation that, while complex in its extended form, is maximally structured and comprehensible at each organizational level. Although this process is computationally intensive, it only needs to be executed once for each action path, as long as the underlying model remains unchanged, making the computational complexity manageable.

Beyond enhancing the comprehensibility of action networks \textit{post-hoc}, this encapsulation process can significantly aid agent behavior. Encapsulated behavioral subunits (whose delimiters are not provided to the agent in advance) can be reused when the same precondition/goal pairs arise, akin to subpolicies in HRL. We do not yet perform this integration of behavior encapsulation with agent's ongoing operations, and present it separately as an illustration of what becomes possible with our approach, but we note it as an important venue for the future.

\section{Continually learning structured visual representations}
\label{sec:mnr}

We now aim to extend our modelling framework to operate on \textit{networks} as observations, rather than raw state variables. Networks can represent diverse domains with a higher dimensionality than what can be represented by lists of state variables, in particular spatial and temporal observations relevant to AI. Our focus here is on vision, specifically 2D shape identification, though the approach is generalizable to any network-representable observation provided an appropriate representational conversion is used. The method is agnostic to how a 2D image is converted into a network. We detail our experimental feature representation in Section \ref{sec:feature_representation}, until then, the reader can think about a generic concept of a visual feature that can refer to edges, colors, gradients, objects, or even raw pixels. We refer to the mechanism described in this section as \textit{Modeller with network refinement} (\textit{MNR} for short). Its design, detailed below, primarily follows from and expands upon, the network refinement mechanism we described for behavior encapsulation in Section \ref{sec:behavior_encaps}.

\subsection{Representational basis for MNR}
\label{sec:mnr_representation_basis}

First, let's define the basis for our representation of observations and input sources for CSVs in this extension:

\begin{definition}
    \label{def:sn_spn}
    \textit{A \textbf{state network (SN)} is a directed graph $(N,E)$, where each node $N$ has an associated \textbf{type}. A list of state networks and their keys, $P=[(k_0, SN_0), (k_1, SN_1), ...]$ is called a \textbf{state polynetwork (SPN).}}
\end{definition}

Node types in state networks represent distinct features (e.g., edges, corners, or objects in visual space), with nodes being observed instances of these features in the current observation (e.g., two edges with the same orientation or two instances of the same object are distinct nodes of the same type; see Figure \ref{fig:assignment_source} in the appendix for an example). Edges ($E$) represent relations between nodes, such as relative positions in visual inputs or succession in temporal domains. A state polynetwork is a collection of distinct state networks with a designator key, enabling the definition of different feature and relation types. In visual space, this could include shape, color gradients, or abstract objects, as well as multi-dimensional relations (e.g., relative positioning). An example SPN is shown in Figure \ref{fig:representation_example_vn} (the means of construction is detailed in the next section).

SPNs will serve as input sources for CSVs in our model, replacing the sets of SVs described in Sec. \ref{sec:Modeller}. When generating a CSV, instead of establishing exhaustive connections with all active state variables at a given moment, we now use the currently observed SPN in the environment as the direct genesis source for that CSV (the new \textit{variation} process). As a result, learning in the model now involves constructing SPN structures that encapsulate the required information—raising the question of how a standing SPN (representing a CSV’s input configuration) should be modified, or “refined” (the new \textit{selection} process) in response to new SPN observations.

\subsection{Network refinement with rerelation}
\label{sec:netref_description}

Refinement, the core learning process in \textit{The Modeller}, reduces two lists of state variables to their intersection. An analogous operation is needed to identify the \textit{shared part} of two or more state polynetworks. For that purpose, we define:

\begin{definition}
    \label{def:spn_satisfied}
    \textit{An SPN $P_0=[(k_0, SN^0_0),\ (k_1, SN^0_1),$ $...(k_N, SN^0_N)]$ is \textbf{satisfied by} another SPN $P_1=[(k_0, SN^1_0), (k_1, SN^1_1), ... (k_N, SN^1_N)]$ (with the same set of keys  $K=[k_0, k_1, ... k_N]$) given a potentially partial \textit{assignment} $f: V(P_0) \rightarrow V(P_1)$, where $N(P_i)$ is the set of all nodes across all state networks of $P_i$, if and only if the following conditions hold: (1) For $\forall n_0 \in N(P_0)$, $f(n_0)$ is defined (has a mapped node in $N(P_1)$), and (2) For $\forall e_0=(n_0, n_1) \in E(SN^0_i)$ where $E(SN^0_i)$ is the set of all edges in state network $SN^0_i$ in $P_0$, there exists a path in $SN^1_i$ from $f(n_0)$ to $f(n_1)$.}
\end{definition}

Intuitively, $P_0$ is satisfied by $P_1$ under an assignment if every node in $P_0$ has a corresponding target in $P_1$, and every edge in $P_0$ has a path in $P_1$ connecting the assigned targets of its endpoints within the same SN. This ensures that all relations in $P_0$ are present in $P_1$, even if mediated by additional entities not in $P_0$ (since paths, not direct edges, are required).

We can now redefine "finding the intersection" of two SPNs $P_0$ and $P_1$ as "minimally refining $P_0$ to be satisfied by $P_1$." This is achieved through \textit{network refinement with rerelation}, where $P_0$ (source) is refined by $P_1$ (refiner). The process, detailed in Algorithm \ref{alg:netref}, relies on two subprocesses:

\begin{itemize}
    \item \textit{Refinement:} Nodes in $P_0$ that are missing in $P_1$, and edges in SNs of $P_0$ that don't have a path between their endpoints in the corresponding SN of $P_1$, are removed.
    \item \textit{Rerelation:} When an edge $(n_0, n_1)$ is removed (including via node removal), a new edge $(p_i, s_i)$ is created for $\forall p_i \in P(n_0), s_i \in S(n_1)$, where $P(n)$ and $S(n)$ are predecessors and successors of node $n$ respectively. (Each edge formed by rerelation is checked for the same conditions of presence as existing ones.)
\end{itemize}

\begin{figure}
     \centering
     \begin{subfigure}{0.15\textwidth}
         \centering
         \includegraphics[width=\textwidth]{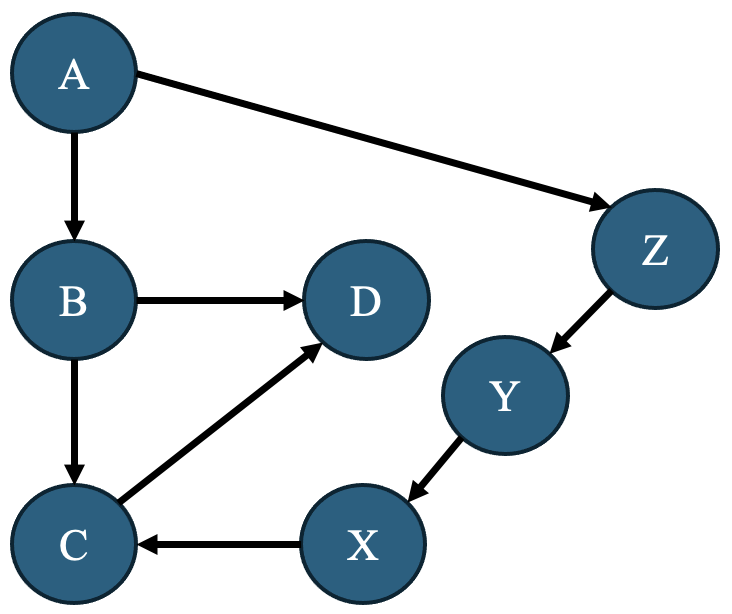}
         \caption{Source SN.}
         \label{fig:netref_source}
     \end{subfigure}
     \begin{subfigure}{0.15\textwidth}
         \centering
         \includegraphics[width=\textwidth]{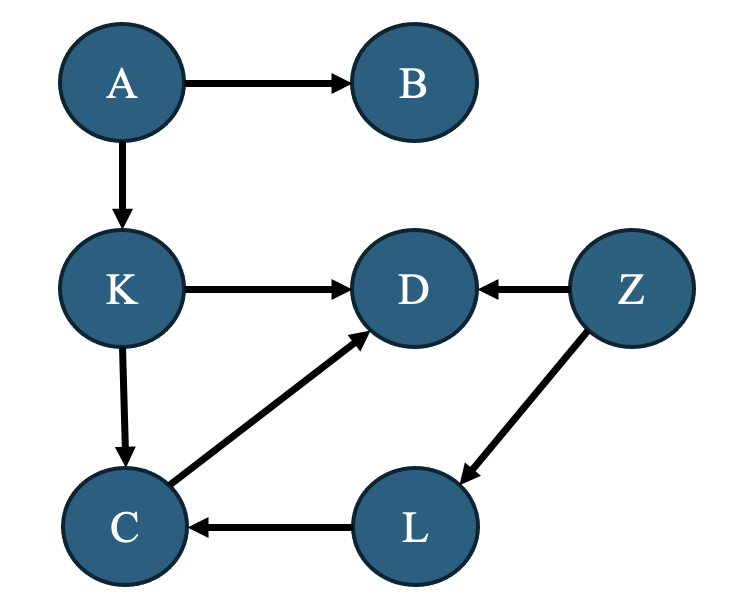}
         \caption{Refiner SN.}
         \label{fig:netref_refiner}
     \end{subfigure}
     \begin{subfigure}{0.15\textwidth}
         \centering
         \includegraphics[width=\textwidth]{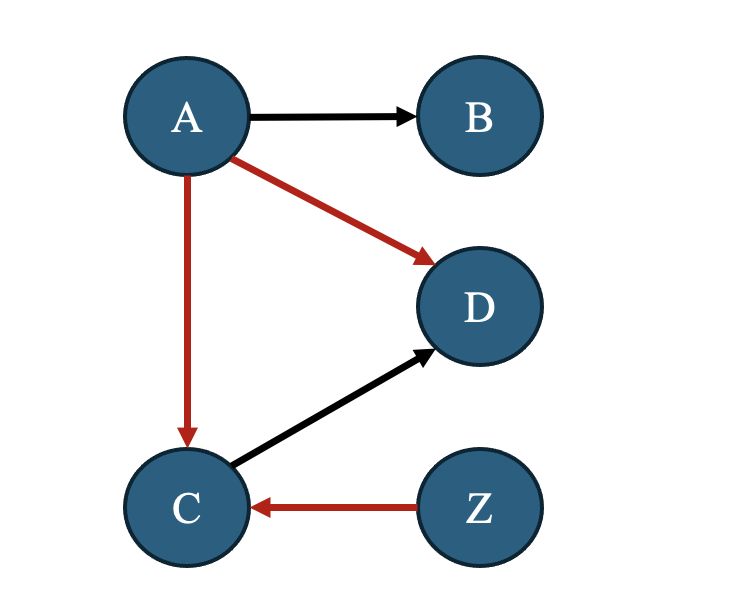}
         \caption{Source, refined.}
         \label{fig:netref_final}
     \end{subfigure}
    \caption{Illustration of network refinement with rerelation. In (c), highlighted edges are created through rerelation. Paths $(A,D)$ and $(A,C)$ exist in both networks but are mediated by different intermediaries (B and K respectively), leading to refined intermediaries and new edges. Similarly, path $(Z,C)$, mediated by $(Y,X)$ in the source and $(L)$ in the refiner, is refined. Edge $(A,Z)$ is removed as it lacks a corresponding path in the refiner SPN. Edge $(A,B)$ is preserved unchanged, as it appears in both networks, despite differing successors of B. (Node positions are illustrative and irrelevant to refinement.)}
    \label{fig:netref}
\end{figure}

Figure \ref{fig:netref} illustrates this process: the source SPN in \ref{fig:netref_source} is refined by the refiner in \ref{fig:netref_refiner}, resulting in the refined SPN in \ref{fig:netref_final}. Paths like $(A,D)$ or $(A,C)$ are preserved despite differing intermediaries. Applying this process sequentially to a source SPN across multiple refiners results in a final SPN representing commonalities across all refiners, ensuring it is satisfied by each refiner retrospectively.\footnote{The local continual learning guarantee in \textit{The Modeller} (Theorem 1) also applies to Alg. \ref{alg:netref}: Nodes and paths in an SPN are analogous to state variables in the base version, thus the theorem's proof holds for node and path removal, with edge refinement being more constrained than path refinement.} The system-level learning flow of MNR is identical to that of base \textit{The Modeller} except for a few small adjustments, which are detailed on Appendix \ref{sec:learning_flow}.

\textit{\textbf{Statistical Refinement}} Given the noisy experimental domain, we enhanced Algorithm \ref{alg:netref} by incorporating node/edge observation statistics. Instead of removing a node/edge upon its first absence, we remove it only if the ratio of its absences exceeds a threshold $T_{ref} \in (0,1)$. This prevents losing important features potentially missed due to noise or misassignment (see below). Being a more constrained removal condition, this maintains \textit{The Modeller}'s continual learning guarantees.

\textit{\textbf{Assignments:}} A key issue in this process is finding a suitable (possibly partial) node mapping $f: V(P_0) \rightarrow V(P_1)$ between two SPNs, ensuring that source nodes map only to nodes of the same type. Multiple valid mappings may exist, leading to different refined structures (see Figure \ref{fig:assignment} in the Appendix). To tackle this, we generate a population of alternative assignments by pairing nodes with shared types. While this works for small node sets, our feature representation (Section \ref{sec:feature_representation}) often requires additional prioritization. For that, we rank node pairs by their positional proximity in both SPNs, using negative softmax of their distances to probabilistically select the best matches. After generating the population, we compute a mismatch score for each assignment as the total number of nodes and paths in the source SPN missing in the refiner SPN. The assignment with the lowest mismatch is chosen. This method yielded effective mappings in our experiments, though it remains the most computationally demanding step of our algorithm. Further research could likely improve this process or even eliminate the need for such assignments altogether.

\subsection{Feature representation for basic shape learning}
\label{sec:feature_representation}

\begin{figure}
     \centering
     \begin{subfigure}[t]{0.14\textwidth}
         \centering
         \includegraphics[width=\textwidth]{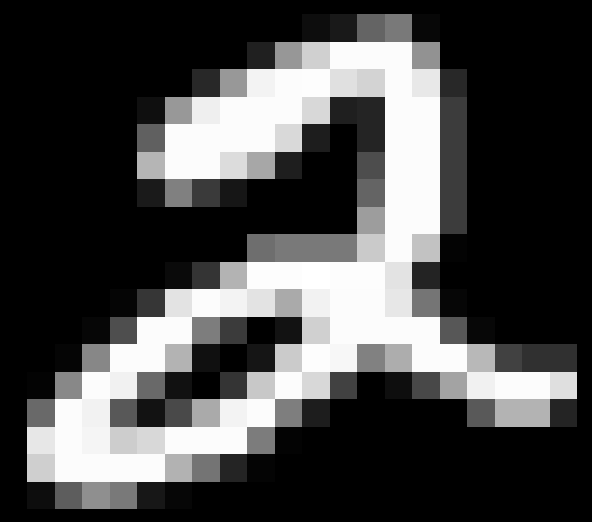}
         \caption{Original image.}
         \label{fig:representation_example_im}
     \end{subfigure}
     \hfill
     \begin{subfigure}[t]{0.15\textwidth}
         \centering
         \includegraphics[width=\textwidth]{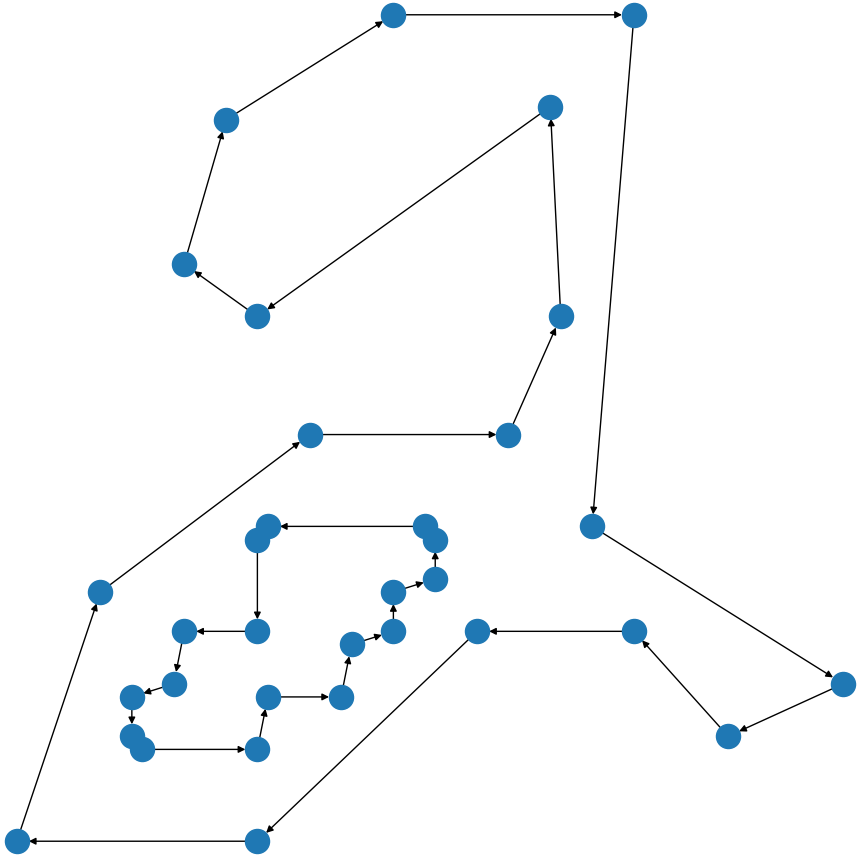}
         \caption{Approx. contours.}
         \label{fig:representation_example_bcn}
     \end{subfigure}
     \hfill
     \begin{subfigure}[t]{0.17\textwidth}
         \centering
         \includegraphics[width=\textwidth]{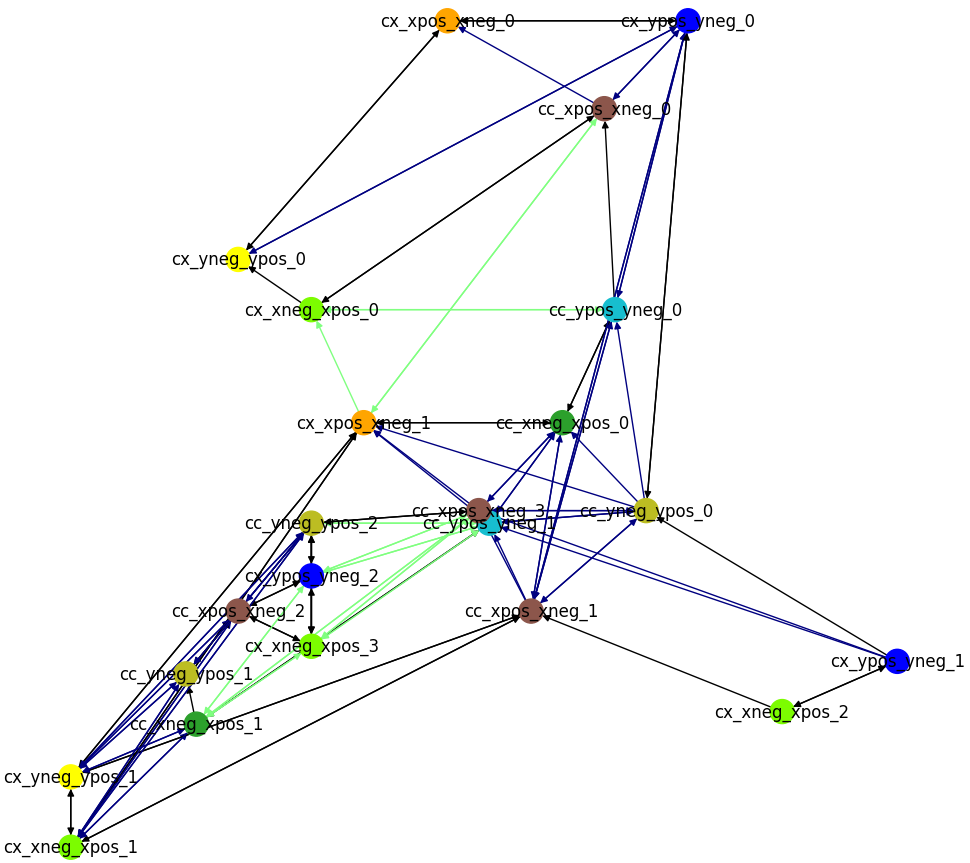}
         \caption{Final SPN.}
         \label{fig:representation_example_vn}
     \end{subfigure}
     \caption{Example SPN construction for 2D shape identification from a given image. SPN is drawn with all networks together. Blue, green, and black edges denote inner, outer, and contour connections, respectively. Nodes, colored by type, represent gradient change points (detailed in main text).}
    \label{fig:feature_representation}
\end{figure}

MNR is applicable to any observation space that can be represented as a network. In our experiments, we focus on demonstrating the method in a basic visual processing domain—2D shape identification—using MNIST as the testbed. Due to space constraints, the details of the feature representation are provided in Appendix \ref{sec:app_feature_representation}. To summarize briefly, we first generate a polygonal approximation of the shape from the binarized image of the digit. Features are then extracted based on the points of gradient orientation change along the $x$ and $y$ axes, effectively capturing second-order variations in shape. An illustrative example is shown in Fig.~\ref{fig:feature_representation}.

This representation is broadly applicable and domain-agnostic for 2D shape detection, relying solely on edge detection. However, it primarily serves to demonstrate our design and has inherent limitations—most notably, limited expressivity due to capturing only gradient sign changes rather than all shape details. This limitation accounts for the detection accuracy shortcomings discussed in the following sections. Future work can extend this approach to more expressive and comprehensive feature representations for 2D shape detection, as well as to other shape types, including 2D features such as color gradients or 3D features incorporating spatial positions, depending on domain requirements. We do not explore these alternatives here, as they fall outside the scope of this work, which focuses on demonstrating the effectiveness of the \textit{learning mechanism} itself (which is \textit{independent} of the particular feature representation being used) rather than the specific feature representation that supports it. It should also be noted that this feature representation applies only to 2D shape detection, which is why our experiments are conducted on MNIST rather than more complex datasets like CIFAR. The latter would require extensions to the feature representation beyond 2D shape detection, a topic beyond the scope of this work centered on validating the learning mechanism.

\section{Experimental Validation 1: Base Modelling and Behavior}
\label{sec:experiments_Modeller}

\subsection{Setup}

\begin{figure}[t]{}
     \centering
     \includegraphics[width=0.4\textwidth]{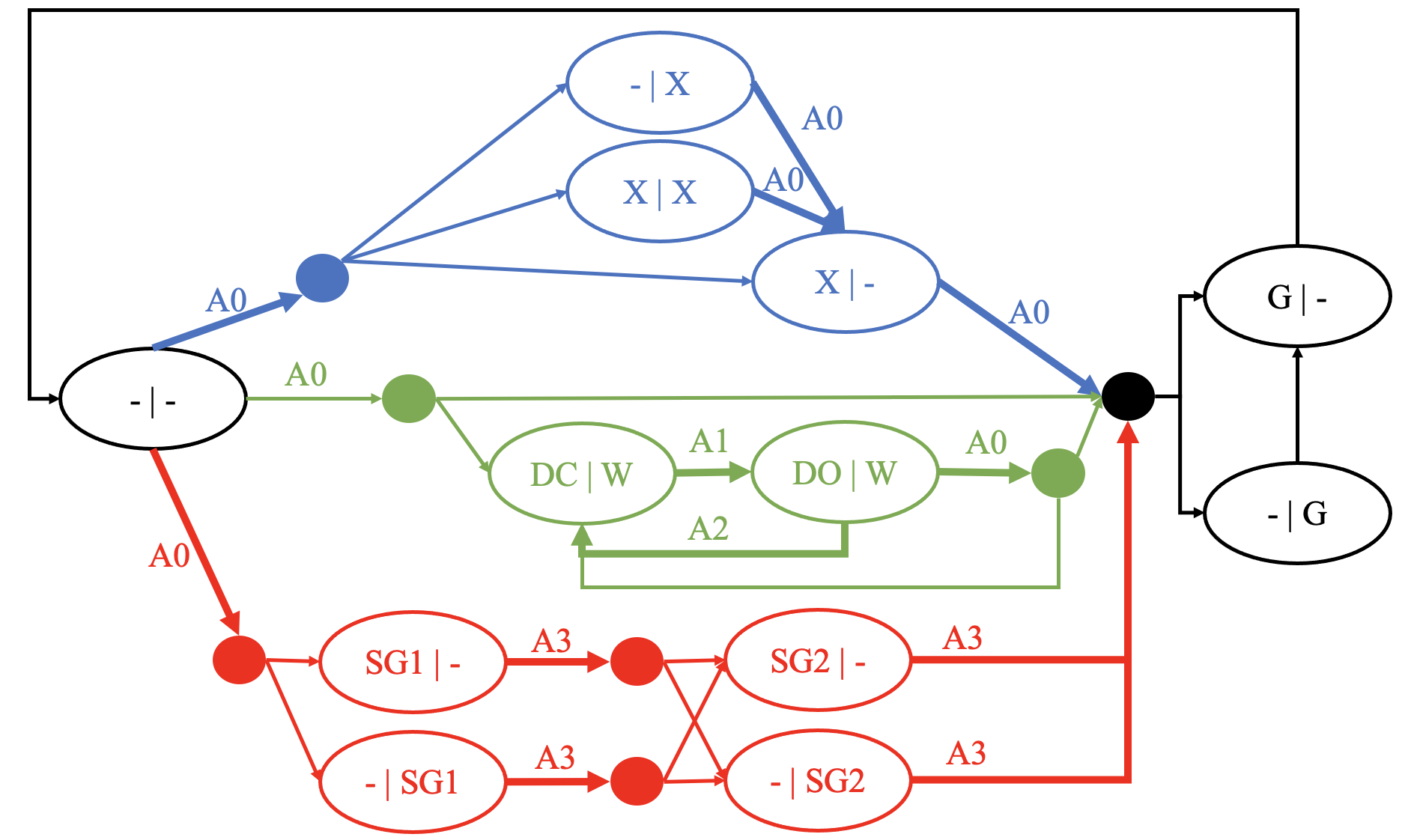}
     \caption{The environment and its subenvironments that we test on, essentially a FSM with two cells each of whom can take one of the states "DO, DC, W, G, SG1, SG2, X" or be empty ("-"). Each state is connected with arrows representing succession relations between them; filled circles correspond to multiple alternatives that can result from it. Green, red and blue portions are "RS", "SG", and "NEG" subtypes respectively (detailed in text), black portion is included in all subtypes. In "Complete" variant, all transitions and states are included. The agent’s goal is to activate state "G" in the first cell, and optimal actions are indicated by bold transitions. The environment has 20 actions, much larger than what is actually useful, in order to make it difficult to reach goal randomly.}
     \label{fig:environment}
 \end{figure}

We demonstrate the operation of The Modeller combined with behavioral mechanisms on a simple test environment, which is a finite-state machine (FSM) with two cells, each capable of seven states or inactivity, as shown in Figure \ref{fig:environment}. The environment includes three subtypes (RS, SG, NEG), illustrated by different colors. This setup was designed to model various types of temporal successions, such as basic succession, correlated changes, alternative causes/outcomes, uncertain transitions, and negative conditons.\footnote{The environment was vaguely inspired from Multiroom environment in Minigrid \cite{MinigridMiniworld23}. For intuition behind this FSM, see the Appendix.} There is also a \textit{random} variant of the environment where two additional states that get activated randomly are introduced, in order to test statistical significance filtering mechanisms. This environment was chosen in order to validate the core operation of our design in a simple and understandable setting, which made in-depth analysis and debug of the design very feasible during development process. There is no inherent limitation to applying to more complex environments, except that the planner implementation should incorporate the changes needed to make search nonexhaustive, and possibility of non-Markovian transitions should be accounted for (see Sections \ref{sec:planner} and \ref{sec:conclusion}). We leave validation on such environments and changes in design as venues for future developments, as this presentation is dense enough as it is.

In our base planning experiments, we compare the performance of an agent that learns a model followed by planning (with a 10\% chance of random actions for exploration) to one that acts purely randomly. The agent starts with 4000 steps of random actions to learn the environment model, then uses the planner for the next 4000 steps. We measure the average steps to reach the goal before and after planning. Next, we conduct continual learning experiments where the agent learns with predefined goals and the environment subtypes switch every 500 steps (with readaptation) or 1000 steps (without readaptation). We test whether the agent can achieve similar performance in different subtypes, both in vanilla and random environment variants without any readaptation of the model, and also analyse learning progression when readaptation is enabled. Finally, we present a demonstrative case of behavior encapsulation on a learned model. For more details on the experimental setup, see Appendix \ref{sec:appendix_expsetup}.


As detailed in Section \ref{sec:background}, to the best of our knowledge, no existing method in the literature is capable of (i) reliably performing unsupervised continual learning of an environment without task boundaries or past sample replay, (ii) executing precise goal-directed behavior on a learned model, or (iii) encapsulating and representing behavior in an automatically generated, arbitrarily hierarchical, and comprehensible structure—let alone addressing all these seemingly disparate challenges within a unified framework. Therefore, we do not include comparisons with existing neural network-based methods in the main text, as we believe they would not provide meaningful insight. For completeness, however, demonstrative experiments and comparisons involving planning with a neural network-based model are presented in Appendix \ref{sec:nn_plan_exps}, and likewise, the performance of continual learning methods under their own limiting assumptions is shown in Appendix \ref{sec:nn_cl_exps}. Both are excluded from the main text due to space limitations and their limited contribution to the central narrative. These sections, which the skeptic reader can refer to, clearly illustrate the shortcomings—and in some cases, outright failure—of neural network-based approaches, even under the current simplified experimental conditions, further underscoring the significance of our proposed design.

\subsection{Results and Discussion}

\begin{figure}[t]{}
     \centering
     \includegraphics[width=0.4\textwidth]{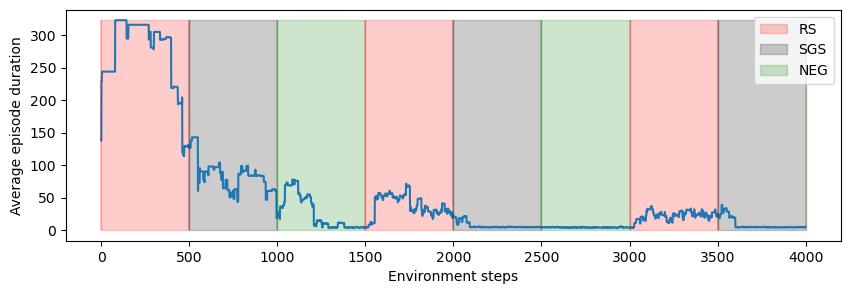}
     \caption{Average (5 trials) episode durations throughout learning with changing environment subtypes (background colors), with model readaptation enabled. Vertical limits show the environment changes, note that the actual step of change varies by a few steps since end of the ongoing episode is awaited.}
     \label{fig:CL_planning_nonrand_readapt}
 \end{figure}

\begin{figure}
     \centering
     \begin{subfigure}[b]{0.20\textwidth}
         \centering
         \includegraphics[width=\textwidth]{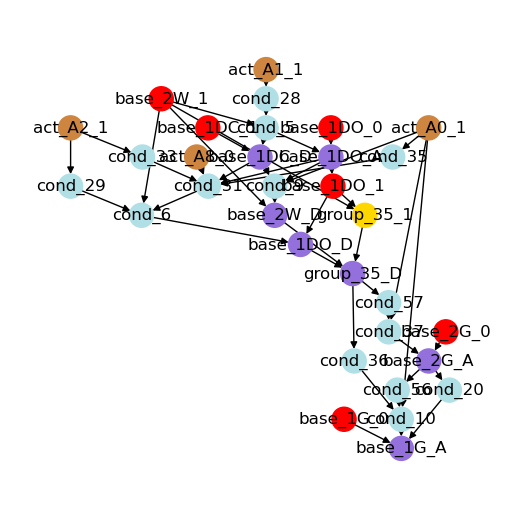}
         \caption{Full action network.}
         \label{fig:fullAN}
     \end{subfigure}
     \begin{subfigure}[b]{0.26\textwidth}
         \centering
         \includegraphics[width=\textwidth]{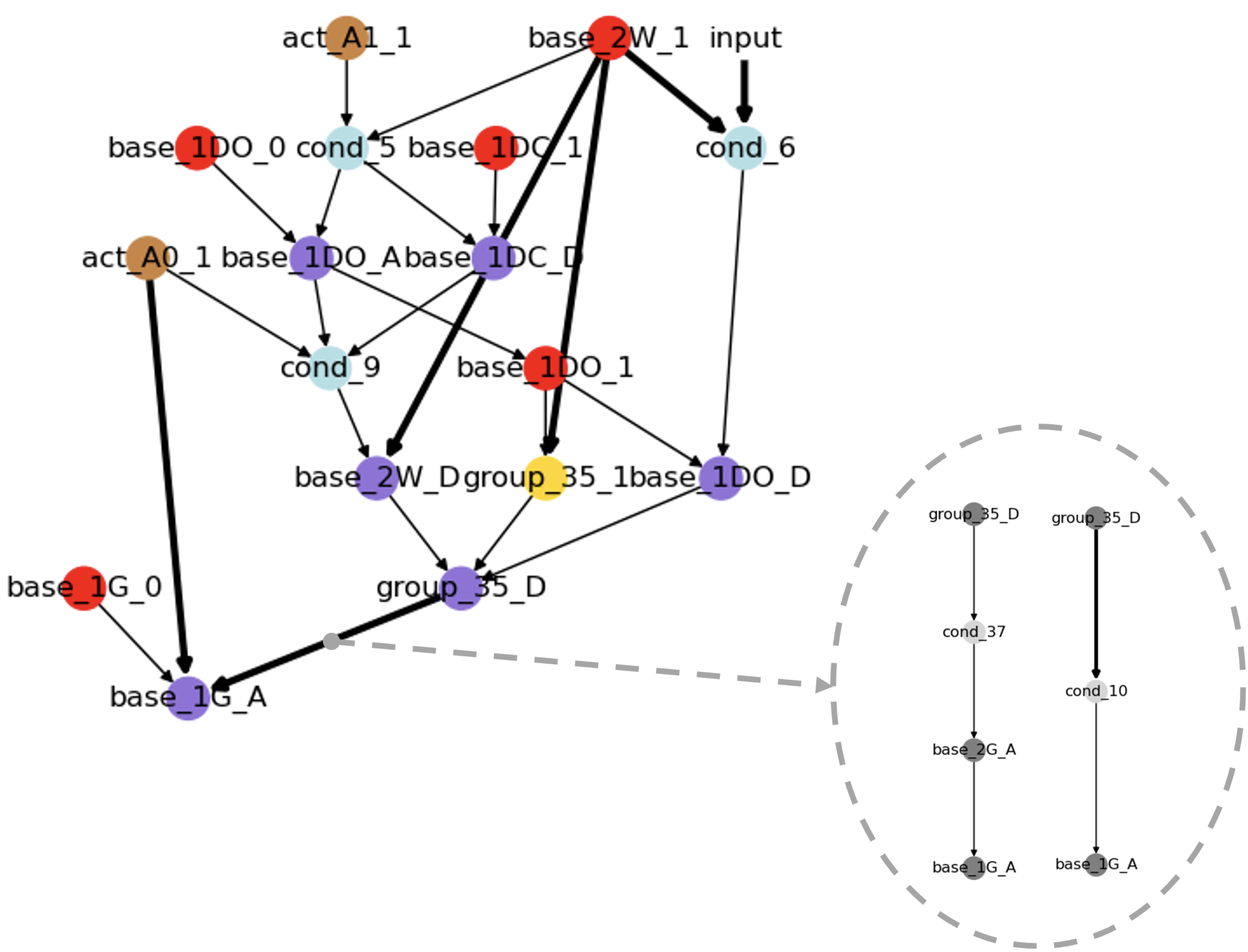}
         \caption{Encapsulated action network.}
         \label{fig:encapsulatedAN}
     \end{subfigure}
    \caption{Example of action networks on test environment. Bold edges are encapsulated. Each node represents a different state variable, and each edge represents conditioning and succession relations between them.}
    \label{fig:actionnets}
\end{figure}

\begin{table}[]
    \centering
    \caption{Base goal-directed behavior. Mean episode durations (across 4000 steps) before and after the introduction of goal (no goal/with goal), for Complete (nonrandom) and Complete-Random variants of the environment. For the latter, The Modeller's statistical significance filtering have been enabled. Actions are chosen randomly before the introduction of the goal. All results are averages across 5 independent trials. Inside paranthesis are standard deviations.}
    \begin{tabular}{c|c|c}
         & No goal & With goal  \\
        \hline
        Complete & 98.1 (17.69) & 7.28 (0.5) \\
        Complete-Random & 99.22 (32.61) & 22.33 (28.2) \\        
    \end{tabular}
    \label{tab:baseplan}
\end{table}

\begin{table*}[]
    \centering
    \caption{Continual learning. Mean episode durations across environment changes for vanilla, random environment, and readaptation variants, with purely random actions provided as a baseline. Columns represent the successive environment subtypes. Subtypes indexed "L" have model learning enabled, "NL" have it disabled (except for "readaptation" variant, which continues learning throughout the end). All results are averages across 5 trials, with standard deviations inside parantheses.}
    \begin{tabular}{c|c|c|c|c|c}
         & RS-L & SGS-L & NEG-L & RS-NL & SGS-NL  \\
        \hline
        Vanilla & 45.58 (25.55) & 5.33 (0.28) & 4.47 (0.22) & 10.38 (1.68)  & 4.3 (0.11)\\
        Random Env. & 190.86 (148.0) & 32.3 (9.93) & 9.87 (3.45) & 121.69 (82.33) & 35.05 (5.42) \\        
        Readaptation & 89.01 (58.72) & 28.19 (21.45) & 6.06 (0.74) & 13.73 (3.45) & 4.71 (0.15) \\       
        \hline
        Random actions & 275.86 & 67.53 & 52.48 & 275.86 & 67.53 \\        
    \end{tabular}
    \label{tab:continual}
\end{table*}

\subsubsection{Base Planning} Table \ref{tab:baseplan} compares episode durations for random actions and planning, the large drop demonstrating the effectiveness in accurately modeling the environment and performing goal-directed behavior on the model. The agent maintains stable performance across 4000 steps after goal introduction, demonstrating its ability to learn the environment independently of the goal and immediately achieve it without additional training—unlike RL-based approaches, which require retraining for changing goals. Randomness, however, does have a notable impact: while planning and modeling remain effective, the presence of additional connections above the significance threshold leads to more redundant action choices. This issue arise from relying only on first-order significance and the challenge of establishing a universal causal effect limit, a limitation is to be addressed in future work—see Appendix \ref{sec:statistical_significance} for details.

\subsubsection{Continual Learning} Table \ref{tab:continual} displays the agent's continual learning performance across changing environments, with the goal defined from the start. Both vanilla and random variants maintain or even improve their performance after exposure to different environments, often outperforming initial learning periods, without readaptation. For instance, the vanilla version averages 5.33 steps on the SGS variant during learning and 4.3 steps after intermittent exposure to other subtypes. Figure \ref{fig:CL_planning_nonrand_readapt} also illustrates this, showing that with model readaptation enabled, the agent performs consistently with its previous endpoint performance in the same environment subtype, without any spikes indicating destructive adaptation. Note that most steps are spent in the RS variant due to the precise timing requirements of the planner (as discussed in Section \ref{sec:planner}).

\subsubsection{Behavior encapsulation} Fig. \ref{fig:actionnets} shows a sample action network and a demonstration of the resulting encapsulated AN. Here the start states are (DC,W), hence encapsulation is between these states (and inactive states for all the rest) and the goal state. The full AN even for this simple environment is clearly very complex; however encapsulation can turn it into a comprehensible, structured, minimal format. On Fig. \ref{fig:encapsulatedAN}, many paths that are seen to be alternatives have been encapsulated (example shown from \textit{Group35-D} to \textit{1G-A}), and only reliable (i.e. necessary) connections remain; which, upon inspection, can be seen to correspond to the transition (DC,W) $\rightarrow$ (DO,W) that is invariably needed for reaching the goal from (DC,W). As discussed before, the identified subgoals and pathways, as well as encapsulated components, can be used as building-block subpolicies for future behavior, though we did not yet incorporate this integration with ongoing agent behavior.

\section{Experimental Validation 2: Modeller with Network Refinement for Vision}
\label{sec:experiments_mnr}

\subsection{Setup}

We experiment on the MNIST dataset to evaluate MNR's continual learning performance, compare its learning progression with neural networks, and analyze learnt representations for proper structure and comprehensibility. MNR's learning process involves randomly selecting $N_C$ classes from the 10 available at the start of each trial. In one \textit{cycle}, the system is exposed to $N_{sample}$ samples from each class sequentially, with one exposure and learning step per sample (no reexposure, as repeated steps have no effect in MNR). Processing samples from one class constitutes an \textit{iteration}. NNs follow the same flow, trained on one sample per step until convergence or maximum number of epochs. Performance is evaluated via per-class accuracy after each iteration ($N_C$ evaluations per cycle). We train for 10 such cycles, simulating a general and realistic continual flow of information with the requirement of ongoing learning without any constraining assumptions. Examples for model comprehensibility demonstrations are drawn from the same experiments. We run our experiments on MNR with $N_C=3,\ 5$ and $10$, and with $N_C=3$ on NNs for comparison of behavior. Reported results are averages of 10 and 5 runs for $N_C=3$ and $5$ respectively.  Further details, including choice of parameters and computation of predictions in MNR, can be found in Appendix \ref{sec:appendix_expsetup}. As discussed in Section \ref{sec:experiments_Modeller}, we do not include a comparison with existing continual learning methods in the main text, as we are unaware of any that offer a meaningful baseline—these methods operate under precisely the assumptions our framework is designed to eliminate. Nonetheless, as was mentioned in Sec. \ref{sec:experiments_Modeller}, Appendix \ref{sec:nn_cl_exps} provides such a comparison with neural network-based approaches, demonstrating that as their required conditions—such as access to replay data or clearly defined task boundaries—are removed, their performance rapidly deteriorates or collapses entirely. This further underscores the significance of The Modeller’s ability to learn continually without relying on such constraints.

\subsection{Results and Discussion}

\subsubsection{Continual Learning} 
\begin{figure*}
     \centering
     \begin{subfigure}{0.32\textwidth}
         \centering
         \includegraphics[width=\textwidth]{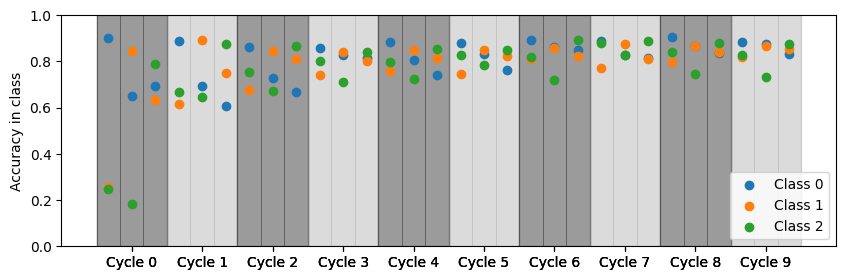}
         \caption{MNR, $N_C=3$.}
         \label{fig:results_cl_my}
     \end{subfigure}
     \begin{subfigure}{0.32\textwidth}
         \centering
         \includegraphics[width=\textwidth]{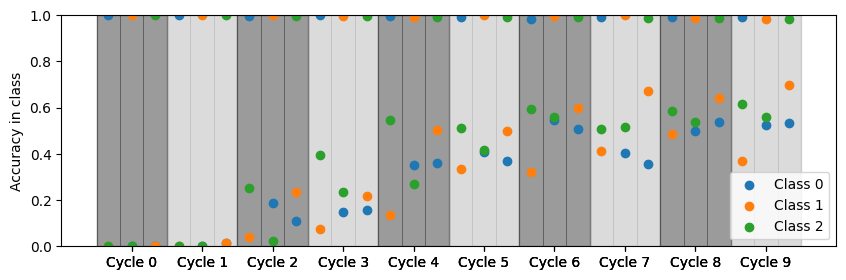}
         \caption{Fully connected neural network,$N_C=3$.}
         \label{fig:results_cl_nn}
     \end{subfigure}
     \begin{subfigure}{0.32\textwidth}
         \centering
         \includegraphics[width=\textwidth]{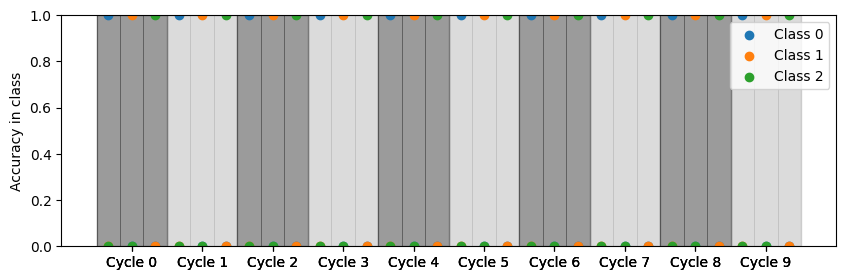}
         \caption{Convolutional neural network, $N_C=3$.}
         \label{fig:results_cl_cnn}
     \end{subfigure}
     \begin{subfigure}{0.35\textwidth}
         \centering
         \includegraphics[width=\textwidth]{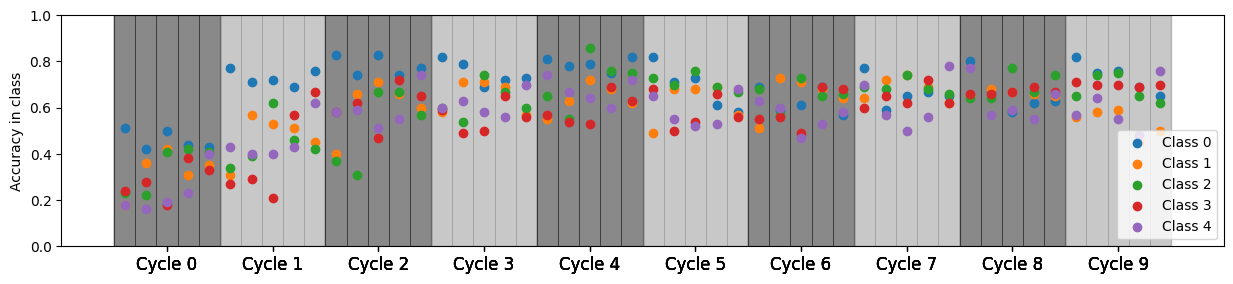}
         \caption{MNR, $N_C=5$.}
         \label{fig:results_cl_my_5c}
     \end{subfigure}
     \begin{subfigure}{0.35\textwidth}
         \centering
         \includegraphics[width=\textwidth]{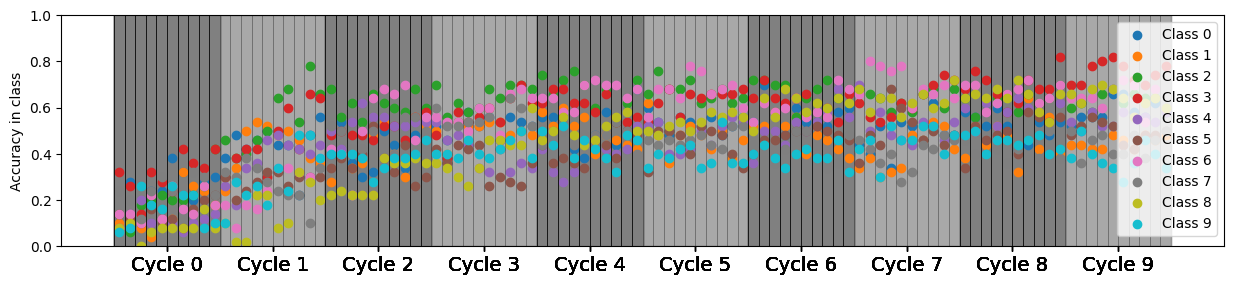}
         \caption{MNR, $N_C=10$.}
         \label{fig:results_cl_my_10c}
     \end{subfigure}
    \caption{Learning performance of MNR, fully connected, and convolutional neural networks on $N_C$-class incremental learning over 10 cycles. Accuracies reflect correct classification ratios for each class. Shaded areas denote cycles, and vertical lines separate iterations within cycles. Results are averaged over 10 (a-c) and 5 (d-e) runs. Note that class indices $i$ are randomly chosen at the start of each run and do \textit{not} necessarily correspond to digit $i$.}
    \label{fig:results_cl}
\end{figure*}

Figure \ref{fig:results_cl} shows learning progression of MNR for $N_C=3$, $4$, and $5$ classes; as well as that of the neural network variants for $N_C=3$ for comparison. MNR's final performance, as expected, does not yet achieve perfect identification, with accuracies of ~85\%, 60\% and 50\% for $N_C=3, 5$ and $10$ respectively after 10 cycles. This stems primarily from limitations of feature representation (Section \ref{sec:feature_representation}) and while it suggests the need for improvement with better representations (see Conclusions), it is not our main focus here. To validate continual learning of our design, we focus on MNR's high retention of learned information, as shown in Figures \ref{fig:results_cl_my}, \ref{fig:results_cl_my_5c}, and \ref{fig:results_cl_my_10c}. Performance on class $i$ remains stable in later iterations ($j>i$) of the same cycle, with early accuracy persisting throughout. This contrasts sharply with neural networks: a fully connected NN (Fig. \ref{fig:results_cl_nn}) loses all information on class $j>i$ in early cycles, and even after 10 cycles, it fails to retain a stable representation, showing $>50\%$ accuracy loss. A convolutional NN performs worse, losing all information repeatedly. Continual learning is most critical in early cycles (first 3-4), as in the long run, with increasing number of cycles, the problem is equivalent to stochastic gradient descent with a slow timescale, reducing the problem to statistical learning with data abundance where NNs already excel. MNR's retention of performance is consistent across tests with 5 and 10 classes as well, albeit with lower baseline accuracies. We note that in MNR, as in The Modeller, even when there are small performance fluctuations, it is by design not due to direct destruction of existing information (unless a conditioner is removed for cumulative insignificance) but stems from over-refinement or negative conditioning.

We briefly note the factors limiting performance compared to the perfect detection achieved by neural networks. First, the representation used lacks full expressivity, capturing only gradient change points rather than all shape features (see Section \ref{sec:feature_representation}). Second, the current statistical refinement approach retains some features not present in every sample of a class, yet SPN satisfaction (Definition \ref{def:spn_satisfied}) requires precise matches. This leads to missed instances, especially outliers. This is rather straightforward to offset by allowing soft satisfaction of SPNs. These limitations were not the focus of this work, which prioritized validating the learning flow with a demonstrative representation, and addressing them are left for future research.

\subsubsection{Comprehensibility of Learned Representations} 
\begin{figure}
     \centering
     \begin{subfigure}{0.15\textwidth}
         \centering
         \includegraphics[width=\textwidth]{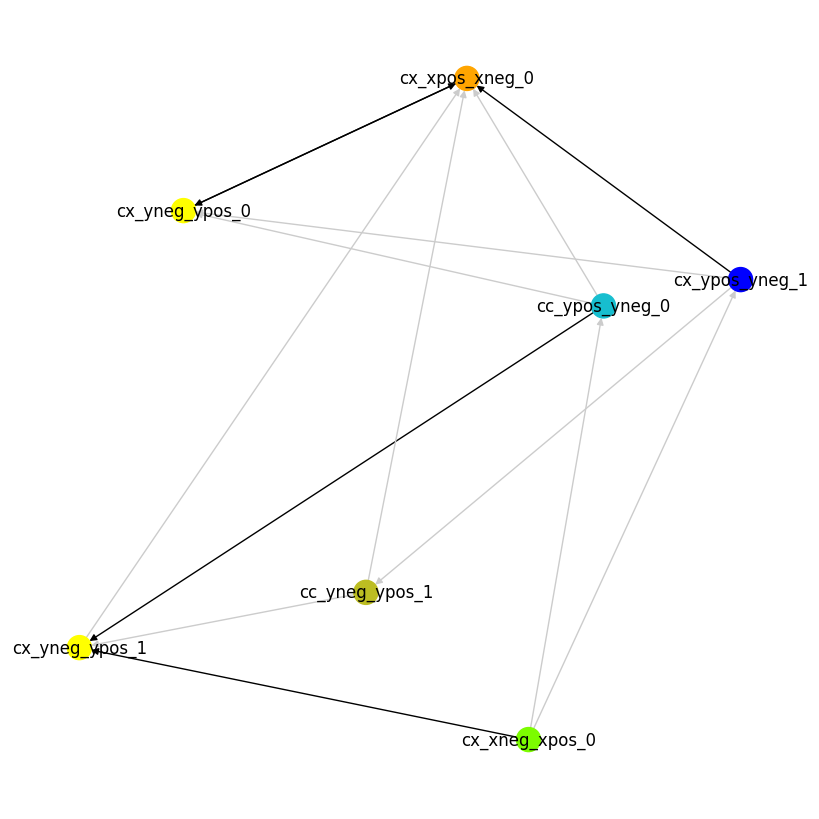}
         \caption{Digit 2, depth 0.}
         \label{fig:results_spns_2_ds}
     \end{subfigure}
      \begin{subfigure}{0.15\textwidth}
         \centering
         \includegraphics[width=\textwidth]{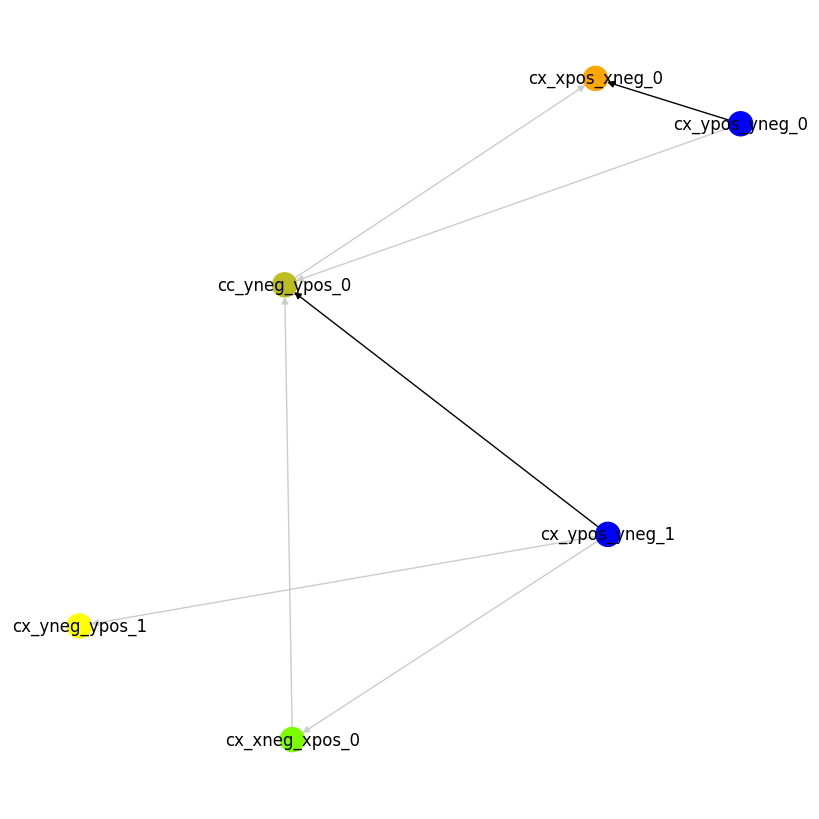}
         \caption{Digit 5, depth 0.}
         \label{fig:results_spns_5_ds}
     \end{subfigure}
     \begin{subfigure}{0.15\textwidth}
         \centering
         \includegraphics[width=\textwidth]{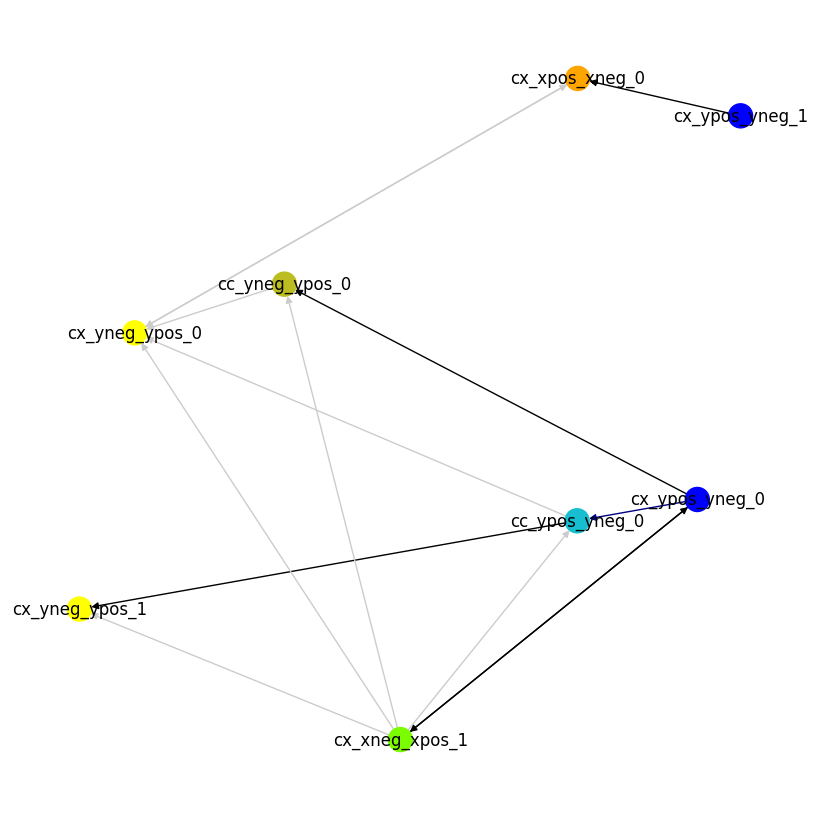}
         \caption{Digit 5, depth 8.}
         \label{fig:results_spns_5_us9}
     \end{subfigure}
    \caption{Sample source SPNs learned by the system, with all SNs combined. Black \& gray edges represent contour \& all-type connections respectively. Depths indicate conditioning distance from the target variable, deeper being more specific. Node positions are the average of all observed positions during the CSV's lifetime. Additional examples are in Fig. \ref{fig:results_spns_appendix} in A.2.}
    \label{fig:results_spns}
\end{figure}

Fig. \ref{fig:results_spns} illustrates samples of learned SPNs, ranging from general representations at lower depths (near the target variable) to more specific ones at higher depths capturing rarer subvariants. These representations are visually intuitive, effectively depicting the digits, features, and interrelations. For example, most contours of digit "2" are preserved in Fig. \ref{fig:results_spns_2_ds}, though features like holes at the lower-left turn (common in some samples, e.g. Fig. \ref{fig:feature_representation}) are omitted, while persistent features like the vertical gradient changes on the left side (\textit{"cx\_yneg\_ypos\_0/1"}) are retained. Similarly, digit "5" in Fig. \ref{fig:results_spns_5_ds} retains key features, including vertical gradient changes on the right and horizontal changes at the top \& bottom, along with correct positional relations. While some contours of "5" are refined at depth 0, they are preserved at the more specific subvariants upstream that provide more details (Fig. \ref{fig:results_spns_5_us9}). Some additional examples are also provided in Fig. \ref{fig:results_spns_appendix} in Appendix.

\section{Conclusion}
\label{sec:conclusion}

The system described in this paper, comprising The Modeller, the planner, and the behavior encapsulator, provides the foundations of a novel solution to core challenges in contemporary ML. This document highlighted its strengths in continual learning, interpretability, seamless integration with planning, and flexible behavior hierarchies. Its effectiveness in these areas arises from a shared foundation: constructing a structured environment model while retaining past knowledge using fundamental principles from evolutionary developmental biology. We also extended this approach to higher-dimensional observation spaces, specifically to vision for 2D shape identification, via network refinement with rerelation. While accuracy-wise performance doesn't yet match NNs due to limitations in feature expressivity, our core design aims were validated by (1) continual learning without constraining assumptions and consistent retention of past knowledge, and (2) producing human-comprehensible representations. We believe both the presented design and the general principles behind our approach offer considerable potential to address the fundamental limitations of existing ML approaches organically.\footnote{One might view the current design's complexity as being oriented toward addressing \textit{fundamental} challenges in learning mechanisms, rather than task complexity per se. While neural networks are capable of solving more complex tasks than those presented in this work, they are unable to perform even simpler tasks (as shown in Section \ref{sec:experiments_mnr}) with the set of properties targeted here—namely, continual learning, comprehensibility, and integration with symbolic methods—without substantial extensions. These extensions, in turn, introduce significant constraints of their own, and fail when they are relaxed, as shown in Appendices \ref{sec:nn_plan_exps} and \ref{sec:nn_cl_exps}. In contrast, while our current approach requires further development to match the task complexity achievable by neural networks, it already delivers the aforementioned properties within its operational domain, doing so without the need for the extensive elaborations required to adapt neural networks to the same ends; with the aim to be extended towards the task complexities that can currently be realized via NNs with further developments.}

The only \textit{inherent} limitation of our general approach is its reliance on discretely-represented spaces (as a result of the principle of \textit{weak linkage}). Handling continuous spaces would require additional preprocessing stages such as digitization. However, at the same time, most problems in AI can be represented or converted to discrete formats, main exceptions being tasks needing fine-tuned control, in which case a Modeller-based system can complement statistical learning for low-level control. Thus, explicit support for continuous spaces may not be necessary, as our approach is designed for cognitive tasks in structured environments rather than low-level control.

\textit{\textbf{Venues for future work}}

Our design is an early proof of concept for a novel approach to artificial learning, showcasing its core mechanisms and addressing multiple major limitations of ML in a unified manner. As such, it naturally opens numerous avenues for future developments, including those for refining the framework, improving efficiency and precision to match methods like NNs, and integrating system components to tackle key ML limitations in relevant domains.

The first one involves addressing the Markovian assumption in our environment modeling approach, modeling only immediate temporal successions. Extending The Modeller to non-Markovian environments requires modeling long-term temporal dependencies. Network refinement with rerelation (MNR), currently used for vision, can handle this by representing temporal events as a unidirectional network, with rerelation capturing non-immediate dependencies through intermediaries. Additionally, to scale our planner for complex environments, selective pathway extension is necessary. This can be achieved using The Modeller's mechanisms, such as returning when a viable path is found or prioritizing based on statistical significance. Likewise, precise timing can be managed by evaluating pathway consequences and excluding those that reverse preconditions or hinder future actions.

As discussed in the main text, the current preprocessing method for visual spaces to construct observation SPNs has limitations. The feature representation, based on gradient sign changes, captures key features but is expressively limited, as shown by performance gaps. Future work should enhance expressivity, possibly by refining this edge-gradient method, which theoretically contains all necessary shape information. Alternatively, computer vision methods like SIFT \cite{lindeberg2012scale} or pretrained models (\cite{oquab2023dinov2}) could be used for feature representation, while retaining MNR for high-level learning. Frequency components, pixel-level detection (if scalability challenges are addressed), and intermediate representations, such as fixed-size filters are also promising directions. Such extensions would not only improve performance but also extend applicability beyond 2D shape identification. A related natural long-term extension of this work is processing more complex visual spaces, which requires hierarchical representations. Simple structures (e.g., eyes or noses) can be learned with the current approach, but more complex entities (e.g., faces or multi-object scenes) need hierarchical organization. MNR is well-suited for such spaces due to their fewer alternative assignments. Future work will explore such extensions into hierarchical representations and their unsupervised construction through learning progression.

Another future direction is integrating the foundational components presented in this paper into unified frameworks, like we currently do with modelling and planning, tailored to specific applications. In particular, incorporating our visual space learning approach (MNR) can extend environment-dynamics modeling and planning capabilities to high-dimensional visual environments, where dynamics are represented by changes in network representations and conditions by visual constructs. Additionally, behavior encapsulation, currently separate from the agent's flow, can be integrated into ongoing operations to enable reusable behavior patterns, a key motivation shared with hierarchical RL. We believe the structured representations learned by our system provide an ideal foundation for this.

Finally, the current implementation is computationally demanding due to redundancies included for simplicity, as optimization was not the focus. This is primarily because the learning flow processes multiple intersecting upstream paths simultaneously. Streamlining this by handling only the "best-matching" path at each step would reduce the upstream network processed. While the model in The Modeller can grow as large as needed, per-step operations should ideally involve a small set of CSVs, with their sources (or SPNs) no larger than the active observations. This would cap per-step complexity to the observation space size, and our goal is to achieve this computational efficiency. A more detailed discussion of this aspect is provided in Appendix \ref{sec:appendix_development_viability}.


This novel design approach clearly requires extensive refinement, optimization, and integration into task-specific workflows to achieve widespread adoption like neural networks—who themselves went through a similar evolution—while retaining the unique capabilities demonstrated here. Our work provides a concrete direction for these developments. Supported by theoretical foundations and promising results, this approach addresses key limitations of neural networks and could drive the next major leap in AI design.

\input{main.bbl}


\begin{IEEEbiography}[{\includegraphics[width=1in,height=1.25in,clip,keepaspectratio]{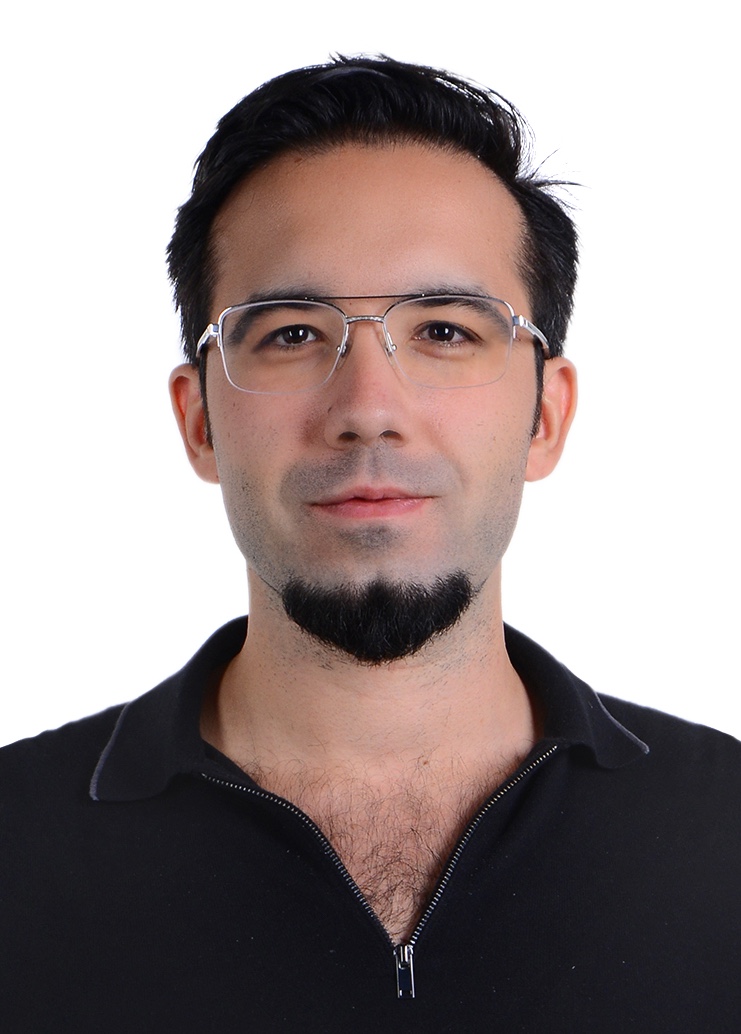}}]{Zeki Doruk Erden} is a researcher whose current work focuses on Artificial Intelligence and Evolutionary Theory, particularly on artificial learning paradigms that enable adaptive growth, structured representations, and recursive improvability. He is currently with Sabancı University. He received his Bachelor’s degree from Middle East Technical University (METU) and his Master’s and Ph.D. from École Polytechnique Fédérale de Lausanne (EPFL).
\end{IEEEbiography}

\begin{IEEEbiography}[{\includegraphics[width=1in,height=1.25in,clip,keepaspectratio]{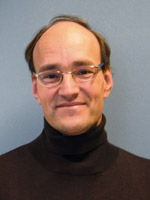}}]{Boi Faltings} was a full professor of computer science at the Ecole Polytechnique Fédérale de Lausanne (EPFL) for 38 years,  where he headed the Artificial Intelligence Laboratory, and has held visiting positions at NEC Research Institute, Stanford University and the Hong Kong University of Science and Technology. He has co-founded 6 companies using AI for e-commerce and computer security and acted as advisor to several other companies. Prof. Faltings has published over 350 refereed papers and graduated 43 Ph.D. students, several of which have won national and international awards. He is a fellow of the European Coordinating Committee for Artificial Intelligence and a fellow of the Association for Advancement of Artificial Intelligence (AAAI). He holds a Diploma from ETH Zurich and a Ph.D. from the University of Illinois at Urbana-Champaign.
\end{IEEEbiography}


\newpage

\section*{Appendix}

\subsection{Conceptual Foundations: Analogy of Evolutionary Theory \& AI, and Evolutionary Developmental Biology}
\label{sec:app_evol_ai}

Evolutionary theory is typically regarded as only tangentially linked to AI. However, a closer look reveals its primary relevance, as this field has traversed an intellectual trajectory remarkably similar to what AI is currently navigating.

Modern Synthesis (MS)—the dominant 20\textsuperscript{th} century view of evolution, combining Darwinian selection with Mendelian inheritance and population genetics \cite{gilbert2000developmental, mayer2013evolution}—despite greatly advancing our understanding of evolution, had left key explanatory gaps. Firstly, MS does not address the generation of biological structures exhibiting modularity, hierarchy, or repetition, instead abstracting structural properties within genetic variation \cite{gilbert2000developmental, marc2005plausibility, carroll2005endless, laland2015extended}. Second, while MS frames evolution as a slow, gradual optimization process at the population level, the history of life reveals periods of rapid transformation 
\cite{gould1977punctuated, marc2005plausibility} and an accelerating pace of phenotypical change and complexification over time \cite{smith1997major, kurzweil2006singularity, gerhart2007theory}.\footnote{An analogy can be drawn between Modern Synthesis and Genetic Algorithms: These algorithms optimize predefined structures (e.g., selecting variables or tuning parameters \cite{bayley2008design}) but cannot spontaneously generate new structures or enable rapid, large-scale adaptation—except for a subset known as \textit{indirect encodings} inspired by developmental processes \cite{meli2021study}.} Recent advances in biology have extended MS by addressing gaps in its statistical optimization framework \cite{laland2015extended}. Among these, evolutionary developmental biology (EDB) has been especially influential \cite{carroll2005endless, marc2005plausibility, west2003developmental, gerhart2007theory}, revealing how shared developmental principles drive structural properties of organisms and a recursively accelerating increase in phenotypic complexity and capabilities.

The Modern Synthesis of evolution and contemporary machine learning share striking parallels \cite{erden2025parallels, erden2025evolutionary}. Both view their respective processes of adaptation fundamentally as statistical optimization processes. They emphasize an averaged final performance and treat their subjects as largely unstructured entities, encapsulating relevant properties in abstract continuous variables (gene frequencies and edge weights). Both are considered incomplete for similar reasons: Modern Synthesis does not fully explain certain aspects of evolutionary progress, namely the speed of adaptation, the inherent adaptability of organisms, or the widely-observed structural properties of biological systems. Analogously, contemporary ML faces criticism for its data inefficiency ($\sim$speed of adaptation), inability to leverage past knowledge when learning new tasks without destroying existing information ($\sim$inherent adaptability), and opaque internal structure ($\sim$structural properties). This conceptual near-equivalence prompts a crucial question: Considering how developmental principles enhance evolutionary theory and address the explanatory gaps in the MS, how can they also underpin the resolution of the corresponding limitations in AI systems, paving the way for a new type of learning systems? We believe two universal principles from evolutionary developmental biology to be particularly relevant to us in this context:

\textit{Adaptation via local variation and selection:} A particular category of developmental processes leverage the \textit{variation and selection} principles underlying biological evolution \textit{locally}, where variants of a substructure are generated and then refined through a selective signal, forming the primary means for an organism to adapt to needs of the environment where no evolutionary preprogramming is possible \cite{marc2005plausibility, west2003developmental, gerhart2007theory}. This is exemplified in the operation of the immune system \cite{rajewsky1996clonal} and in neural development \cite{hiesinger2021self, petanjek2023dendritic}. Importantly, overparameterized neural networks trained through gradient descent can also be seen as a process of variation and selection: An initially abundant pool of variation (randomly initialized weight patterns) is subjected to a selective force (gradient descent), amplifying "beneficial" weights that reduce error while diminishing others. Crucially, \textit{this variation-selection cycle is neither local nor iterative}, but is a singular process affecting the complete network: there is no mechanism for locally regenerating variation atop learned structures without disrupting existing knowledge, in contrast to the biological processes that can locally generate variation on demand, shedding a new light on the continual learning problem. At the heart of the learning mechanism introduced below lies a process of local variation and selection which not only enables structural adaptation in the model, but also underpins the system’s continual learning capability, as demonstrated in Theorem 1.

\textit{Regulation of weakly-linked core processes:} A large number of fundamental biological processes are highly conserved across distantly related organisms \cite{marc2005plausibility, gerhart2007theory}, including processes responsible for generating basic \cite{holstein2012evolution} as well as higher-order structures \cite{lemons2006genomic}. Such \textit{core processes} evolved early in evolutionary history and have been mostly preserved since then. Major evolutionary changes since these early developments have primarily occurred at their \textit{regulation}, which control the expression and interaction of these core processes \cite{tickle2017sonic, wang2016divergence, suresh2023comparative}. These regulatory modifications not only drive rapid evolvability, but the regulatory networks formed via these pathways can themselves become conserved and, via being combined or reconfigured to create more complex functions (e.g. vertebrate eye \cite{lamb2007evolution, lamb2008origin}), serve as substrate for further evolution, hence recursively improving future evolutionary potential. Underlying this versatility is the connection of core processes via a \textit{weak linkage}, where regulatory signals (inputs/conditions) for their activation are very simple, as opposed to complex and precise ones \cite{gerhart2007theory, marc2005plausibility}, greatly facilitating regulatory changes or recombinations. Hence, small changes in regulation, when unfolded across development, can lead to dramatic alterations in the adult organism \cite{halder1995induction}. The structural properties observed in organisms (modularity, hierarchy, repetition, correlated changes), which in turn facilitate evolution \cite{carroll2005endless, clune2013evolutionary, mengistu2016evolutionary, szilagyi2020phenotypes, carroll1995homeotic}, stem from the architecture of these regulatory networks, such as reuse of regulatory mechanisms across various processes, distinct gene sets regulating different parts of the organism, or higher-level regulatory genes controlling the expression of downstream genes \cite{carroll2005endless, marc2005plausibility, gerhart2007theory}. The representation introduced in Section \ref{sec:Modeller} is built upon \textit{weakly linked components} and the potential for the \textit{regulation (termed "conditioning") of learned relationships}, enabling on-demand complexification and convenient decomposability, as will be detailed further below.

As discussed in Section \ref{sec:background}, the potential of structural organization in learned models to mitigate destructive adaptation, enhance comprehensibility \& engineerability, and enable integration with established "symbolic" algorithms like planning is well recognized in the literature. However, despite the intuitive appeal of these principles, advancements towards this goal remain surprisingly limited. The key insight from evolutionary developmental biology in this regard—and the critical gap in existing approaches to incorporating structure in ML systems—is the integration of the \textit{developmental principles} discussed above. These principles govern the structure of biological organisms and recursively-increasing pace of evolution. Without developmental principles—such as learning through local variation-selection instead of a static network, or weak linkage at lower levels of organization instead of densely interconnected, fine-tuned units—operating at the foundational level to generate structure rather than imposing it from above, it becomes impossible to achieve the desired properties \textit{adaptively} and \textit{across all levels of organization}. Complementarily, the complexity of core control processes in biological systems indicates that the appropriate design level for adaptive systems is \textit{not} high-level integration, but \textit{more proficient low-level organizational units}—standing in contrast to the majority of current ML research, which focuses primarily on high-level mechanisms operating on networks of artificial neurons \cite{wan2024towards, colelough2025neuro} or building blocks with qualitatively similar capabilities \cite{bal2024rethinking}.

These conceptual foundations form the basis of the design outlined in this paper.

\subsection{Local preservation of past responses}
\label{sec:app_lppr}

Continual learning with preservation of existing knowledge is central to our design. We first examine the conditions necessary for this and their limitations in contemporary methods. Since system-wide retention is hard to analyze and likely infeasible without replay, we focus on \textit{local preservation of past responses} at the computational unit level—a practical proxy and likely prerequisite for continual learning.

We are interested in computational units whose response can be phrased as $y=f(\sum w_i x_i + b)$ where $w_i$ are the weights, $x_i$ are the inputs and $b$ is the bias of the unit, and $f(\cdot)$ is a nonlinear function. Artificial neurons are examples of such units (as well as CSVs in our design, see next section). Assume that the unit, in the past, has been exposed to an input instance $\textbf{X}_t = [x_{0t}, ... x_{Nt}]$ at some step $t$. With the standing weights $\textbf{W} = [w_0, ... w_N]$, its response to this input given as $y_t=f(\sum w_i \cdot x_{it} + b)$. Suppose that after an arbitrary process of learning in response to other instances, weights of the unit are updated to $\textbf{W}' = [w'_0, ... w'_N]$, $w'_i = w_i + \Delta w_i$ and bias as $b' = b_i + \Delta b$ at step $t=T$. New response to past input $\textbf{X}_t$ is now $y'_t = f(\sum w'_i x_{it} + b')$. We want to ensure $y'_t=y$, or  $f(\sum w'_i x_{it} + b') = f(\sum w_i x_{it} + b)$. This is generally valid if (and for monotonous $f(\cdot)$ as frequently used in NNs, only if) $ \sum (w_i + \Delta w_i) x_{it} + (b + \Delta b) = \sum w_i x_{it} + b$, reducing to:

\begin{equation}
    \label{eq:popr_condition}
    \sum \Delta w_i x_{it} + \Delta b = 0
\end{equation}

Eq. \ref{eq:popr_condition} is the condition for local preservation of information. A learning update rule defining $\Delta w_i$ and $\Delta b$ should satisfy this condition if response of the unit to previous inputs $\textbf{X}_t$, $t \leq T$ is to be preserved. Some of the terms in the sum of Eq. \ref{eq:popr_condition} may be $0$ due to $\Delta w_i$ being 0, as defined by the learning rule, based on the current value of $w_i$. So we can rewrite Eq. \ref{eq:popr_condition} as $\sum_{i:\Delta w_i \neq 0} \Delta w_i x_{it} + \Delta b = 0$. Note that this equation is in general dependent on $x_{it}$ of $t \leq T$ for $i:\Delta w_i \neq 0$, which we assume to be not explicitly available. The only possibility for an update rule to satisfy the condition is to make it independent of $x_{it}$. This would only be possible if $\forall i:\Delta w_i \neq 0$, and $\forall t \leq T$, $x_{it}$ can be readily deduced from the available information, in which case $w_i$. This can be written as:

\begin{equation}
    \label{eq:x_g_rel}
     \forall i:\Delta w_i \neq 0,\ \forall t \leq T,\  w_i = g(x_{it})\rightarrow x_{it} = g^{-1}(w_i)
\end{equation}

So that Eq. \ref{eq:popr_condition} becomes independent of $x_{it}$: $\sum_{i:\Delta w_i \neq 0} \Delta w_i g^{-1}(w_i) + \Delta b = 0$. Notice that it follows from Eq. \ref{eq:x_g_rel} that:
   
\begin{equation}
    \label{eq:xs_rel}
    x_{i0} = x_{i1} = ... = x_{iT},\ \forall t \leq T,  \forall i:\Delta w_i \neq 0\
\end{equation}

Neural networks trained via gradient descent do not satisfy these conditions because (1) learned weights lack direct correspondence to the inputs they modulate (Eq. \ref{eq:x_g_rel}), and (2) past inputs are not guaranteed to be identical for a given weight (Eq. \ref{eq:xs_rel}). In contrast, our design, The Modeller, described in the main text, satisfies these conditions, by the following equivalences by CSVs as computational units:
\begin{enumerate}
    \item Sources to which the CSV is no longer connected to are the instances that took different values in the past (equivalent to having $w_i=0$), and they are known to have $\Delta w = 0$ since a removed source remains disconnected.
    \item For all the remaining connections ($\Delta w$ not necessarily $= 0$, i.e. can undergo change) we know that $x_{it}=1,\ \forall t \leq T$ since the ongoing presence of an input $i$ means that it has not been observed absent in the past.
    \item The observation condition of a CSV (same as the condition for the satisfaction of its sources) is equivalent to $f(.)$ being a step function, $w_i$ being 1 for all connections, and $b = -|\textbf{W}_c|$, $\textbf{W}_c$ being the set of all connected sources.
\end{enumerate}

Hence, The Modeller locally preserves past responses by this analysis, providing a different view to Theorem 1 in the main text.

\subsection{Details of The Modeller system components}
\label{sec:Modeller_details}

We define a state variable (SV) as a variable that can take three values: 1 for \textit{active}, -1 for \textit{inactive}, and 0 which can be interpreted as \textit{unobserved, undefined,} or \textit{irrelevant} depending on context. Note that the numerical values are given only as shorthand notation and do not participate in an algebraic operation anywhere. The phrase \textit{nonactive} refers to any SV that is not active. The SV construct comes in three subtypes: Base SVs (BSVs), Dynamics SVs (DSVs), Conditioning SVs (CSVs).

\textit{BSV:} BSVs are the externally-specified SVs whose states, which is assumed to be either 1 or -1, are provided externally to the system at each time instant. These can be regarded as the direct observations from the environment.

\textit{DSV:} Each BSV comes with two associated DSVs, for activation (A-DSV) and deactivation (D-DSV) respectively. Activation at timestep $t$ is defined as the transition of a BSV state from -1 in step $t-1$ to 1 in step $t$; and likewise deactivation at $t$ is defined from 1 in $t-1$ to -1 in $t$. At step $t$, A-DSV is deduced active (state 1) if activation is observed at step $t$, inactive (-1) if a BSV is inactive at $t-1$ and no activation is observed at $t$, and undefined (0) if the BSV is already active. Symmetrically, at step $t$, D-DSV is deduced active (state 1) if deactivation is observed at step $t$, inactive (-1) if a BSV is active at $t-1$ and no deactivation is observed at $t$, and undefined (0) if the BSV is already inactive. The BSVs are modelled only through changes in their states via their associated DSVs, and are not predicted by themselves.

\textit{CSV:} A CSV is a SV that conditions either DSVs or other CSVs (but not BSVs since they are not subject to direct modelling of their states); that is, predicts their activation. More specifically; each CSV comes with a set of positive and negative sources, where each source is either a BSV or DSV; and a set of targets, which correspond to the SVs that this CSV conditions. At steady state, a CSV’s source conditions are said to be satisfied when all its positive sources were active and all its negative sources were nonactive in the previous step - in other words, the satisfaction corresponds to the condition $all(positive\ sources)\ and\ not(any(negative\ source))$ in the previous step. A CSV state is undefined (0) if its source conditions are not satisfied. If its source conditions are satisfied; a CSV’s state is active (1) if the state of all its targets are either active or unobserved; and inactive (-1) if the state of all its targets are either inactive or unobserved. In case inactive and active targets are observed together, the CSV is duplicated to encompass the corresponding subsets of targets (as detailed below), hence we always ensure that one of the two above conditions will be satisfied with respect to the states of the targets. A CSV is to be interpreted as a state variable that represents the observance of a particular relationship - it being active means that this particular relationship (e.g. a change, as represented by a DSV, is observed conditioned on some sources) is observed, and it being inactive means that this relationship is not observed. The CSV being undefined or unobserved corresponds to the case in which the conditions for the observation of the relationship are not satisfied in the first place.

Potential targets of conditioning (i.e. DSVs and CSVs), when they are not undefined, are expected to be active if one of their conditioners are active; and inactive otherwise. Furthermore, these types of SVs also possess an \textit{unconditionally} flag, that allow for exceptions in this activity prediction, and are used to model uncertainty regarding activation of SVs. This flag can take three values: It starts with a value "unconditional" at the creation of the CSV and, if the CSV is observed to always be active whenever its sources were satisfied, it remains so. At the first observation of a case where the sources of the CSV are satisfied without the CSV being active, this flag changes to "conditional," signalling that sources alone do not suffice for the activation of the CSV and activity of one of its upstream conditioners is expected. The "conditional" value persists until the first observation of a case where CSV is observed active without any upstream conditioner being active and no new conditioner could be formed (see below and the main text); in which case the flag changes to "possibly unconditional" and remains as such.

Over the course of interaction with the environment, The Modeller learns a model that predicts the BSV states at the next step indirectly via the prediction of the DSV states. Within the predictions uncertainty is also represented where needed, as apparent from the description of the SVs. Since uncertainty is represented in a local basis (by unconditionality flags of individual SVs), and since CSVs are points of connection relating potentially multiple sources to potentially multiple targets; the uncertainty representation can represent alternative correlated outcomes in a tree-like manner where each downstream “branch” corresponding to the alternative outcomes in one direction or another can include multiple outcomes that occur together - we note that representation of uncertainty as such is not possible in a local manner with e.g. classical neural networks.

\begin{figure}[t]{}
     \centering
     \includegraphics[width=0.2\textwidth]{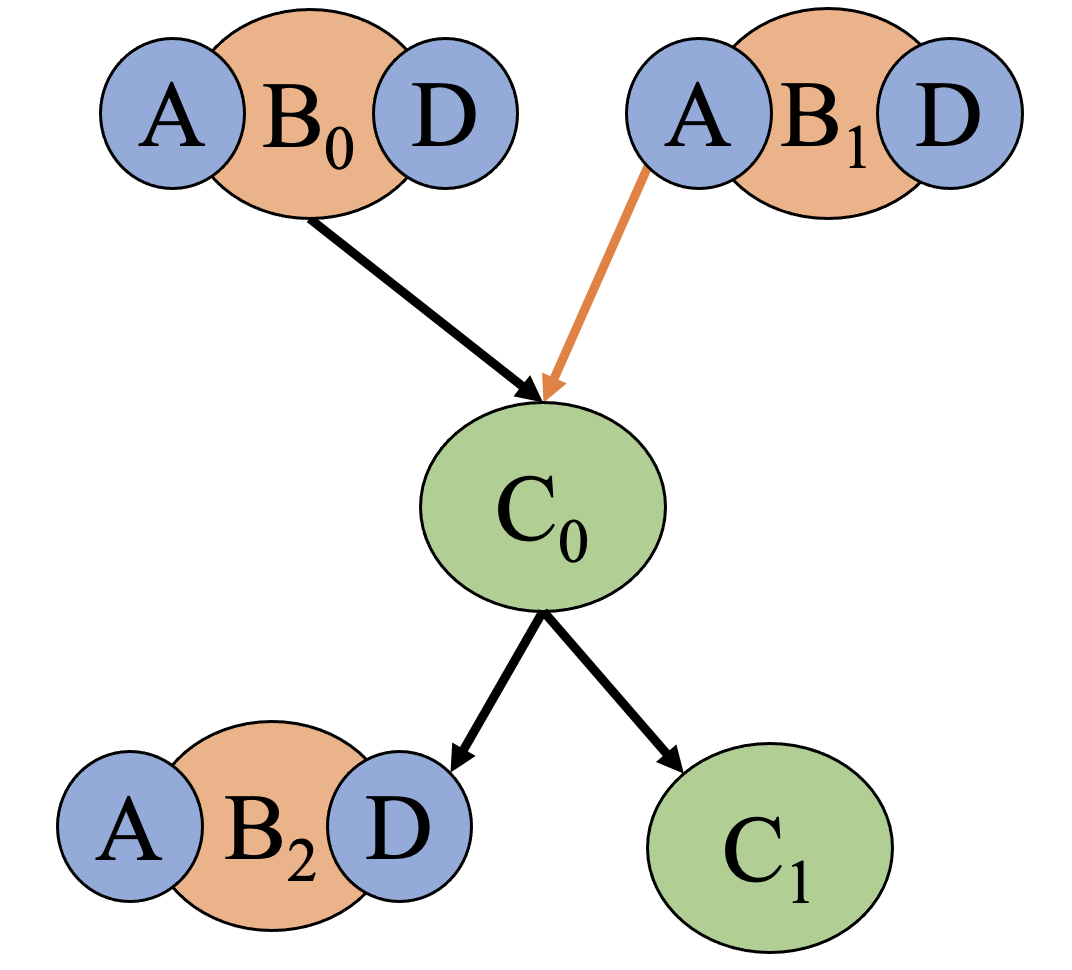}
     \caption{Illustration of SV types and relationships. The figure shows BSVs ($B_i$), their DSVs for activation (A) and deactivation (D), and CSVs ($C_i$). Here, CSV $C_0$ takes as positive source BSV $B_0$, as negative source the activation DSV of $B_1$; and conditions the CSV $C_1$ as well as the deactivation of $B_2$, modelling "$B_2$ is deactivated and $C_1$ is active if $B_0$ is active and $B_1$ is not activated."}
     \label{fig:svs}
 \end{figure}

\subsection{Learning the model}
\label{sec:Modeller_details_learning}

First, we provide an overview of the learning process in one step of interaction with the environment. During a step, the model is traversed, and the states of all its SVs are computed. For CSVs sources and targets are modified to be able to match the current states to the predictions/explanations of the CSV, so that the model is consistent with the environment at each step. After that, new CSVs are generated for the DSVs and CSVs that lack an explanation at the current step. The new CSV takes as positive sources all currently active eligible SVs in an exhaustive manner. Finally, model is refined by removal of unnecessary state variables.

The learning process is summarized formally on Algorithms \ref{alg:algorithm_adaptationloop} and \ref{alg:algorithm_csvstate}. Below, we provide a detailed breakdown of the processes described on those algorithms.

\begin{algorithm*}[tb]
\caption{Pseudocode of the main The Modeller adaptation loop; formed of state computations followed by CSV generation for unexplained SVs.}

\label{alg:algorithm_adaptationloop}
\textbf{Parameter}: $N$ Set of all target nodes\\
\textbf{Function} \textit{ProcessEnvironmentStep}(observations)
\begin{algorithmic}[1] 
    \STATE $BSVStates \leftarrow \ observations$
    \STATE $ComputeDSVStates()$ \textit{Comment: Computes DSV states by BSV events}
    \FOR{$level\ \in\ reverse(ComputationLevels)$}
        \FOR{$CSV\ \in\ SVs_in(level)$}
            \STATE $ComputeState(CSV)$
        \ENDFOR
    \ENDFOR
    \STATE $UnexplainedSVs \leftarrow [SV:\ SV.state = 1\ \AND\ NoConditionerActive(SV)]$
    \STATE $sources \leftarrow [SV:\ SV in\ [BSVs, DSVs]\ \AND\ SV.state = 1\ \AND\ isEligible(SV)]$
    \STATE $ NewCSV = CreateCSV(sources, [SV: SV\ in\ UnexplainedSVs\ \AND\ TargetEligible(SV)])$
    \STATE $ModelRefinement()$ \textit{Comment: Removes CSVs with no source or target}
\end{algorithmic}
\end{algorithm*}

\begin{algorithm*}[tb]
\caption{Pseudocode for CSV state computation.}

\label{alg:algorithm_csvstate}

\textbf{Function} \textit{ComputeState}($CSV$)
\begin{algorithmic}[1] 
    \IF{$AnySourceActive()$}
        \STATE $SeparateActiveInactiveTargets()$ \textit{Comment: Creates two CSVs from current one with active and inactive targets in either of them}
        \IF{$AnyTargetObserved()$}
            \STATE $State = 1$
            \STATE $PosSources \leftarrow [source:\ source\ in\ PosSources\ \AND\ source.state=1]$
            \STATE $NegSources \leftarrow [source:\ source\ in\ NegSources\ \AND\ source.state!=1]$
        \ELSIF{$AnyTargetInactive()$}
            \IF{$not(AllSourcesActive())$}
                \STATE $State = 1$
            \ELSE
                \IF{$AnyNegativeSourceActive()$}
                    \STATE $State = 0$
                    \STATE $NegSources \leftarrow [source:\ source\ in\ NegSources\ \AND\ source.State=1]$
                \ELSE
                    \STATE $State = -1$  \textit{Comment: No negative source active to explain inactivity of targets}
                \ENDIF
            \ENDIF
        \ENDIF
    \ELSE
        \STATE $State = 0$ \textit{Comment: Unobserved if targets are not observed}
    \ENDIF

    \IF{$State = -1$}
        \IF{$NegativeConnectionsFormed$}
            \STATE $FormNegativeConnections()$
        \ELSE
            \STATE $unconditionality = "isConditional"$
        \ENDIF
    \ENDIF

\end{algorithmic}

\end{algorithm*}

Initially, the model is generated with only BSVs and their associated DSVs, and without any CSV. At every step, the current and previous states of all the SVs are recorded, as well as the current and previous events (activation and deactivation) of every BSV. 

At each step, the effective network created by DSVs and CSVs are traversed in the reverse order of computation, similar to backpropagation algorithm; starting from DSVs, then the CSVs that condition these BSVs, then the conditioners of these CSVs, and so on. Each traversed SV gets their state computed, and additionally CSV compositions are changed where needed, as in Figure \ref{fig:csvform} and detailed below.

\subsubsection{Processing of a CSV}

The process for CSVs are carried as follows: If no positive source of a CSV is observed at a given step, its state is deduced as 0 (undefined/unobserved). If at least one source is observed, and if there are both active and inactive targets among the CSV targets, then the CSV is duplicated with different target sets to create one copy that includes active targets and one copy that includes inactive targets (and any undefined targets are shared by both). This ensures that the CSV remains consistent, since it’s activation represents the activation of all its targets provided they are not undefined. There is no way to say whether an undefined target will be consistent with one duplicate or another after the changes to the CSV described below without observing a non-undefined state in them, so they are put into both copies and do not otherwise affect the state deduction of the CSV (except if all targets are undefined, see below).

Following this operation, if a CSV has any target active, then its state is deduced as active (1). If there is no perfect match with the standing sources of CSV and their activations (i.e. there are either inactive positive sources or active negative sources), these source lists are refined so that the remaining sources correspond perfectly to the current state of the network - in other words, any positive source that is inactive and any negative source that is active is removed. This refinement eliminates parts of the previously-posited relationships “hypothesized” to be necessary by the CSV in an exhaustive manner (see details on CSV formation, below) that are observed to be not necessary for the observation of the effect that the CSV models (Figure \ref{fig:csvform_3}).

If, on the other hand, the CSV has any inactive target (which is exclusive with any target being active due to the duplication-differentiation operation made above) and if not all its positive sources are active, then the state is deduced as 0, being consistent with the interpretation of a CSV as being defined only if all its positive sources are active. If however, all positive sources are active; then we look if any negative source is active that can justify the inactivation of the targets of the CSV. If there is at least one negative source that is active, we deduce the state as 0 since source conditions are not satisfied; and refine the negative targets that are not currently active in the same manner we described in the previous paragraph (due to the observation that they are seen to be not necessary for the suppression of the CSV - Figure \ref{fig:csvform_5}).

If, instead, all the targets of CSV are undefined, then the CSV is undefined as well.

A CSV is always created with only positive sources at first and no negative sources, and a CSV always starts as an unconditional CSV for whom we never expect to observe an inactive state (see below part for details on the generation of CSVs). At the observation of an inactive state in the CSV (i.e. one in which sources are active but targets are inactive), only once after the creation of the CSV, we duplicate the CSV and separate the targets that are currently undefined (to protect them from the change being made). In the duplicate that has the inactive targets, we connect the CSV with the negative sources by forming a negative sources list that encompasses all the currently-active eligible BSVs and DSVs in the model, which will be subject to future refinement (criteria of \textit{eligibility} is detailed in the Appendix, essentially corresponding to SVs that do not yield useful information). This, essentially, attempts to explain the CSV’s observed inactivation. If, however, an inactive state is observed despite already having formed connection with negative sources, then the unconditionally flag of the CSV is set to "conditional", representing that the CSV’s state is now uncertain (setting aside its possible conditioners).

\subsubsection{CSV generation and model refinement}
\label{subsec:csv_form}

After the traversal of SVs for computation of their states and modifications in CSV compositions, all DSVs and CSVs who are observed active but are neither unconditional nor have an active conditioner that explains their activation are labelled as \textit{unexplained}. We then form a CSV that, as positive sources, has all the eligible, currently-active BSVs and DSVs; and as target, has all the eligible SVs in unexplained list (Figure \ref{fig:csvform_1}). Any target which is left outside of this CSV, and hence remain unexplained, have their unconditionally flags set to "possibly conditional" (which basically signals that the SV can go active without any explanation or predictor).

Finally, at the end of the step, we refine the general model by removing any CSVs that may be duplicates of other CSVs (ending up representing the same thing from different histories), as well as any CSV that has no sources or targets left as a result of refinement or duplication operations.

\subsubsection{Source eligibility for CSVs}

To reduce model complexity and avoid the need for repeated exposures to the environment, we pre-filter sources during CSV formation or CSV negative-sources formation by their eligibility as follows: We define \textit{trivial sources} of a CSV as the sources of all the SVs that lie downstream starting from this CSV (i.e. SVs conditioned by this CSV, and CSVs conditioned by them, and so on), plus the associated BSV if a DSV is reached. Intuitively, these are the sources whose states can be determined by the knowledge that the CSV is active (since a CSV being active means that it’s target will be active as well, which will inform us about the states of its sources), and hence wouldn’t be informative sources for the current CSV as any information conveyed by them will be trivial. When forming a CSV, among all the currently-active BSV and DSVs, we filter those that provide trivial information to all the unexplained SVs (i.e. prospective targets for the generated CSV) out as positive sources, and take only those that do not provide trivial information as source to at least one of them. Furthermore, after this filtering, if there is a prospective target for which all the remaining prospective sources provide trivial information, then this target is not taken as a target of the CSV and hence remains unexplained.

In a similar spirit, when forming negative sources, we filter out all the candidates that provide trivial information for the CSVs. In addition, however, we filter out any upstream positive source (that is, the cumulative list of all positive sources among all upstream CSVs of this CSV, i.e. its conditioners and conditioners of its conditioners, including itself) because we already know (by the definition of the conditioning process) that there was an instance in which this CSV was observed when the SVs in this list of positive conditioners was also observed; and hence these negative sources would be eliminated in exposure with the same instance again.

\subsubsection{Conditioner formation for unconditional CSVs}

Here we note a modification that we do not employ currently, but is possible: Currently we allow no CSVs to condition unconditional CSVs since they are not informative and hence prevent the model from being minimal. However, we note that allowing for conditioners to be formed to unexplained (no active conditioners)  unconditional CSVs as well could result in these CSVs already having some conditioners learned from the previous encounters with the environment in case they ever turn conditional, reducing the required number of interactions for the learning of the full environment model, at the cost of making the model more exhaustive in terms of what is being modelled. This would require two changes: (1) At CSV formation, not excluding the unexplained CSVs that are unconditional; and (2) when refining positive sources, we create a CSV which takes as its initial positive sources that are being removed, and that conditions the CSV whose sources are being refined currently. This way, instead of removing what was observed to be active at previous encounters at which the CSV was active, we push them to an upper level of computation to represent an alternative condition in which the CSV was observed to be active before.

\subsubsection{Computational complexity}

The computational complexity of The Modeller learning loop, with respect to model complexity, is worst-case $O(N)$, where $N$ denotes the number of CSVs. This arises because, in the worst-case scenario, Algorithm~\ref{alg:algorithm_adaptationloop} may traverse the entire set of CSVs within the model. However, note that such full traversal is unlikely in practice; typically, only a relevant subset of CSVs (those that pertain to or activate in response to the current observation) is traversed.

\subsection{Proof of Theorem 1}
\label{sec:app_proof}

Let $X_P^i$ and $X_N^i$ be positive and negative sources of $C$ respectively that remains \textit{after} refinements that instance $y_i$ causes. Since we know that $C$ does not undergo negative sources formation, and that $y_0$ comes before $y_1$, we can say that $X_P^1 \subseteq X_P^0$ and $X_N^1 \subseteq X_N^0$ since only refinements are allowed on $X_P$ and $X_N$ sets of $C$ by our definition of operations.

We now analyse the two possible cases with respect to satisfaction of sources:
\begin{itemize}
    \item If, in the original encounter with $y_0$ the sources of $C$ were satisfied, then we had $S_x=1 \forall x \in X_P^0$ and $S_x=1 \forall x \in X_P^0$. Since $X_P^1 \subseteq X_P^0$ and $X_N^1 \subseteq X_N^0$, we will also have $S_x=1\ \forall x \in X_P^1$ and $S_x=1\ \forall x \in X_P^1$ at the new encounter with instance $y_0$. Hence, if sources of $C$ were satisfied in the previous encounter with $y_0$, they will remain satisfied in the new encounter. The value of $S_C$ can be -1 or 1 if and only if sources of $C$ are satisfied; in which case it is exclusively determined by the state of its targets (-1 if targets are inactive and 1 if targets are active). Since the states of targets are determined by $y_0$ and hence is the same across the past and new encounter with $y_0$; if $S_C=1 (-1)$ in the past exposure with $y_0$, then it will be $1(-1)$ in the new exposure as well.
    \item If, in the original encounter with $y_0$ the sources of $C$ were not satisfied (and hence original encounter yielded $S_C=0$), then we either had $S_x \neq 1\ \forall x \in X_P^0$ or $S_x = 1\ \forall x \in X_N^0$ (note that we defined $X_P^i$ and $X_N^i$ as source sets \textit{after} the refinements; and hence we know that in both cases it will be the whole of positive/negative source sets that have the property, and not a subset of them; since the source SVs that were not a part of that subset will have been refined). Since $X_P^1 \subseteq X_P^0$ and $X_N^1 \subseteq X_N^0$, we will also have either $S_x \neq 1\ \forall x \in X_P^1$ (if former) or $S_x = 1\ \forall x \in X_N^1$ (if latter), both of them not satisfying the sources conditions of $C$ (hence the new encounter with $y_0$ also yielding $S_C=0$.
\end{itemize}

Therefore, in all cases, response to $y_0$ remains identical before and after exposure to $y_1$.

\subsection{Learning the statistical significance of encountered relations}
\label{sec:statistical_significance}

The base mechanisms of The Modeller as described in the main text rest on an attempt of prediction of all encountered changes in state variables in the environment, forming an explanatory/predictive relationship between any two observed events in that attempt of full modelling of the environment. Unlike neural networks (or other statistical learning methods), the naive algorithm does not depend on, but also does not naturally incorporate, a method of statistically averaging and filtering learned relationships. Such a means of estimation of statistical significance of learned relationships can be incorporated into the models learned by The Modeller in a straightforward manner into the learned relationships locally, which in turn can be used to filter out non-significant relationships, hence preventing overcomplexification of the model.

Let $C$ be a CSV, and let $T$ be a target SV of that CSV. We define the event \textit{sources satisfied}, $SS(C)$, to be the event where all positive sources of $C$ are active and all negative sources are nonactive. For each target, we define an \textit{observation} of the target $O(T)$ to be when the target is observed (i.e. either active or inactive, state 1 or -1, as defined in the main text) and an \textit{incidence} of the target $I(T)$ to be when the target is active (state 1). We define the event \textit{concurrence} to be the event where both the sources of $C$ are satisfied and there is an indicence of target, $CC(C,T)=SS(C) \wedge I(T)$.

We quantify the statistical significance of a learned relationship between a set of sources of a CSV and one of its targets as the \textit{amount of increase in the probability of the incidence of the target given the satisfaction of the sources of the CSV}. We define \textit{normalized causal effect (NCE)} as the amount of increase in probability of incidence of $T$ that satisfaction of sources of CSV $C$ causes, normalized by the original probability of incidence:

\begin{equation}
    NCE = \frac{P(I(T)|SS(C)) - P(I(T))}{P(I(T))}
\end{equation}

The conditional probability in the nominator can be expanded as:

\begin{equation}
    P(I(T)|SS(C)) = \frac{P(I(T), SS(C))}{P(SS(C))} = \frac{P(CC(C,T))}{P(SS(C))}
\end{equation}

by our definition of concurrence $CC(C,T)$ above. All of the probabilities can be computed by locally tracking of the number of instances that the corresponding events are observed, when the target is observed (i.e. $O(T)=1$). When the target is unobserved/undefined, by extension none of the other events are observed.

A positive NCE means that $SS(C)$ increases probability of $I(T)$ and a negative NCE means that $SS(C)$ decreases it. An NCE of e.g. 2.0 means that $SS(C)$ increases probability of $I(T)$ to 3 times the original probability. Within the context of our modelling mechanism, a negative NCE means that the relationship between sources of $C$ and $T$ has been learned in the wrong direction - actual negative relations learned in proper direction will still result in positive NCE, because the sources of that relation will go within the negative sources of $C$ instead of the positive ones, still in the end resulting in the $SS(C)$. The lower the magnitude of NCE, the less significant the relationship is.

Given NCE values for each relationship, one can set a positive threshold $\epsilon_T$, where NCE values with magnitude below it are regarded as statistically insignificant. $\epsilon_T$ represents the trade-off between complete modelling and model complexity. After that separation of relationships into significant and insignificant ones, one can proceed either with their removal, or simply with blocking further conditioner formation for them to prevent overcomplexification in an attempt to predict a near-random relationship (i.e. to prevent "fitting the noise"). Since our main aim in employing this mechanism is to prevent overcomplexification, and since removal of such insignificant relationships from the model completely would result in their re-learning if the agent is exposed to them again; we opt for the latter option and block further conditioner formation for them.

NCE values may have other utilities for the processes of the agent. An example might be that it can be used in the prioritization of subgoals in the planner (see main text), where more reliable causal relationhips are prioritized over less reliable ones. We do not investigate into such utilities at this stage.

It’s important to note that the statistical estimates are not precise during the transient phase. This is due to the refinement mechanism, which prioritizes structural revisions and adjustments to make a given CSV align with observations where feasible. During this phase, estimates tend to overemphasize significance. However, these transients are brief, and NCEs insignificant CSVs quickly diminish once the refinements are complete and the CSV sources settle into their final form. Furthermore, this final form is typically less constrained, leading to more exposures over time in the same environment. Alternatively, we could eliminate these inaccuracies by resetting recorded statistics after each change to the CSV's composition, though this would increase the time needed for an NCE value to be deemed reliable. We do not use this approach here, as we do not find the temporary bias toward significance in transient SVs to be an issue, but it can be employed where precision has priority over efficiency.

We use the explained NCE metric to quantify the significance of relations in our base Modeller experiments. For MNR, we adopt a simpler approach to filtering learned relations, removing a conditioner $C$ of target $T$ if $P(SS(C)|I(T)) < \epsilon_{sign}$ (as in NCE, $SS(C)$ and $I(T)$ represent the satisfaction of $C$'s sources and the observation of $T$'s state, respectively). This approach was chosen to keep our MNR implementation simpler, as the precise representations required for environment modeling are unnecessary in this case, given that the task is a simple classification task. Experimentally, we find that both of these metrics are effective in quantifying the significance of learned relations.

\textbf{\textit{Effect on continual learning}} Notice that there is no change (particularly no decay) in neither NCE nor $P(SS(C)|I(T))$ if the target is not observed - hence, these measures of statistical significance do not decay (relationship "forgotten") in case of a changed environment in which the new one does not display the co-occurance of the two events (target and CSV sources being satisfied), as long as its target is not observed in isolation as well. If its target is observed in the new environment, two cases may occur:
\begin{enumerate}
    \item $P(I(T))$ is stable. This would be expected in an already-mature model or in environments where there is not much variability in the occurance of individual targets (even if the conditions under which they occur differ). In this case, there is no change in the metrics.
    \item $P(I(T))$ changes. In this case, the metrics will change according to $P(I(T))$. Note, however, that additional exposure can only mean a more accurate estimate of the true $P(I(T))$ value - any change in $P(I(T))$ hence does not have a detrimental effect, but instead makes the causal effect estimate more reliable in the context of the complete model; provided that the new environment itself does not have a probability of $P(I(T))$ in itself that is non-representative of the general probability, in particularly one that is excessively higher than the general one. This latter possibility (an immature estimate of $P(I(T))$ and an unnaturally high $P(I(T))$ in the new environment) is the only case in which a previously-learned correct relationship can be wrongly destroyed in case of a changing environment. But even such cases would have no long-term ramifications as $P(I(T))$ for any given target $T$ would reach to a reliable estimate after a few cycles of exposures to environments where $T$ is observed.
\end{enumerate}

The current method of computing and filtering based on statistical significance has one drawback, however; and it is that only first-order significance of relations are considered. In other words: If we have a CSV C0 with a target D0, and C0 (possibly unconditional) is conditioned by another CSV C1, then whether C0-D0 relationship will be regarded as significant or not depends only on the observations of sources of C0 and D0; and will \textit{not} consider their dependency on C1. This may result in unnecessary filtering in cases where a said statistical relationship is insignificant in the absence of a particular upstream conditioner, but becomes significant with that - we also see effects of this limitation to some degree in our results in the main text. Resolving this limitation requires considering and conditioning on higher-order conditioners when computing the metric values. This is left for future work, as it is not essential at the current initial stage of our design. It represents a more detailed mechanism for quantifying statistical relations, specifically one that can identify and construct \textit{significance groups} across variables in the model.

\subsection{Modifications to learning flow in MNR}
\label{sec:learning_flow}

The learning flow of MNR largely follows base Modeller's learning flow, except for the following adjustments:

\begin{enumerate}

    \item Unlike base Modeller, where upstream CSVs integrate with lower-level CSVs via an \textit{and} condition, our implementation treats upstream CSVs as observed \textit{subvariants} of their CSV targets, with source SPNs encompassing the source SPNs of them. This allows assignments from lower-order CSVs to propagate upstream, eliminating redundant assignments and simplifying the learning flow by avoiding the need for additional subnetwork definitions.

    \item Unlike base Modeller, which creates a common CSV for all targets in a step, we assume a single target per CSV and create separate CSVs (with identical sources) for each target. This change supports the upstream assignment propagation in previous point.

    \item Instead of embedding negative (suppressing) sources within a CSV, we externalize them into separate CSVs. A negatively-conditioning CSV is formed when a state variable with an inactive state and no active negative conditioner is observed (with potentially multiple formed per target). We also redefine the \textit{unconditionality} flag to deactivate upon the first observation of an inactive state, allowing simultaneous positive and negative upstream conditioning.
    
\end{enumerate}

The major changes, outlined in points 1 and 2, primarily affect CSV composition without altering the core learning flow or CSV state definitions. These changes were made for implementation simplicity and will be modified in future framework developments to include multiple targets and distinct upstream networks, improving representational efficiency (see Section \ref{sec:experiments_mnr}).

\begin{figure}
     \centering
     \begin{subfigure}{0.2\textwidth}
         \centering
         \includegraphics[width=\textwidth]{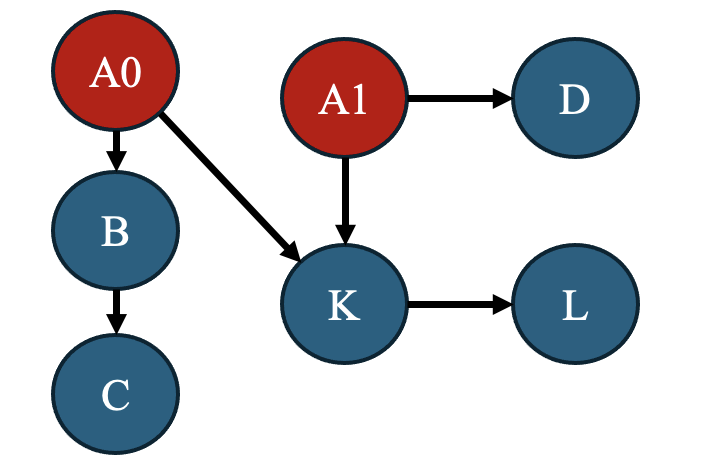}
         \caption{Source SN.}
         \label{fig:assignment_source}
     \end{subfigure}
     \hspace{1cm}
     \begin{subfigure}{0.2\textwidth}
         \centering
         \includegraphics[width=\textwidth]{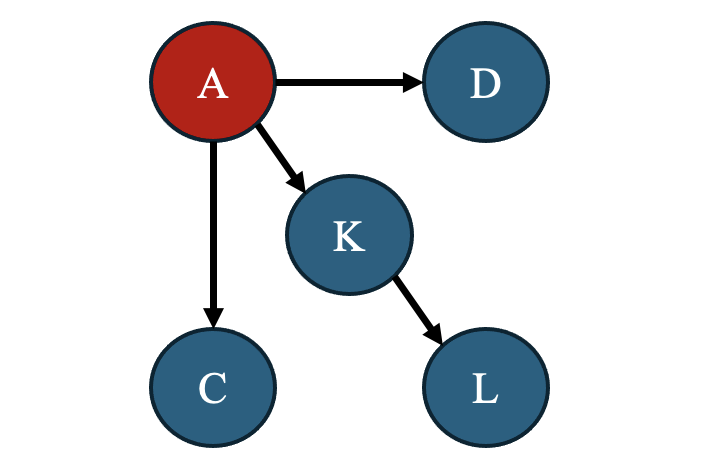}
         \caption{Refiner SN.}
         \label{fig:assignment_refiner}
     \end{subfigure}
     \vfill
     \begin{subfigure}{0.2\textwidth}
         \centering
         \includegraphics[width=\textwidth]{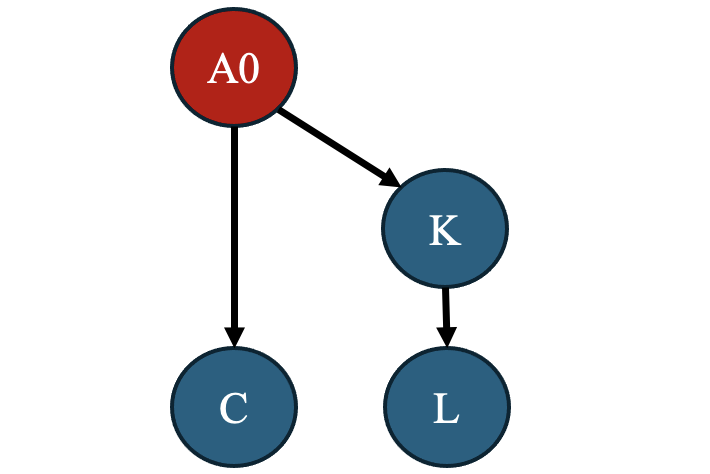}
         \caption{Refined, with $A \rightarrow A0$.}
         \label{fig:assignment_final_0}
     \end{subfigure}
     \hspace{1cm}
      \begin{subfigure}{0.2\textwidth}
         \centering
         \includegraphics[width=\textwidth]{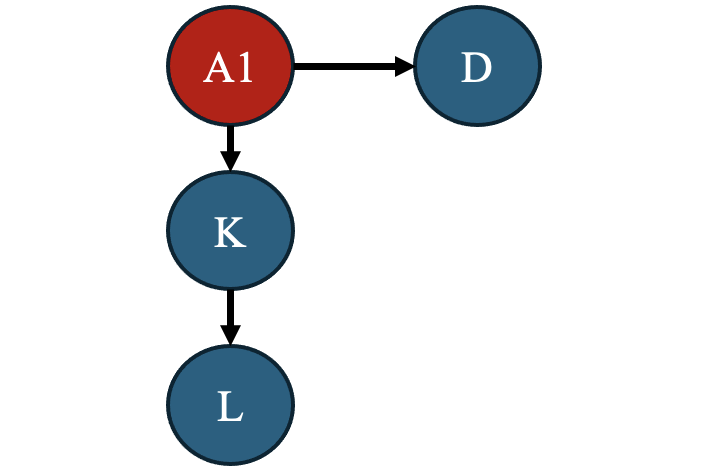}
         \caption{Refined, with $A \rightarrow A1$.}
         \label{fig:assignment_final_1}
     \end{subfigure}
    \caption{An illustration of multiple viable assignments in MNR. A node ($A$) of a given type (red) in the refiner SPN can map to multiple nodes in the source network ($A0$ or $A1$), resulting in two post-refinement networks with no clear superior assignment.}
    \label{fig:assignment}
\end{figure}

\subsection{Details of feature representation for MNR}
\label{sec:app_feature_representation}

MNR is applicable to any observation space representable as networks. For our experiments, we focus on demonstrating the method in a basic visual processing domain: 2D shape identification, using MNIST as the test domain. Below, we detail the feature representation (image-to-network conversion) used. We note that this is only demonstrative for the use of our approach in a simple context, but future work can extend it to other types, including 2D features like color gradients or 3D features with spatial positions, as the domain requires.

To ensure generality in shape detection, we use gradient orientation change points in the x and y axes. The process involves: (1) thresholding (50\%) the grayscale image, (2) approximating contours with OpenCV’s Ramer–Douglas–Peucker algorithm \cite{opencv_shape_analysis} (Fig. \ref{fig:representation_example_bcn}) and (3) computing gradient signs at corners in x and y dimensions based on edge orientation (e.g., a right-facing edge having a negative x-gradient). Nodes are created where gradient directions in either axis change from the predecessor to the successor edges as we traverse the contour, with node types defined by the axis, direction, and corner convexity. For instance, a top-right convex corner with a y-gradient change from positive to negative is classified as \textit{"convex, +y to -y"}. This defines the SPN’s nodes.

Our SPN includes four SN types: \textit{contour}, \textit{inner}, \textit{outer}, and \textit{all}. \textit{Contour} networks link nodes along the contour, while \textit{inner} and \textit{outer} SNs connect nodes within the inner (value 1) or outer (value 0) regions. \textit{All}-type SNs combine all connections to preserve relative positional relationships even when node types vary. Each type has \textit{horizontal} and \textit{vertical} variants, with directed edges indicating positional relationships along the respective axis. For example, a directed edge $(n_0, n_1)$ in horizontal means $n_0$ is to the left of $n_1$. See Fig. \ref{fig:representation_example_vn} for an example SPN.

\textit{Limitations and possible extensions:} This representation is broadly applicable and domain-agnostic for 2D shape detection, relying only on edge detection. However, it’s mainly for demonstrating our design and has limitations, such as limited expressivity from not capturing all details of the shape but only those that represent gradient sign changes (e.g., only three of five corners at the bottom are used in Fig. \ref{fig:representation_example_vn}). This affects final identification accuracy, making it fall short from a perfect performance as discussed in Section \ref{sec:experiments_mnr}. A more complete approach would consider all gradient orientation changes, increasing complexity but also completeness. Likewise, it is currently tailored for shape detection, but could be extended to other 2D features or 3D spaces with similar logic, as well as to alternative representations like pixel-level processing, CNN filters, or frequency-based transformations, as discussed in more detail in Sec \ref{sec:conclusion}. Finally, we note that multiple feature types can be used with SPN representation, either in separate networks or within the same network, capturing positional relations between features across domains. While we don't explore this here, it’s an interesting direction for future work.

\subsection{Details of experimental framework}
\label{sec:appendix_expsetup}

\paragraph{Significance filtering for The Modeller} The Modeller's mechanism of filtering based on statistical significance (i.e. NCE) is enabled only for the random variant of the environment. When enabled, we used a cutoff NCE of 0.25 for blocking upstream conditioner formations (i.e. no more upstream conditioners are formed if the CSV does not cause a >25\% in the probability of occurrence of its target).

\paragraph{Intuition regarding the design of environment in Figure \ref{fig:environment}} The environment was inspired from Multiroom environment in Minigrid. The states represent closed door (DC), open door (DO), wall (W), subgoal 1/2 (SG1/2), goal (G) and a random variable (X); "RS" stands for "rooms" and represents an agent going through multiple rooms opening doors in each, and "SGS" represents one in which agent reaches two subgoals and then reaches the goal afterwards, and "NEG" represents a case where goal appears conditioned on one positive and one negative conditon. In all, the goal can be moving. Alternative outcomes are present in all environment subtypes, since each of them allows for multiple outcomes following an empty ("-/-") state. Alternative predecessors are tested in "SGS" environment where SG2 can be preceded by SG1 in either of the two cells; and likewise in general the appearance of G can be preceded by any of the alternatives associated with different environment subtypes. The capability to represent positive and negative relations together is tested in subtype "NEG", in which G appears only if X is enabled in the first cell and not the second one.

\paragraph{Detailed settings for MNR experiments} We use $N_{sample}=20,\ 10,\ 5$ and test set size of $50,\ 20,\ 10$ samples per class for experiments with $N_C=3,\ 5,\ 10$ respectively. Reported results are averages of 10 runs for $N_C=3$ and 5 runs for $N_C=5$. Population size for generating assignments was chosen as $10$ for both learning and prediction. For MNR, we choose refinement threshold $T_{ref}=0.05$, significance threshold $\epsilon_{sign}=0.05$. $\epsilon$ for polygonal approximation is $0.01L$ where $L$ is the arc length of contour being approximated. Our fully connected neural network architecture used for comparison is of 2 hidden layers with 128 neurons each, while the CNN architecture has a pair of convolutional (32 filters with 3x3 kernels) and max-pooling (with pool size 2x2) layer, repeated twice sequentially, followed by a dense layer of 128 neurons. All NNs use ReLU activation in hidden layers and softmax in output. We use a maximum of 100 epochs and an early stopping patience of 10 epochs. All remaining settings are Keras defaults.

\paragraph{Prediction in MNR} To predict targets for a given observed SPN, we follow this procedure: First, we attempt to find an assignment (as described in \ref{sec:netref_description}) that \textit{satisfies} the source SPN of the CSV. If no assignment is found, the CSV is considered inactive. If an assignment is found, we compute the activation probability of the CSV. If the CSV is unconditional (no positive/negative conditioners), the probability is $p = P(I(T)|SS(C))$, tracked over the learning process. If the CSV is conditional, the probability is $p = (1 - p^{max}-) \times p^{max}+$, where $p^{max}-$ and $p^{max}+$ are the maximum activation probabilities of negative and positive conditioners, respectively. Intuitively, this approach prioritizes the most-upstream representations (most closely matching the observed SPN) when calculating the final probability, resolving conflicts between activating and suppressing pathways by multiplying their probabilities.

\paragraph{Computation resources} All experiments were run on a 2.4GHz 8-Core Intel Core i9 processor with 32 GB 2667MHz DDR4 memory. No GPU was used. Giving an accurate estimate for computation time is impossible since experiments were run in parallel to unevenly-distributed independent workloads.

\subsection{A sample model learned on SMR}

\begin{figure}
  \centering
  \includegraphics[width=0.5\textwidth]{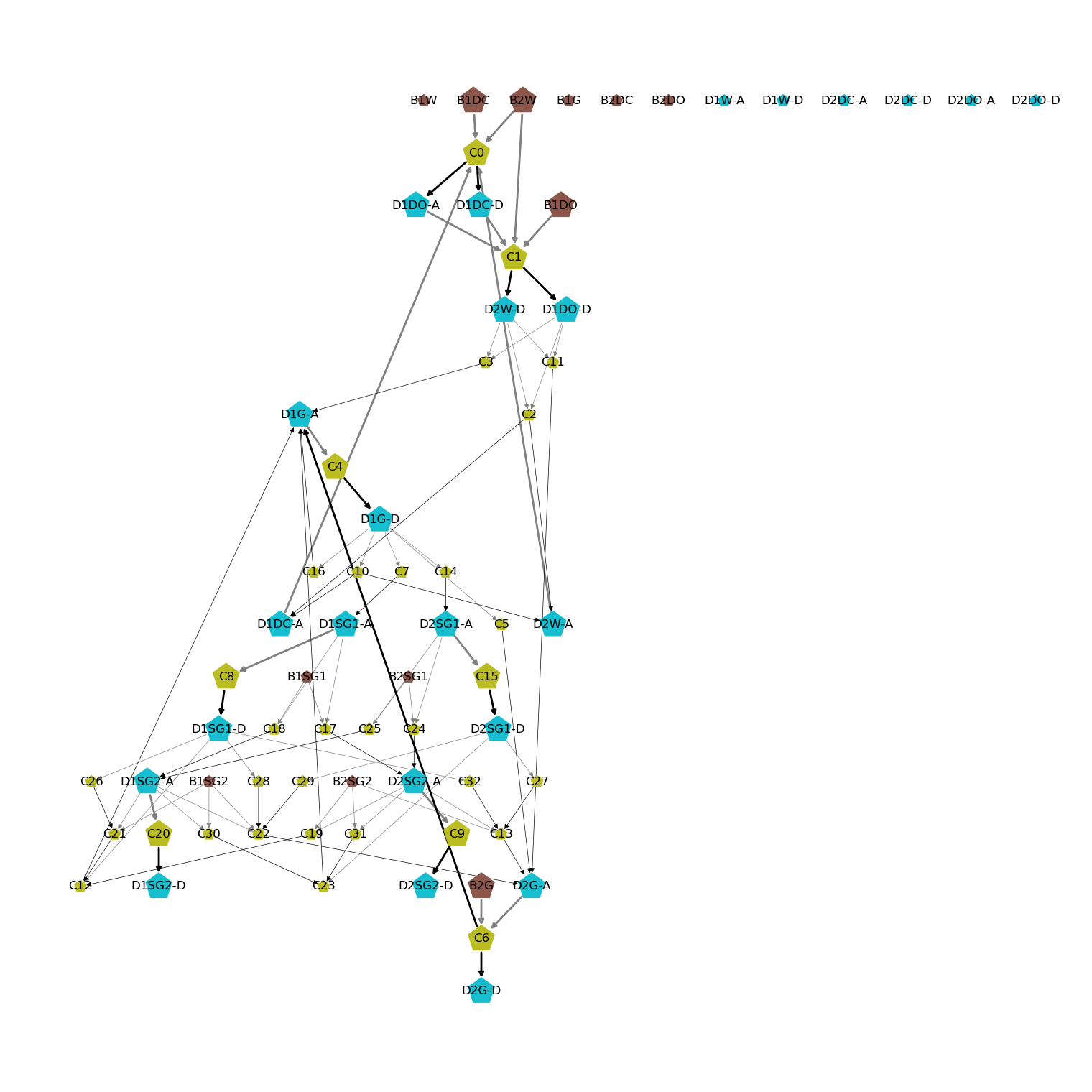}
  \caption{A sample environment model learned by The Modeller. In the visualized model, brown nodes are BSVs, blues are DSVs, and the rest are CSVs. The enlarged pathways (bold arrows and large nodes) are reliable outcomes (i.e. unconditional CSVs) and the rest are uncertain (possibly conditional) ones. Black arrows represent conditioning relationships and gray arrows represent source relationships (all positive in this example). Disconnected SVs (those that can never be activated by environment design) are cut for visual clarity.}
  \label{fig:sample_model}
\end{figure}

\begin{figure}
  \centering
  \includegraphics[width=0.5\textwidth]{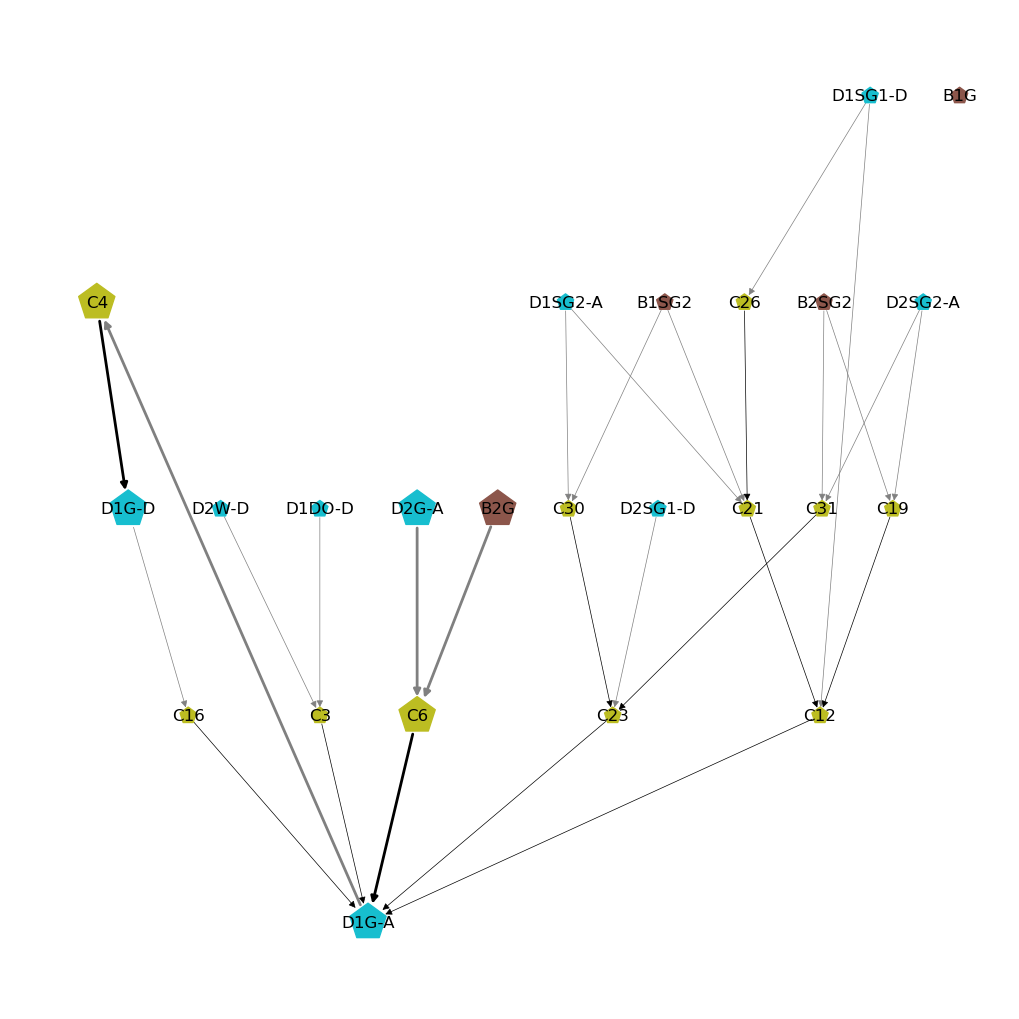}
  \caption{Same model as Figure \ref{fig:sample_model}, but for the predictive pathway of BSV 1G only. Many pathways for the activation of 1G can be seen in a human-comprehensible way in this model via the distinct CSVs preceding it (C3, C6, C12 C16, C23) and that the only reliable one of them is C6, and whose further sources can be seen by pursuing them upstream. In contrast, interpretation of a neural network model is much less straightforward due to nonlinearities, continuous parameters, and extensive connectivity that ties each neuron at the output to virtually all other neurons in the network.}
  \label{fig:sample_model_goal}
\end{figure}

\begin{figure}
  \centering
  \includegraphics[width=0.5\textwidth]{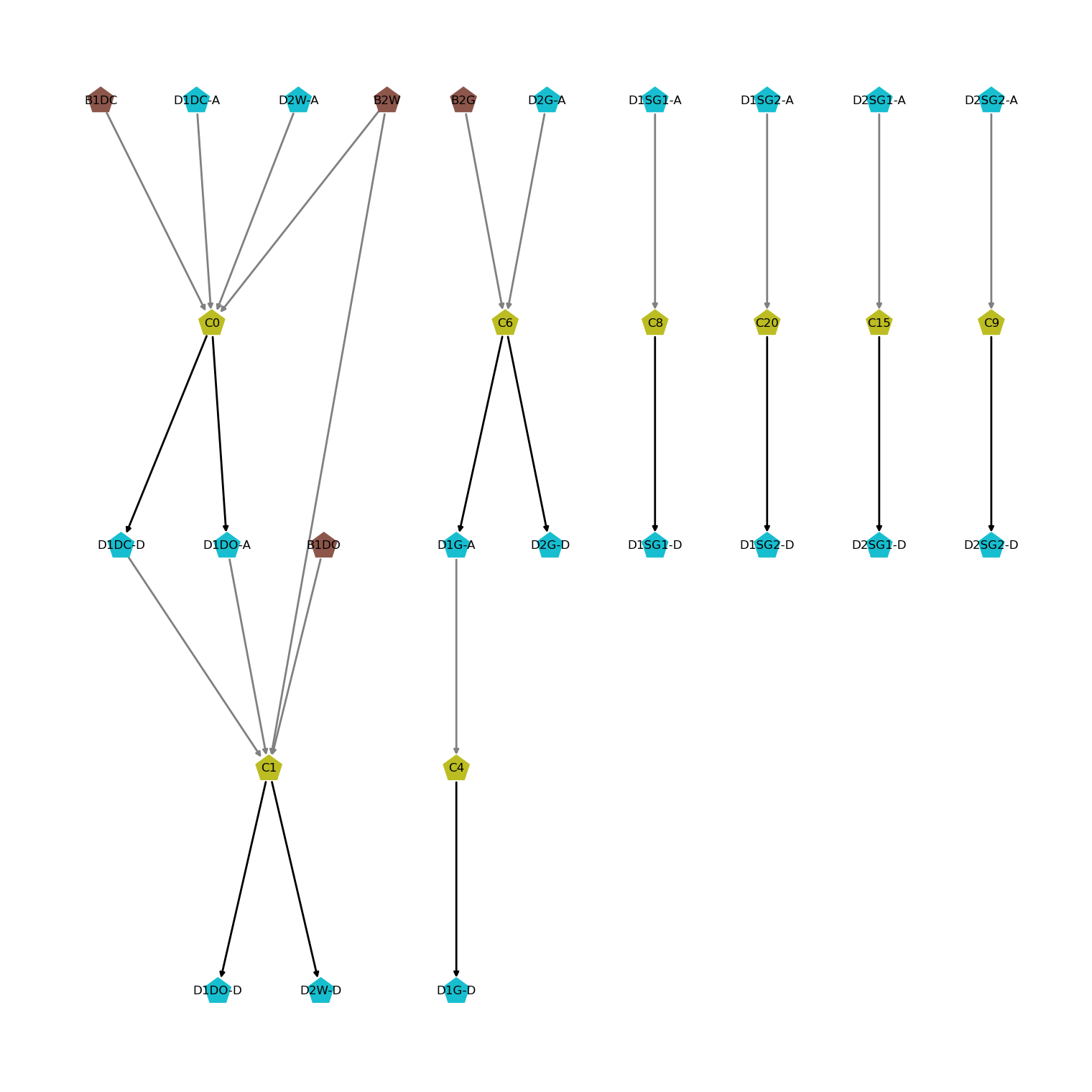}
  \caption{Same model as Figure \ref{fig:sample_model}, but with reliable pathways only, showing "islands of certainty" as potential candidates for abstraction.}
  \label{fig:sample_model_reliable}
\end{figure}

A sample model learned on the SMR environment (Figure \ref{fig:environment}) is provided on Figure \ref{fig:sample_model}. Figure \ref{fig:sample_model_goal} provides, as an example, the pathway of BSV 1G (state G at cell 1), in which the specific pathways connecting to this BSV can be seen more clearly in a human-comprehensible manner. Figure \ref{fig:sample_model_reliable} shows the whole model, but only with reliable connections; clearly showing "islands of certain state transitions" which can be an example of a delimiting criterion that can be used for abstractions as discussed in the main text.

\begin{algorithm*}[tb]
\caption{Overview of the planning algorithm, relying on recursive generation of upstream \textit{action networks} (the graph of behaviors required to realize the desired goals from the currently active SVs).}

\label{alg:planner}
\textbf{Function} Plan(currentActiveSVs, goalSVs)
\begin{algorithmic}[1] 
    \STATE ActionNetwork $\leftarrow$ \ EmptyNet
    \FOR{SV, target $\in$ goalSVs}
        \STATE GenerateUpstreamAN(SV, target)
    \ENDFOR
\end{algorithmic}

\textit{Comment: Argument "target" states what the desired state is in the SV, which can be activation (A), deactivation (D), active (1) or nonactive (0). Irrelevant for CSVs.}

\textbf{Function} GenerateUpstreamAN(SV, target)
\begin{algorithmic}[1] 
    \STATE \textbf{if} satisfiedByCurrentActives(SV, target): return \textbf{True} 
    \STATE pathways $\leftarrow$ EmptyList
    \IF{type(SV) in [BSV, GSV]}
        \STATE pathways.add(Precondition(sv, target))
        \STATE{\textit{Comment: These are the preconditions for target to occur in a SV. For (A, D, 1, 0) they are (0, 1, A, D) respectively; since a SV must be activated for itself to be active, needs to be inactive for itself to get activated, and so on.}}
        \STATE pathways.add(Constituents(sv), target)
        \STATE pathways.add(Constituencies(sv), target)
        \STATE \textbf{if} target in ['A', 'D']: pathways.add(Conditioners(sv, target))
    \ELSIF{type(SV) is CSV}
        \STATE pathways.add(Sources(sv))
        \STATE pathways.add(Conditioners(sv))
    \ENDIF
    \STATE \textbf{if} pathways is Empty: return \textbf{False}

    \FOR{upstreamSV, upstreamTarget in pathways}
        \STATE ActionNetwork.AddEdge((upstreamSV, upstreamTarget), (SV, target))
        \STATE GenerateUpstreamAN(upstreamSV, upstreamTarget)
    \ENDFOR
\end{algorithmic}
\end{algorithm*}

\begin{algorithm*}[tb]
\caption{Behavior encapsulation.}

\label{alg:behenc}
\textbf{Function-}\textit{EncapsulateBehavior}(ActionNetworks)

\textbf{Parameters:} ActionNetworks: The list of action networks over which behavior encapsulation will be conducted.

\begin{algorithmic}[1] 

    \STATE \textit{Comment: If only one action network is provided, return that network directly.}
    
    \IF{length(ActionNetworks) is 1}
        \STATE \textbf{return} ActionNetworks[0]
    \ENDIF

    \STATE EncapsulatedAN = constructEncapsulatedAN(ActionNetworks)
    
    \FOR{(sourceNode, targetNode) in EncapsulatedAN.edges}
        \STATE connectingSubnetworks = [getConnectingSubnetwork(net, sourceNode, targetNode) for net in actionNetworks]
        \STATE CSNEntry = encapsulateBehavior(connectingSubnetworks)
        \STATE EncapsulatedAN.edges[sourceNode, targetNode]['connectingSubnetworks'] = CSNEntry
    \ENDFOR

\end{algorithmic}

\textbf{Function-}\textit{constructEncapsulatedAN}(ActionNetworks)

\begin{algorithmic}[1] 
    \STATE EncapsulatedAN = ActionNetworks[0].copy()

    \FOR{n $\in$ nodes(EncapsulatedAN)}                    
        \IF{not(n $\in$ net for all net in ActionNetworks)}
            \FOR{$(n_0, n_1) \in edges(EncapsulatedAN, n)$}
                \STATE RemoveWithRerelation($net, n_0, n_1)$
            \ENDFOR
            \STATE net.RemoveNode($n$)
        \ENDIF
    \ENDFOR
    
    \FOR{$(n_0, n_1) \in edges(net)$}
        \IF{not($path(n_0,n_1)$ $\in$ net for all net in ActionNetworks)}
            \STATE RemoveWithRerelation($net, n_0, n_1)$
        \ENDIF
    \ENDFOR
\end{algorithmic}

\textbf{Function-}\textit{getConnectingSubnetwork}(net, sourceNode, targetNode)

\begin{algorithmic}[1] 

    \STATE \textit{Comment: Get the intermediate subnetwork connecting source and target.}
    
    \STATE intermediateNodes = $\bigcup$ allSimplePaths$(net, sourceNode, targetNode)$
    
    \STATE intermediateSubnetwork = createNetworkCopy(inducedSubgraph$(net, intermediateNodes)$)
    
    \STATE \textit{Comment: Get the upstream subnetwork of the nodes in the intermediate subnetwork.}
    
    \STATE upstreamNodes = $\emptyset$
    \FOR{$n \in nodes(intermediateSubnetwork)$}
        \IF{$n \neq sourceNode$ \AND $n \neq targetNode$}
            \STATE upstreamNodes $\mathrel{+}= \{n\} \cup$ ancestors$(net, n)$
        \ENDIF
    \ENDFOR
    
    \STATE upstreamSubnetwork = createNetworkCopy(inducedSubgraph$(net, upstreamNodes)$)
    
    \STATE \textit{Comment: Return the combination of the intermediate subnetwork and its upstream subnetwork.}
    \STATE \RETURN merge$(intermediateSubnetwork, upstreamSubnetwork)$
    
\end{algorithmic}

\textbf{Function-}\textit{RemoveWithRerelation}($net$, $n_0$, $n_1$)

\begin{algorithmic}[1] 
    \FOR{$(p,s)\ \in\ prod(P_{net}(n_0),\  S_{net}(n_1))$}
        \STATE $net$.AddEdge($p$, $s$)
    \ENDFOR
    \STATE $net$.RemoveEdge($n_0$, $n_1$)
\end{algorithmic}

\end{algorithm*}

\begin{algorithm*}[tb]
\caption{Network refinement with rerelation.}

\label{alg:netref}
\textbf{Function-}\textit{RefineBy}($P_0$, $P_1$, f)

\textbf{Parameters:} $P_0$, source SPN. $P_1$, refiner SPN. $f$, a partial assignment between nodes in $P_0$ to nodes in $P_1$.

\begin{algorithmic}[1] 
    \FOR{$SN^0_i \in P_0$}
        \FOR{$n \in nodes(SN^0_i)$}                    
            \IF{$f(n)$ not defined}
                \FOR{$(n_0, n_1) \in edges(SN^0_i, n)$}
                    \STATE RemoveWithRerelation($SN^0_i, n_0, n_1)$
                \ENDFOR
                \STATE $SN^0_i$.RemoveNode($n$)
            \ENDIF
        \ENDFOR
        
        \FOR{$(n_0, n_1) \in edges(SN^0_i)$}
            \IF{$path(f(n_0), f(n_1))$ not in $SN^1_i$}
                \STATE RemoveWithRerelation($SN^0_i, n_0, n_1)$
            \ENDIF
        \ENDFOR
    \ENDFOR

    \STATE \textit{Comment: Function RemoveWithRerelation is as defined in Alg. \ref{alg:behenc}.}

\end{algorithmic}

\end{algorithm*}

\subsection{Additional illustrations of representations learned by MNR}

\begin{figure*}
     \centering
     \begin{subfigure}{0.45\textwidth}
         \centering
         \includegraphics[width=\textwidth]{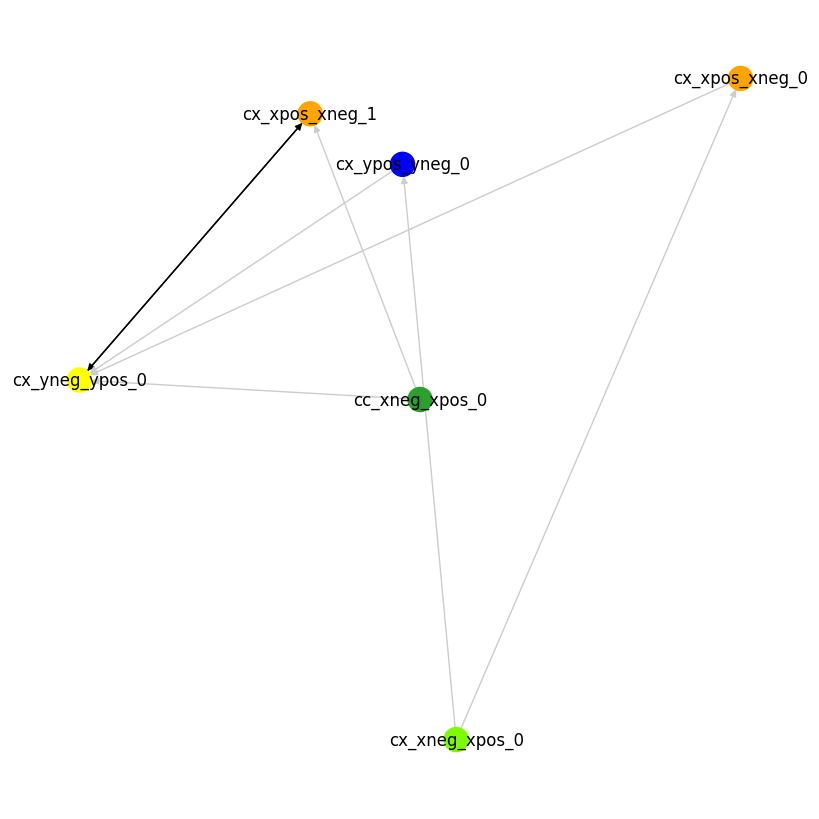}
         \caption{Digit 4, depth 0.}
         \label{fig:results_spns_4_ds}
     \end{subfigure}
     \begin{subfigure}{0.45\textwidth}
         \centering
         \includegraphics[width=\textwidth]{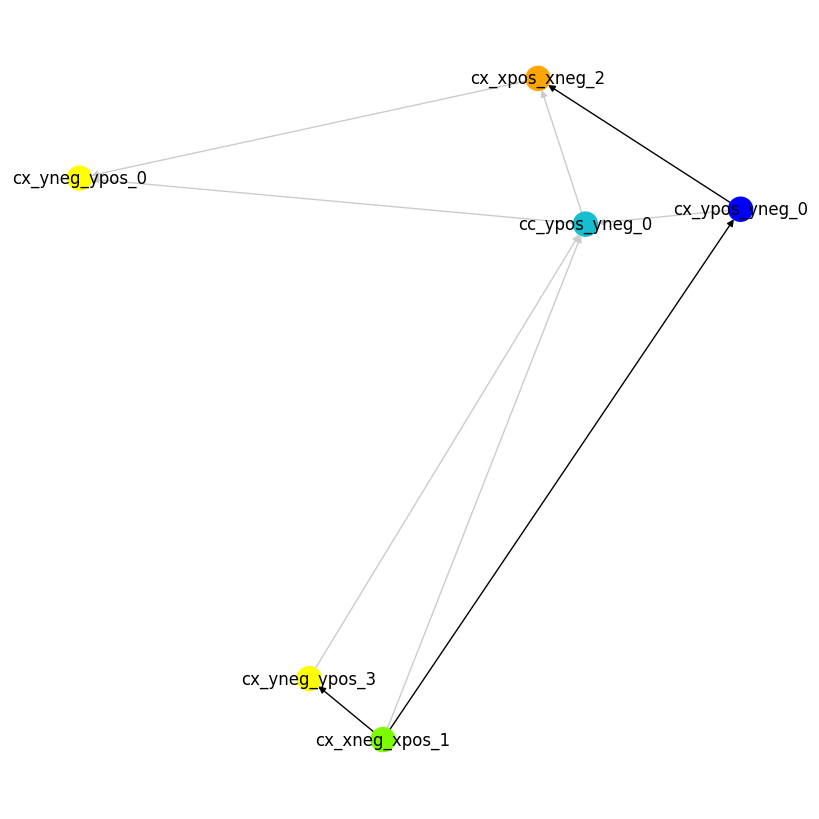}
         \caption{Digit 7, depth 0.}
         \label{fig:results_spns_7_ds}
     \end{subfigure}
      \begin{subfigure}{0.45\textwidth}
         \centering
         \includegraphics[width=\textwidth]{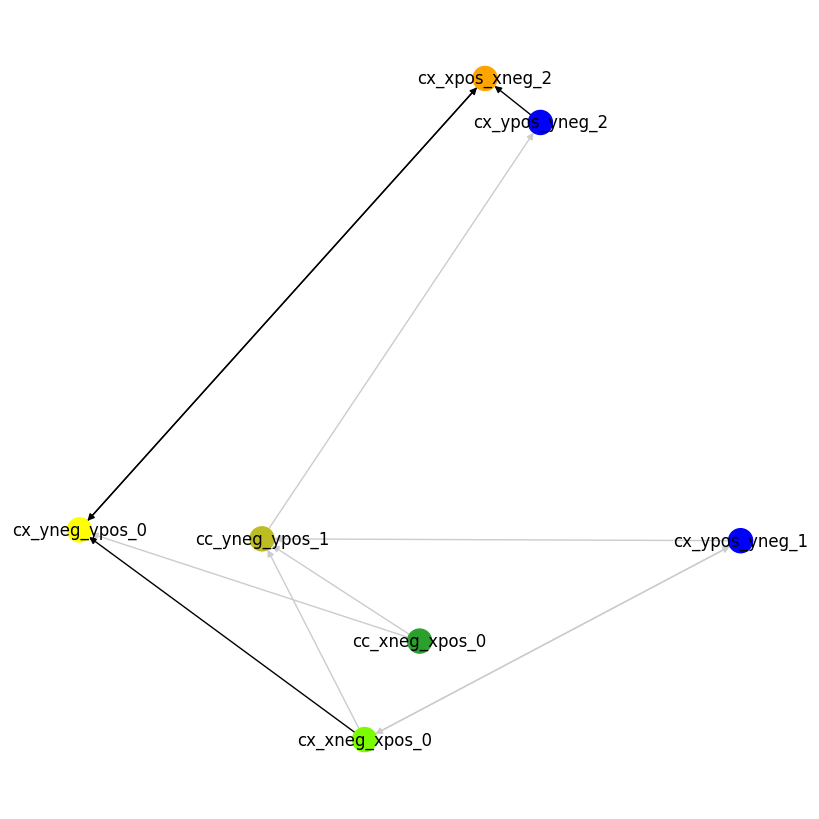}
         \caption{Digit 6, depth 0.}
         \label{fig:results_spns_6_ds}
     \end{subfigure}
     \begin{subfigure}{0.45\textwidth}
         \centering
         \includegraphics[width=\textwidth]{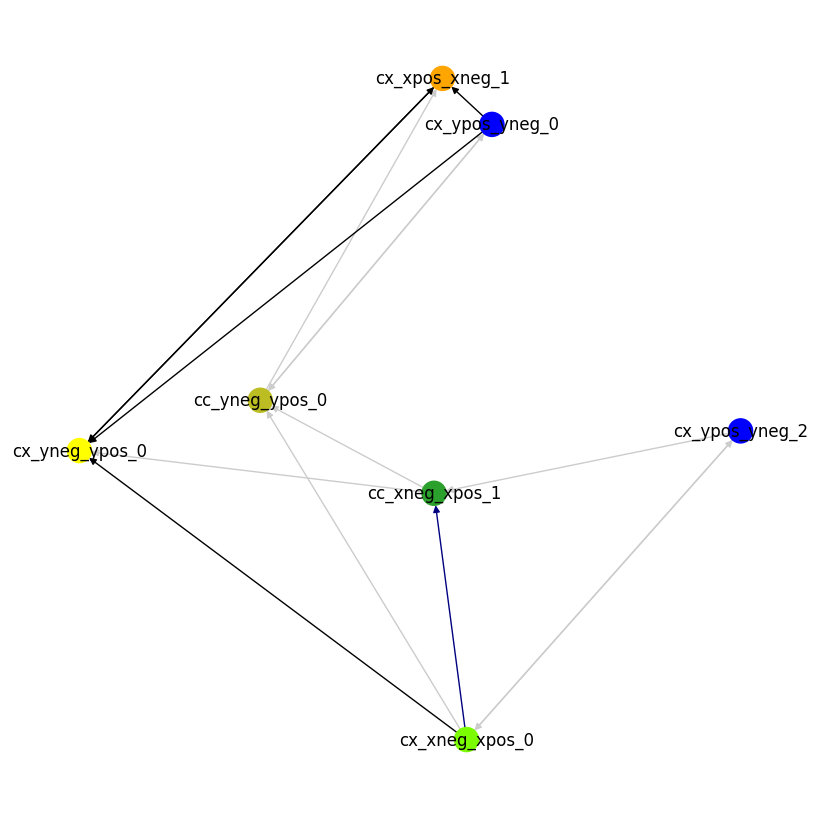}
         \caption{Source of a negative conditioner (suppressing connection) for digit 6, initialized from an instance of digit 4.}
         \label{fig:results_spns_6_negfrom4}
     \end{subfigure}
    \caption{Additional examples of representations of digits learned by MNR. Blue, black, and gray edges represent inner, contour, and all-type connections, respectively. Depths indicate conditioning distance from the target variable (number of CSVs in the conditioning path). Node positions reflect the average of all observed positions during the CSV’s lifetime.}
    \label{fig:results_spns_appendix}
\end{figure*}

Figure \ref{fig:results_spns_appendix} presents some additional learned representations in the runs during our experiments presented in the main text.

\subsection{On computational viability of developmental processes}
\label{sec:appendix_development_viability}

A possible concern regarding developmental processes may be computational viability. From the perspective of model size, we can be sure of feasibility—neural networks are already overparameterized for the tasks they learn \cite{du2018power}. In contrast, with proper minimal-growth methods, it is possible to construct a model that does not exceed the model complexity required by the task, aside from possible transients of local variation (e.g. exhaustive formation in The Modeller, Sec. \ref{sec:Modeller}). Concerning runtime, while individual learning steps would indeed exceed those of NNs, the capacity for continual learning (entailing online learning) removes the necessity for repetitive passes through the dataset (see our continual learning experiments in MNIST for a clear demonstration), which are the main culprits for the prolonged training sessions of traditional NNs. Notably, this stepwise-slower yet online learning regime is what is seen in animals, with even humans displaying nothing like the instantaneous gradient updates in NNs. Hence this approach is computationally feasible from both a size and runtime standpoint. This is not to say that feasibility will be immediately evident (e.g. we run into some size issues in both of our experimental regimes in this paper) but it means that there are no insurmountable limitations beyond those already imposed by current methods.

\subsection{Experimental performance of neural network-based continual learning systems}
\label{sec:nn_cl_exps}


This section demonstrates the performance of two types of neural network-based continual learning systems: (1) a method based on direct replay of past samples, and (2) a method based on creating multiple experts while detecting previously observed or novel tasks through autoencoders. These two methods respectively exemplify approaches that require the \textit{constraining assumptions} of (1) continued availability of past data, and (2) clearly defined task boundaries, as discussed in Section~\ref{sec:background}. Consequently, they are not directly comparable with The Modeller (which operates without these assumptions), and the analysis here is provided more for completeness rather than direct comparison. The main goal is to examine what happens as these assumptions are gradually relaxed, to the extent possible. For clarity of demonstration, all experiments are conducted on MNIST with 3 enabled classes, using the same training regime as in Section~\ref{sec:experiments_mnr}, and results are compared to MNR, since this setup focuses specifically on the continual learning aspect.

\subsubsection{Continual learning with sample replay}

For the first baseline, we test a continual learning method based on direct sample replay, relying on the assumption of continued availability of past data, which MNR does not require. We follow the training regime in Section~\ref{sec:experiments_mnr}, where training on one task in a cycle consists of 20 training iterations, each iteration comprising a batch of 10 samples. A fraction \( R_{replay} \) of each batch is drawn from a replay buffer. This buffer has size \( N_{buffer} = 10 \times R_{replay} \), filled with randomly selected samples evenly drawn from all previous (global) iterations. Replay recording is done per training iteration (not at task end) to remove dependency on task boundaries, which we want to test the replay method without. The central constraining assumption behind this method is the sustained availability of relevant past data in sufficient volume and distribution to be effective alongside new task data. To test the influence of the validity of that assumption, we vary \( R_{replay} \) from 0.8 to 0 to observe how performance degrades with weakening of the assumption.

\begin{figure*}
     \centering     
     \begin{subfigure}{0.4\textwidth}
         \centering
         \includegraphics[width=\textwidth]{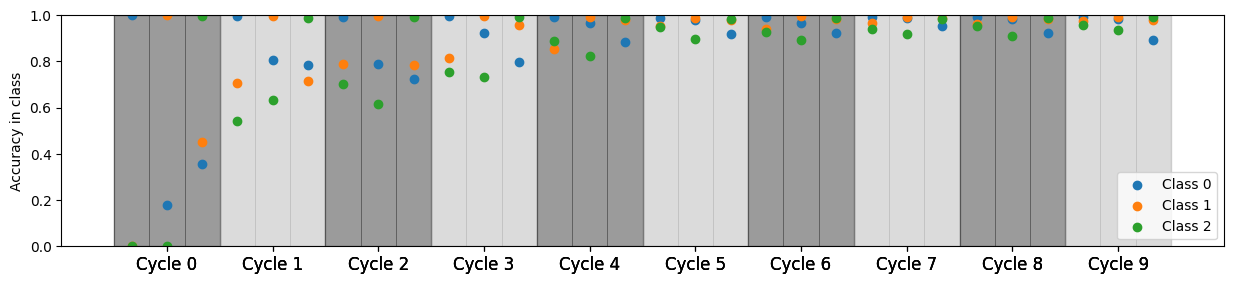}
         \caption{$R_{replay}=0.8$.}
         \label{fig:cl_replay_ratio0p8}
     \end{subfigure}
     \begin{subfigure}{0.4\textwidth}
         \centering
         \includegraphics[width=\textwidth]{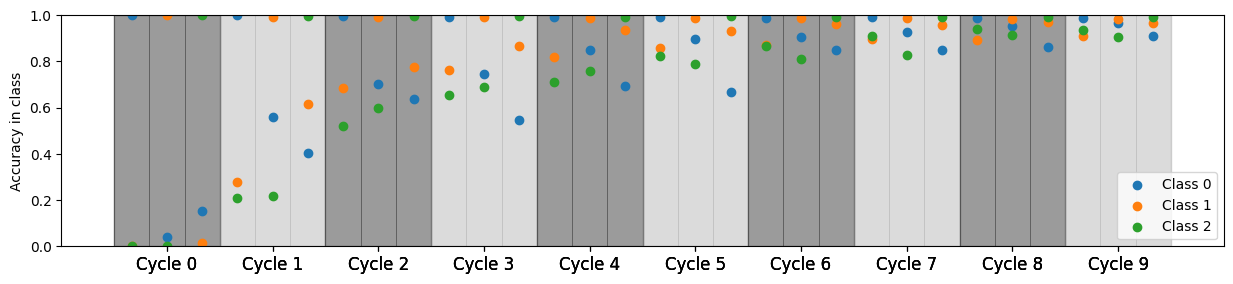}
         \caption{$R_{replay}=0.5$.}
         \label{fig:cl_replay_ratio0p5}
     \end{subfigure}
     \begin{subfigure}{0.4\textwidth}
         \centering
         \includegraphics[width=\textwidth]{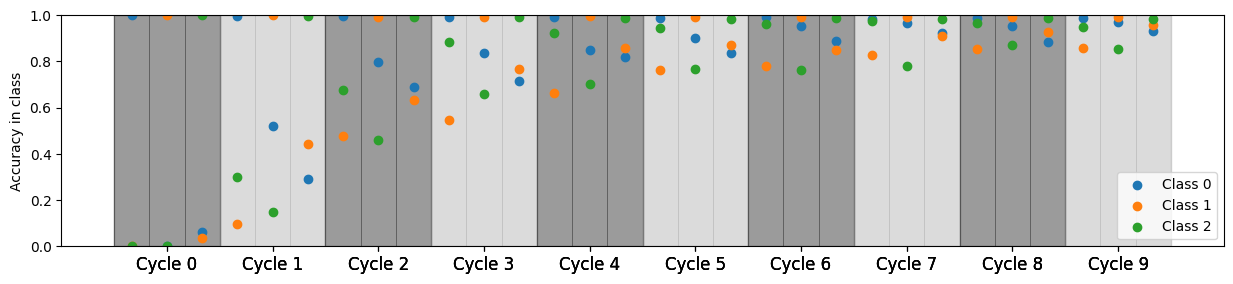}
         \caption{$R_{replay}=0.2$.}
         \label{fig:cl_replay_ratio0p2}
     \end{subfigure}
     \begin{subfigure}{0.4\textwidth}
         \centering
         \includegraphics[width=\textwidth]{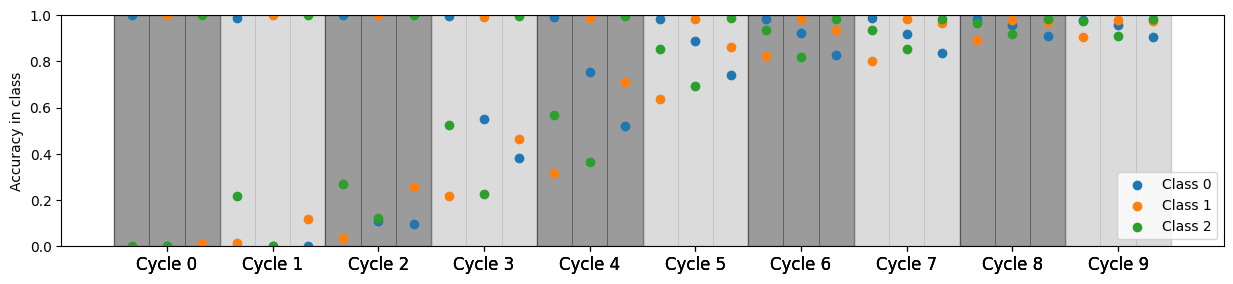}
         \caption{$R_{replay}=0$.}
         \label{fig:cl_replay_ratio0}
     \end{subfigure}
    \caption{Learning performance of replay-based neural network on $N_C$-class incremental learning over 10 cycles, for different ratios of the number of replay samples within the total batch. Accuracies reflect correct classification ratios for each class. Shaded areas denote cycles, and vertical lines separate iterations within cycles. Results are averaged over 8 runs. Note that class indices $i$ are randomly chosen at the start of each run and do \textit{not} necessarily correspond to digit $i$.}
    \label{fig:results_cl_replay}
\end{figure*}

Figure~\ref{fig:results_cl_replay} shows the results. We observe that (1) reaching stable performance takes more cycles of exposure to each task with lower replay ratios, and (2) visible catastrophic forgetting occurs in all cases, especially as replay ratio is reduced\footnote{Notice that this is because, as detailed above, we do not assume visibility of task boundaries and update the replay buffer at every training iteration, in order to bring this learning flow a bit closer to the unconstrained setup of MNR.}. Even with replay, the system shows significantly weaker retention than MNR. Importantly, performance during earlier iterations is key—since, as discussed in the main text, at long timescales the learning reduces to SGD with slower dynamics, which neural networks handle well. In summary, in terms of early retention, MNR clearly outperforms replay-based methods. We emphasize again that on top of the performative advantage, this is still not a fair comparison of methods: replay-based continual learning critically depends on the availability of past data, an assumption not shared by MNR.

\subsubsection{Continual learning with multiple experts and novelty detection}

As the second baseline, we use a continual learning method based on the creation of multiple, isolated neural network experts per task, combined with an autoencoder-based task–expert matching and new task detection mechanism. This approach, similar to \cite{jacobson2022task, erden2024directed, erden2025continual}, represents the least constrained among the NN-based multiple-expert methods we know of. (However, it still relies—albeit to a lesser extent—on the presence of task boundaries, unlike the approaches proposed in this paper.)

The learning flow of this regime is as follows:
\begin{itemize}
\item There are multiple experts (each being a neural network), and each expert has an associated autoencoder. Every autoencoder maintains an average reconstruction error and its standard deviation.
\item At each training iteration, the agent receives a batch of size $N_{batch}$ from the current task. The reconstruction error of this batch is computed for each autoencoder. If any autoencoder’s reconstruction error on the current batch is lower than its stored average plus $k$ (varied throughout experiments) standard deviations, the batch is matched to that expert (if multiple matches are found, we assign the batch to the expert with the lowest reconstruction error). If no match is found, a new expert and corresponding autoencoder are created.
\item The selected expert and autoencoder—whether newly created or existing—are then trained for one iteration (i.e., one gradient step) on this batch. For the autoencoder, the mean and standard deviation of reconstruction error are updated using this batch (if the expert already existed).
\item If, at any point, more than 30 experts are created (on average corresponding to 10 experts per class), training is halted and subsequent performance is marked as “failure.” This is because the system becomes computationally unscalable, and from an operational standpoint, the presence of 10 or more experts for a single class indicates an unreasonably over-specialized structure.
\end{itemize}


Notice that this approach follows a similar line to \cite{jacobson2022task, erden2024directed, erden2025continual}. However, the key difference is that in these prior works, the learning flow explicitly assumes that each predictive network and its associated autoencoder (both corresponding to a single task) are trained until convergence and then frozen. Furthermore, it is assumed that all subsequent data during that phase continues to pertain to the same task. This implies a strong task-boundary assumption, effectively simplifying the problem within that context. In contrast, we do not impose such constraints. Our method allows for task changes at any point, without requiring convergence or freezing of experts, thereby simulating a more realistic and unconstrained flow. Still, a weaker form of task-boundary assumption remains (which our design, The Modeller does not have), as we assume that all samples in a batch correspond to the same task. To examine the relaxation of this assumption as well, we conduct experiments not only with the default setup (batch size of 20 samples and one training iteration per class per cycle) but also with an alternative setup where the batch size is reduced to 1, and 20 training iterations are performed per class per cycle.

\begin{figure*}
     \centering
     \begin{subfigure}{0.32\textwidth}
         \centering
         \includegraphics[width=\textwidth]{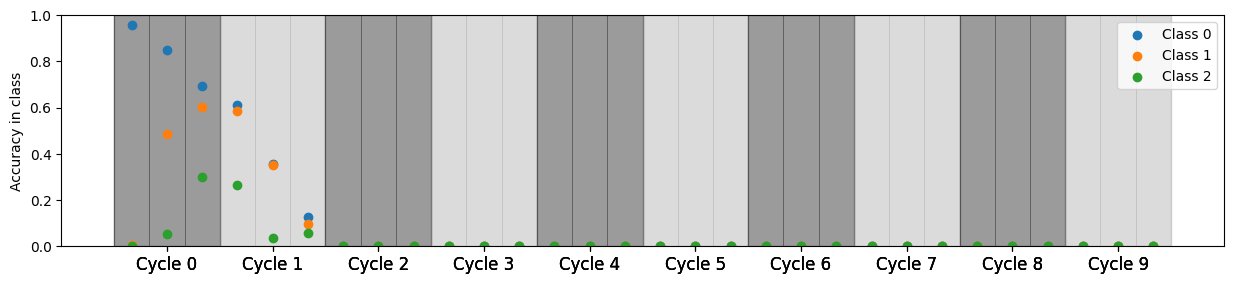}
         \caption{Batch size 1, 20 training iterations per class in cycle.}
         \label{fig:cl_aec_bs1}
     \end{subfigure}
     \begin{subfigure}{0.32\textwidth}
         \centering
         \includegraphics[width=\textwidth]{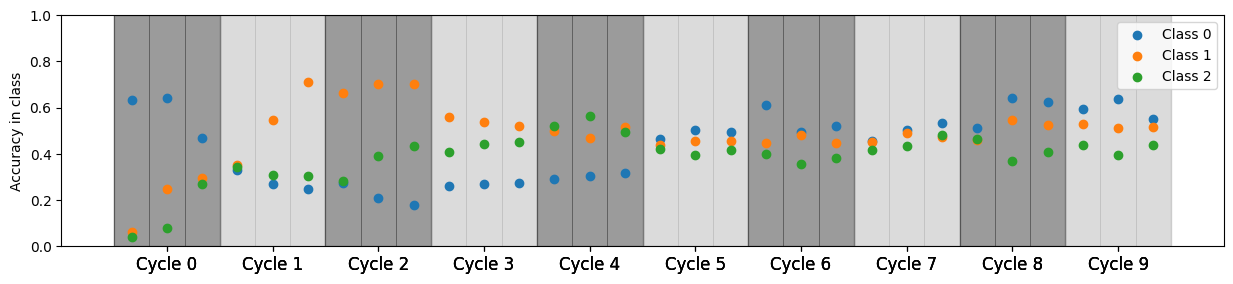}
         \caption{Batch size 20, $k=0$, one training iteration per class in cycle.}
         \label{fig:cl_aec_bs20_k0}
     \end{subfigure}
     \begin{subfigure}{0.32\textwidth}
         \centering
         \includegraphics[width=\textwidth]{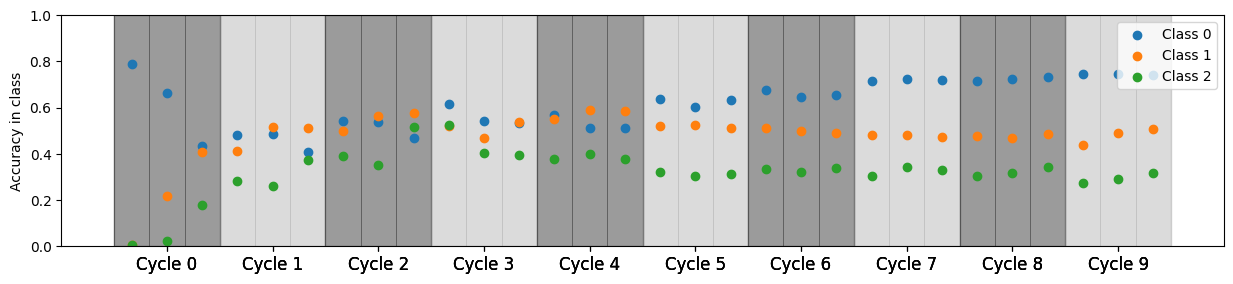}
         \caption{Batch size 20, $k=0.5$, one training iteration per class in cycle.}
         \label{fig:cl_aec_bs20_k0p5}
     \end{subfigure}
     \begin{subfigure}{0.32\textwidth}
         \centering
         \includegraphics[width=\textwidth]{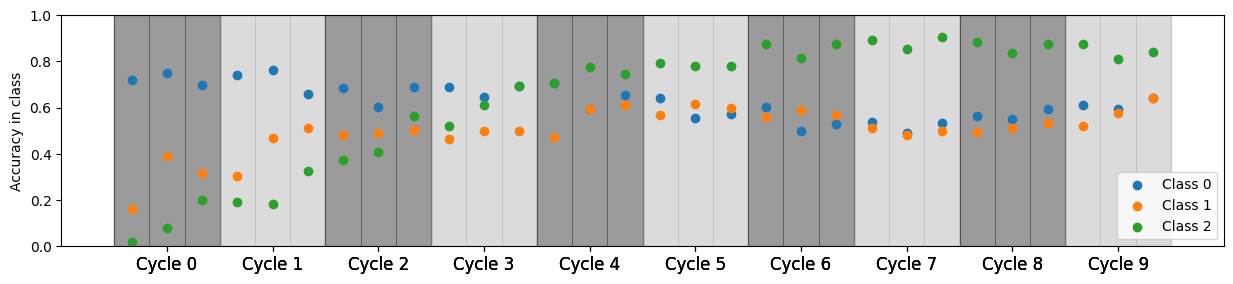}
         \caption{Batch size 20, $k=1$, one training iteration per class in cycle.}
         \label{fig:cl_aec_bs20_k1}
     \end{subfigure}
     \begin{subfigure}{0.32\textwidth}
         \centering
         \includegraphics[width=\textwidth]{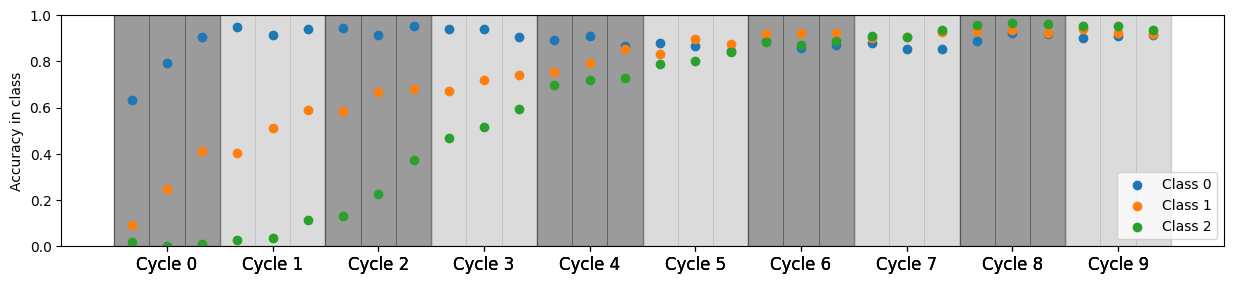}
         \caption{Batch size 20, $k=2$, one training iteration per class in cycle.}
         \label{fig:cl_aec_bs20_k2}
     \end{subfigure}
    \caption{Learning performance of the used multiple-experts system on $N_C$-class incremental learning over 10 cycles, for different batch sizes and $k$ values (see the text for details). Accuracies reflect correct classification ratios for each class. Shaded areas denote cycles, and vertical lines separate iterations within cycles. Results are averaged over 10 (a-c) and 5 (d-e) runs. Note that class indices $i$ are randomly chosen at the start of each run and do \textit{not} necessarily correspond to digit $i$. Note that the trials where training is stopped due to the exceeding of 30 experts, performance is taken as 0.}
    \label{fig:results_cl_aec}
\end{figure*}

Figure~\ref{fig:results_cl_aec} shows the performance of this approach. When using a batch size of 1, the system rapidly exceeds the 30-expert limit (on average after 4.375 global iterations), clearly indicating that an excessive number of experts are created for individual samples within the same task. (As noted in the figure, performance is recorded as 0 after training is halted in these cases.)

With a batch size of 20, we observe that the system still performs poorly—despite the fact that neural networks can, in principle, learn perfect classification accuracy. This suggests that the primary limitation lies in the task-assignment mechanism. Performance remains well below that of MNR (despite its representational limitations discussed in the main text), and while catastrophic forgetting is not directly observed (partly due to already-low accuracy) retention and adaptability are still lacking. Only the case with $k=2$ achieves good final performance, but this occurs primarily because all samples are assigned to the same experts: A larger $k$ increases leniency in matching, which prevents the creation of specialized experts. As a result, learning of new classes is significantly delayed, particularly during early iterations. In this scenario, the system effectively reduces to one large expert trained via stochastic gradient descent with slower timescales. We emphasize that even with 20 samples per batch, the method retains an implicit task-boundary assumption—namely, that all samples in a batch originate from the same task. This is a structural assumption that MNR explicitly avoids.

In summary, the performance of the multi-expert method breaks down as soon as the task-boundary assumption is even partially relaxed—that is, when continuous inflow of task-specific data until convergence is no longer guaranteed. In most cases, both retention and overall performance remain inferior to MNR, which makes no such assumption at all.

\subsection{Experimental performance of neural network-based planning}
\label{sec:nn_plan_exps}

In this section, we present the experimental performance of a neural network-based planner, primarily to demonstrate why using neural networks to learn an environment model for planning is infeasible—serving here for completeness of the narrative. We use a deliberately simple planning mechanism that does \textit{not} incorporate any heuristics, value estimations, or other goal-conditioned signals. This constraint—planning purely on top of a pre-learned environment model, without any prior knowledge or goal-directed learning—is consistent with the assumptions used in the main text, and corresponds to the regime under which The Modeller and its planner extension are tested.

To evaluate this, the system in this section operates as follows:

\begin{itemize}

    \item First, 10,000 samples are collected from the environment through unfolding agent–environment interaction. During this phase, the agent selects actions randomly.
    
    \item These samples are used to train a fully connected neural network that serves as a \textit{transition model}. The inputs to this model are the current-state observations and the chosen action; the target outputs are the corresponding next-state observations.
    
    \item Once the model is trained, a testing phase is conducted over 5 episodes. In each episode, the agent is placed in the environment and given a predefined goal (as in the main text, activating state 1G). At every decision step, the agent takes the current observations and combines them with 2 randomly selected actions. For each of these, it generates 2 predicted next-state samples, resulting in 4 total state-action alternatives. It then recursively simulates the next step by choosing two further random actions for each predicted next state, again generating 2 samples per action. This recursive prediction continues for 3 steps total, effectively implementing a random search of depth 3. If the goal state is found within this simulated search tree, the agent executes the first action on the corresponding path; otherwise, it selects an action at random.

\end{itemize}




The number of next-state and action samples (2 each), along with the plan search depth (3 steps), was selected to remain within the bounds of computational feasibility. Beyond this setup, computational time increases to impractical levels—even in this relatively simple environment. This method of planning is inherently inefficient; however, in the absence of any goal-specific pre-training or the use of predefined heuristics (which, as noted in the main text, aligns with our experimental assumptions), such a random search constitutes the only viable alternative when relying solely on a neural network-based environment model.

As in the main text experiments, we use the SMR environment. We report results only for the setting in which distinct environment subtypes are enabled (with the process above applied independently to each subtype), as this setup alone sufficiently demonstrates the clear limitations of this approach.

Table~\ref{tab:nn_plan_exps} presents the results of this planning method across the three environment subtypes. Compared to Table~\ref{tab:continual}, which employs planning on a clean, structured model learned by The Modeller, all environment subtypes without exception require significantly longer to reach the goal. Most notably, the Vanilla variants exhibit differences of orders of magnitude—for example, RS requires 10 steps (after having learned, column “RS-NL” in Table~\ref{tab:continual})  with The Modeller versus 266 steps here, and SGS requires 4 steps versus 75 steps—clearly indicating the failure of this neural network–based planning approach \textit{without} any goal-dependent or conditioned pre-training or user-defined heuristics.

\begin{table}[]
    \centering
    \caption{Planning with a neural network-based environment model. Mean episode durations across environment changes for vanilla and random variants of the environment subtypes, with purely random actions provided as a baseline. All results are averages across 5 trials, with standard deviations inside parantheses.}
    \begin{tabular}{c|c|c|c}
         & RS & SGS & NEG \\
        \hline
        Vanilla & 266.28 (96.54) & 75.92 (10.51) & 54.40 (10.99) \\
        Random Env. & 232.96 (113.44) & 56.44 (10.40) & 47.96 (4.22) \\        
    \end{tabular}
    \label{tab:nn_plan_exps}
\end{table}


\vfill

\end{document}

%% file: main.bbl

%% file: main.bbl
\begin{thebibliography}{10}
\providecommand{\url}[1]{#1}
\csname url@samestyle\endcsname
\providecommand{\newblock}{\relax}
\providecommand{\bibinfo}[2]{#2}
\providecommand{\BIBentrySTDinterwordspacing}{\spaceskip=0pt\relax}
\providecommand{\BIBentryALTinterwordstretchfactor}{4}
\providecommand{\BIBentryALTinterwordspacing}{\spaceskip=\fontdimen2\font plus
\BIBentryALTinterwordstretchfactor\fontdimen3\font minus \fontdimen4\font\relax}
\providecommand{\BIBforeignlanguage}[2]{{%
\expandafter\ifx\csname l@#1\endcsname\relax
\typeout{** WARNING: IEEEtran.bst: No hyphenation pattern has been}%
\typeout{** loaded for the language `#1'. Using the pattern for}%
\typeout{** the default language instead.}%
\else
\language=\csname l@#1\endcsname
\fi
#2}}
\providecommand{\BIBdecl}{\relax}
\BIBdecl

\bibitem{clune2019ai}
J.~Clune, ``Ai-gas: Ai-generating algorithms, an alternate paradigm for producing general artificial intelligence,'' \emph{arXiv preprint arXiv:1905.10985}, 2019.

\bibitem{zador2019critique}
A.~M. Zador, ``A critique of pure learning and what artificial neural networks can learn from animal brains,'' \emph{Nature communications}, vol.~10, no.~1, p. 3770, 2019.

\bibitem{marcus2018deep}
G.~Marcus, ``Deep learning: A critical appraisal,'' \emph{arXiv preprint arXiv:1801.00631}, 2018.

\bibitem{lecun2022path}
Y.~LeCun, ``A path towards autonomous machine intelligence version 0.9. 2, 2022-06-27,'' \emph{Open Review}, vol.~62, no.~1, 2022.

\bibitem{erden2025foundations}
Z.~D. Erden, ``Foundations of a new learning paradigm in {AI} grounded in the principles of evolutionary developmental biology,'' Ph.D. dissertation, EPFL, 2025.

\bibitem{vandeven2024continuallearningcatastrophicforgetting}
\BIBentryALTinterwordspacing
G.~M. van~de Ven, N.~Soures, and D.~Kudithipudi, ``Continual learning and catastrophic forgetting,'' 2024. [Online]. Available: \url{https://arxiv.org/abs/2403.05175}
\BIBentrySTDinterwordspacing

\bibitem{hadsell2020embracing}
R.~Hadsell, D.~Rao, A.~A. Rusu, and R.~Pascanu, ``Embracing change: Continual learning in deep neural networks,'' \emph{Trends in cognitive sciences}, vol.~24, no.~12, pp. 1028--1040, 2020.

\bibitem{rolnick2019experience}
D.~Rolnick, A.~Ahuja, J.~Schwarz, T.~Lillicrap, and G.~Wayne, ``Experience replay for continual learning,'' \emph{Advances in neural information processing systems}, vol.~32, 2019.

\bibitem{buzzega2021rethinking}
P.~Buzzega, M.~Boschini, A.~Porrello, and S.~Calderara, ``Rethinking experience replay: a bag of tricks for continual learning,'' in \emph{2020 25th International Conference on Pattern Recognition (ICPR)}.\hskip 1em plus 0.5em minus 0.4em\relax IEEE, 2021, pp. 2180--2187.

\bibitem{buzzega2020dark}
P.~Buzzega, M.~Boschini, A.~Porrello, D.~Abati, and S.~Calderara, ``Dark experience for general continual learning: a strong, simple baseline,'' \emph{Advances in neural information processing systems}, vol.~33, pp. 15\,920--15\,930, 2020.

\bibitem{masse2018alleviating}
N.~Y. Masse, G.~D. Grant, and D.~J. Freedman, ``Alleviating catastrophic forgetting using context-dependent gating and synaptic stabilization,'' \emph{Proceedings of the National Academy of Sciences}, vol. 115, no.~44, pp. E10\,467--E10\,475, 2018.

\bibitem{jacobson2022task}
M.~J. Jacobson, C.~Q. Wright, N.~Jiang, G.~Rodriguez-Rivera, and Y.~Xue, ``Task detection in continual learning via familiarity autoencoders,'' in \emph{2022 IEEE International Conference on Systems, Man, and Cybernetics (SMC)}.\hskip 1em plus 0.5em minus 0.4em\relax IEEE, 2022, pp. 1--8.

\bibitem{kirkpatrick2017overcoming}
J.~Kirkpatrick, R.~Pascanu, N.~Rabinowitz, J.~Veness, G.~Desjardins, A.~A. Rusu, K.~Milan, J.~Quan, T.~Ramalho, A.~Grabska-Barwinska \emph{et~al.}, ``Overcoming catastrophic forgetting in neural networks,'' \emph{Proceedings of the national academy of sciences}, vol. 114, no.~13, pp. 3521--3526, 2017.

\bibitem{iyer2022avoiding}
A.~Iyer, K.~Grewal, A.~Velu, L.~O. Souza, J.~Forest, and S.~Ahmad, ``Avoiding catastrophe: Active dendrites enable multi-task learning in dynamic environments,'' \emph{Frontiers in neurorobotics}, vol.~16, p. 846219, 2022.

\bibitem{rusu2016progressive}
A.~A. Rusu, N.~C. Rabinowitz, G.~Desjardins, H.~Soyer, J.~Kirkpatrick, K.~Kavukcuoglu, R.~Pascanu, and R.~Hadsell, ``Progressive neural networks,'' \emph{arXiv preprint arXiv:1606.04671}, 2016.

\bibitem{erden2024directed}
Z.~D. Erden and B.~Faltings, ``Directed structural adaptation to overcome statistical conflicts and enable continual learning,'' in \emph{AAAI 2024}, ser. Deployable AI Workshop, 2024.

\bibitem{lee2020neural}
S.~Lee, J.~Ha, D.~Zhang, and G.~Kim, ``A neural dirichlet process mixture model for task-free continual learning,'' \emph{arXiv preprint arXiv:2001.00689}, 2020.

\bibitem{hafez2023continual}
M.~B. Hafez and S.~Wermter, ``Continual robot learning using self-supervised task inference,'' \emph{IEEE Transactions on Cognitive and Developmental Systems}, vol.~16, no.~3, pp. 947--960, 2023.

\bibitem{damleactive}
S.~Damle, S.~Lokam, and N.~Goyal, ``How do active dendrite networks mitigate catastrophic forgetting?'' in \emph{The First Workshop on NeuroAI@ NeurIPS2024}.

\bibitem{qu2021recent}
H.~Qu, H.~Rahmani, L.~Xu, B.~Williams, and J.~Liu, ``Recent advances of continual learning in computer vision: An overview,'' \emph{arXiv preprint arXiv:2109.11369}, 2021.

\bibitem{galashov2023continually}
A.~Galashov, J.~Mitrovic, D.~Tirumala, Y.~W. Teh, T.~Nguyen, A.~Chaudhry, and R.~Pascanu, ``Continually learning representations at scale,'' in \emph{Conference on Lifelong Learning Agents}.\hskip 1em plus 0.5em minus 0.4em\relax PMLR, 2023, pp. 534--547.

\bibitem{wang2022continual}
Z.~Wang, L.~Liu, Y.~Duan, Y.~Kong, and D.~Tao, ``Continual learning with lifelong vision transformer,'' in \emph{Proceedings of the IEEE/CVF Conference on Computer Vision and Pattern Recognition}, 2022, pp. 171--181.

\bibitem{du2018power}
S.~Du and J.~Lee, ``On the power of over-parametrization in neural networks with quadratic activation,'' in \emph{International conference on machine learning}.\hskip 1em plus 0.5em minus 0.4em\relax PMLR, 2018, pp. 1329--1338.

\bibitem{su2024focuslearn}
Q.~Su, C.~Kloukinas, and A.~d. Garcez, ``Focuslearn: Fully-interpretable, high-performance modular neural networks for time series,'' in \emph{2024 International Joint Conference on Neural Networks (IJCNN)}.\hskip 1em plus 0.5em minus 0.4em\relax IEEE, 2024, pp. 1--8.

\bibitem{goyal2020object}
A.~Goyal, A.~Lamb, P.~Gampa, P.~Beaudoin, S.~Levine, C.~Blundell, Y.~Bengio, and M.~Mozer, ``Object files and schemata: Factorizing declarative and procedural knowledge in dynamical systems,'' \emph{arXiv preprint arXiv:2006.16225}, 2020.

\bibitem{pateria2021hierarchical}
S.~Pateria, B.~Subagdja, A.-h. Tan, and C.~Quek, ``Hierarchical reinforcement learning: A comprehensive survey,'' \emph{ACM Computing Surveys (CSUR)}, vol.~54, no.~5, pp. 1--35, 2021.

\bibitem{andreas2017modular}
J.~Andreas, D.~Klein, and S.~Levine, ``Modular multitask reinforcement learning with policy sketches,'' in \emph{International conference on machine learning}.\hskip 1em plus 0.5em minus 0.4em\relax PMLR, 2017, pp. 166--175.

\bibitem{devin2017learning}
C.~Devin, A.~Gupta, T.~Darrell, P.~Abbeel, and S.~Levine, ``Learning modular neural network policies for multi-task and multi-robot transfer,'' in \emph{2017 IEEE international conference on robotics and automation (ICRA)}.\hskip 1em plus 0.5em minus 0.4em\relax IEEE, 2017, pp. 2169--2176.

\bibitem{sahni2017learning}
H.~Sahni, S.~Kumar, F.~Tejani, and C.~Isbell, ``Learning to compose skills,'' \emph{arXiv preprint arXiv:1711.11289}, 2017.

\bibitem{goyal2019reinforcement}
A.~Goyal, S.~Sodhani, J.~Binas, X.~B. Peng, S.~Levine, and Y.~Bengio, ``Reinforcement learning with competitive ensembles of information-constrained primitives,'' \emph{arXiv preprint arXiv:1906.10667}, 2019.

\bibitem{yang2020multi}
R.~Yang, H.~Xu, Y.~Wu, and X.~Wang, ``Multi-task reinforcement learning with soft modularization,'' \emph{Advances in Neural Information Processing Systems}, vol.~33, pp. 4767--4777, 2020.

\bibitem{xu2019explainable}
F.~Xu, H.~Uszkoreit, Y.~Du, W.~Fan, D.~Zhao, and J.~Zhu, ``Explainable ai: A brief survey on history, research areas, approaches and challenges,'' in \emph{Natural Language Processing and Chinese Computing: 8th CCF International Conference, NLPCC 2019, Dunhuang, China, October 9--14, 2019, Proceedings, Part II 8}.\hskip 1em plus 0.5em minus 0.4em\relax Springer, 2019, pp. 563--574.

\bibitem{kashefi2023explainability}
R.~Kashefi, L.~Barekatain, M.~Sabokrou, and F.~Aghaeipoor, ``Explainability of vision transformers: A comprehensive review and new perspectives,'' \emph{arXiv preprint arXiv:2311.06786}, 2023.

\bibitem{ghallab2016automated}
M.~Ghallab, D.~Nau, and P.~Traverso, \emph{Automated planning and acting}.\hskip 1em plus 0.5em minus 0.4em\relax Cambridge University Press, 2016.

\bibitem{ccalicsir2019model}
S.~{\c{C}}al{\i}{\c{s}}{\i}r and M.~K. Pehlivano{\u{g}}lu, ``Model-free reinforcement learning algorithms: A survey,'' in \emph{2019 27th signal processing and communications applications conference (SIU)}.\hskip 1em plus 0.5em minus 0.4em\relax IEEE, 2019, pp. 1--4.

\bibitem{jimenez2012review}
S.~Jim{\'e}nez, T.~De~La~Rosa, S.~Fern{\'a}ndez, F.~Fern{\'a}ndez, and D.~Borrajo, ``A review of machine learning for automated planning,'' \emph{The Knowledge Engineering Review}, vol.~27, no.~4, pp. 433--467, 2012.

\bibitem{mordoch2023learning}
A.~Mordoch, B.~Juba, and R.~Stern, ``Learning safe numeric action models,'' in \emph{Proceedings of the AAAI Conference on Artificial Intelligence}, vol.~37, no.~10, 2023, pp. 12\,079--12\,086.

\bibitem{verma2021asking}
P.~Verma, S.~R. Marpally, and S.~Srivastava, ``Asking the right questions: Learning interpretable action models through query answering,'' in \emph{Proceedings of the AAAI Conference on Artificial Intelligence}, vol.~35, no.~13, 2021, pp. 12\,024--12\,033.

\bibitem{stern2017efficient}
R.~Stern and B.~Juba, ``Efficient, safe, and probably approximately complete learning of action models,'' \emph{arXiv preprint arXiv:1705.08961}, 2017.

\bibitem{moerland2023model}
T.~M. Moerland, J.~Broekens, A.~Plaat, C.~M. Jonker \emph{et~al.}, ``Model-based reinforcement learning: A survey,'' \emph{Foundations and Trends{\textregistered} in Machine Learning}, vol.~16, no.~1, pp. 1--118, 2023.

\bibitem{moerland2020framework}
T.~M. Moerland, J.~Broekens, and C.~M. Jonker, ``A framework for reinforcement learning and planning,'' \emph{arXiv preprint arXiv:2006.15009}, vol. 127, 2020.

\bibitem{hammersley2013monte}
J.~Hammersley, \emph{Monte carlo methods}.\hskip 1em plus 0.5em minus 0.4em\relax Springer Science \& Business Media, 2013.

\bibitem{mcmahon2022survey}
T.~McMahon, A.~Sivaramakrishnan, E.~Granados, K.~E. Bekris \emph{et~al.}, ``A survey on the integration of machine learning with sampling-based motion planning,'' \emph{Foundations and Trends{\textregistered} in Robotics}, vol.~9, no.~4, pp. 266--327, 2022.

\bibitem{otte2015survey}
M.~W. Otte, ``A survey of machine learning approaches to robotic path-planning,'' \emph{University of Colorado at Boulder, Boulder}, 2015.

\bibitem{dai2019nest}
X.~Dai, H.~Yin, and N.~K. Jha, ``Nest: A neural network synthesis tool based on a grow-and-prune paradigm,'' \emph{IEEE Transactions on Computers}, vol.~68, no.~10, pp. 1487--1497, 2019.

\bibitem{evci2022gradmax}
U.~Evci, B.~van Merrienboer, T.~Unterthiner, M.~Vladymyrov, and F.~Pedregosa, ``Gradmax: Growing neural networks using gradient information,'' \emph{arXiv preprint arXiv:2201.05125}, 2022.

\bibitem{mitchell2023self}
R.~Mitchell, M.~Mundt, and K.~Kersting, ``Self expanding neural networks,'' \emph{arXiv preprint arXiv:2307.04526}, 2023.

\bibitem{maile2022and}
K.~Maile, E.~Rachelson, H.~Luga, and D.~G. Wilson, ``When, where, and how to add new neurons to anns,'' in \emph{International Conference on Automated Machine Learning}.\hskip 1em plus 0.5em minus 0.4em\relax PMLR, 2022, pp. 18--1.

\bibitem{maile2022structural}
K.~Maile, L.~Herv{\'e}, and D.~G. Wilson, ``Structural learning in artificial neural networks: A neural operator perspective,'' 2022.

\bibitem{ding2023structural}
Z.~Ding, H.~Xie, P.~Li, and X.~Xu, ``A structural developmental neural network with information saturation for continual unsupervised learning,'' \emph{CAAI Transactions on Intelligence Technology}, vol.~8, no.~3, pp. 780--795, 2023.

\bibitem{kitson2023survey}
N.~K. Kitson, A.~C. Constantinou, Z.~Guo, Y.~Liu, and K.~Chobtham, ``A survey of bayesian network structure learning,'' \emph{Artificial Intelligence Review}, vol.~56, no.~8, pp. 8721--8814, 2023.

\bibitem{erden2025parallels}
Z.~D. Erden and B.~Faltings, ``On the parallels between evolutionary theory and the state of ai,'' \emph{arXiv preprint arXiv:2505.23774}, 2025.

\bibitem{erden2025evolutionary}
------, ``Evolutionary developmental biology can serve as the conceptual foundation for a new design paradigm in artificial intelligence,'' \emph{arXiv preprint arXiv:2506.12891}, 2025.

\bibitem{marc2005plausibility}
K.~Marc, \emph{The plausibility of life}.\hskip 1em plus 0.5em minus 0.4em\relax Yale University Press, 2005.

\bibitem{west2003developmental}
M.~J. West-Eberhard, \emph{Developmental plasticity and evolution}.\hskip 1em plus 0.5em minus 0.4em\relax Oxford University Press, 2003.

\bibitem{gerhart2007theory}
J.~Gerhart and M.~Kirschner, ``The theory of facilitated variation,'' \emph{Proceedings of the National Academy of Sciences}, vol. 104, no. suppl\_1, pp. 8582--8589, 2007.

\bibitem{holstein2012evolution}
T.~W. Holstein, ``The evolution of the wnt pathway,'' \emph{Cold Spring Harbor Perspectives in Biology}, vol.~4, no.~7, p. a007922, 2012.

\bibitem{lemons2006genomic}
D.~Lemons and W.~McGinnis, ``Genomic evolution of hox gene clusters,'' \emph{Science}, vol. 313, no. 5795, pp. 1918--1922, 2006.

\bibitem{Erden2024Modelleyen}
Z.~D. Erden and B.~Faltings, ``Modelleyen: Continual learning and planning via structured modelling of environment dynamics,'' in \emph{Brain Informatics: 17th International Conference, BI 2024, Bangkok, Thailand, December 13–15, 2024, Proceedings}, S.~Itthipuripat, G.~Ascoli, A.~Li, N.~Pat, and H.~Kuai, Eds.\hskip 1em plus 0.5em minus 0.4em\relax Springer Singapore, 2024.

\bibitem{MinigridMiniworld23}
M.~Chevalier-Boisvert, B.~Dai, M.~Towers, R.~de~Lazcano, L.~Willems, S.~Lahlou, S.~Pal, P.~S. Castro, and J.~Terry, ``Minigrid \& miniworld: Modular \& customizable reinforcement learning environments for goal-oriented tasks,'' \emph{CoRR}, vol. abs/2306.13831, 2023.

\bibitem{lindeberg2012scale}
T.~Lindeberg, ``Scale invariant feature transform,'' 2012.

\bibitem{oquab2023dinov2}
M.~Oquab, T.~Darcet, T.~Moutakanni, H.~Vo, M.~Szafraniec, V.~Khalidov, P.~Fernandez, D.~Haziza, F.~Massa, A.~El-Nouby \emph{et~al.}, ``Dinov2: Learning robust visual features without supervision,'' \emph{arXiv preprint arXiv:2304.07193}, 2023.

\bibitem{gilbert2000developmental}
\BIBentryALTinterwordspacing
S.~F. Gilbert, \emph{Developmental Biology}, 6th~ed.\hskip 1em plus 0.5em minus 0.4em\relax Sunderland, MA: Sinauer Associates, 2000, a New Evolutionary Synthesis. [Online]. Available: \url{https://www.ncbi.nlm.nih.gov/books/NBK10128/}
\BIBentrySTDinterwordspacing

\bibitem{mayer2013evolution}
\BIBentryALTinterwordspacing
G.~C. Mayer and C.~L. Craig, ``Evolution, theory of,'' in \emph{Encyclopedia of Biodiversity (Second Edition)}, second edition~ed., S.~A. Levin, Ed.\hskip 1em plus 0.5em minus 0.4em\relax Waltham: Academic Press, 2013, pp. 392--400. [Online]. Available: \url{https://www.sciencedirect.com/science/article/pii/B9780123847195000484}
\BIBentrySTDinterwordspacing

\bibitem{carroll2005endless}
\BIBentryALTinterwordspacing
S.~Carroll, \emph{Endless Forms Most Beautiful: The New Science of Evo Devo and the Making of the Animal Kingdom}, ser. ISSR Library.\hskip 1em plus 0.5em minus 0.4em\relax W.W. Norton \& Company, 2005. [Online]. Available: \url{https://books.google.ch/books?id=CnnGKjzw3xMC}
\BIBentrySTDinterwordspacing

\bibitem{laland2015extended}
K.~N. Laland, T.~Uller, M.~W. Feldman, K.~Sterelny, G.~B. M{\"u}ller, A.~Moczek, E.~Jablonka, and J.~Odling-Smee, ``The extended evolutionary synthesis: its structure, assumptions and predictions,'' \emph{Proceedings of the royal society B: biological sciences}, vol. 282, no. 1813, p. 20151019, 2015.

\bibitem{gould1977punctuated}
S.~J. Gould and N.~Eldredge, ``Punctuated equilibria: the tempo and mode of evolution reconsidered,'' \emph{Paleobiology}, vol.~3, no.~2, pp. 115--151, 1977.

\bibitem{smith1997major}
J.~M. Smith and E.~Szathmary, \emph{The major transitions in evolution}.\hskip 1em plus 0.5em minus 0.4em\relax OUP Oxford, 1997.

\bibitem{kurzweil2006singularity}
R.~Kurzweil, \emph{The Singularity Is Near: When Humans Transcend Biology}.\hskip 1em plus 0.5em minus 0.4em\relax Penguin (Non-Classics), 2006.

\bibitem{bayley2008design}
D.~J. Bayley, R.~J. Hartfield~Jr, J.~E. Burkhalter, and R.~M. Jenkins, ``Design optimization of a space launch vehicle using a genetic algorithm,'' \emph{Journal of Spacecraft and Rockets}, vol.~45, no.~4, pp. 733--740, 2008.

\bibitem{meli2021study}
C.~Meli, V.~Nezval, Z.~K. Oplatkova, V.~Buttigieg, and A.~S. Staines, ``A study of direct and indirect encoding in phenotype-genotype relationships,'' in \emph{Artificial Intelligence and Soft Computing: 20th International Conference, ICAISC 2021, Virtual Event, June 21--23, 2021, Proceedings, Part II 20}.\hskip 1em plus 0.5em minus 0.4em\relax Springer, 2021, pp. 290--301.

\bibitem{rajewsky1996clonal}
K.~Rajewsky, ``Clonal selection and learning in the antibody system,'' \emph{Nature}, vol. 381, no. 6585, pp. 751--758, 1996.

\bibitem{hiesinger2021self}
P.~R. Hiesinger, ``The self-assembling brain: how neural networks grow smarter,'' 2021.

\bibitem{petanjek2023dendritic}
Z.~Petanjek, I.~Banovac, D.~Sedmak, and A.~Hladnik, ``Dendritic spines: synaptogenesis and synaptic pruning for the developmental organization of brain circuits,'' in \emph{Dendritic spines: structure, function, and plasticity}.\hskip 1em plus 0.5em minus 0.4em\relax Springer, 2023, pp. 143--221.

\bibitem{tickle2017sonic}
C.~Tickle and M.~Towers, ``Sonic hedgehog signaling in limb development,'' \emph{Frontiers in cell and developmental biology}, vol.~5, p.~14, 2017.

\bibitem{wang2016divergence}
P.~Wang, D.~Zhao, S.~Rockowitz, and D.~Zheng, ``Divergence and rewiring of regulatory networks for neural development between human and other species,'' \emph{Neurogenesis}, vol.~3, no.~1, pp. 5730--5743, 2016.

\bibitem{suresh2023comparative}
H.~Suresh, M.~Crow, N.~Jorstad, R.~Hodge, E.~Lein, A.~Dobin, T.~Bakken, and J.~Gillis, ``Comparative single-cell transcriptomic analysis of primate brains highlights human-specific regulatory evolution,'' \emph{Nature Ecology \& Evolution}, vol.~7, no.~11, pp. 1930--1943, 2023.

\bibitem{lamb2007evolution}
T.~D. Lamb, S.~P. Collin, and E.~N. Pugh, ``Evolution of the vertebrate eye: opsins, photoreceptors, retina and eye cup,'' \emph{Nature Reviews Neuroscience}, vol.~8, no.~12, pp. 960--976, 2007.

\bibitem{lamb2008origin}
T.~D. Lamb, E.~N. Pugh, and S.~P. Collin, ``The origin of the vertebrate eye,'' \emph{Evolution: Education and Outreach}, vol.~1, pp. 415--426, 2008.

\bibitem{halder1995induction}
G.~Halder, P.~Callaerts, and W.~J. Gehring, ``Induction of ectopic eyes by targeted expression of the eyeless gene in drosophila,'' \emph{Science}, vol. 267, no. 5205, pp. 1788--1792, 1995.

\bibitem{clune2013evolutionary}
J.~Clune, J.-B. Mouret, and H.~Lipson, ``The evolutionary origins of modularity,'' \emph{Proceedings of the Royal Society b: Biological sciences}, vol. 280, no. 1755, p. 20122863, 2013.

\bibitem{mengistu2016evolutionary}
H.~Mengistu, J.~Huizinga, J.-B. Mouret, and J.~Clune, ``The evolutionary origins of hierarchy,'' \emph{PLoS computational biology}, vol.~12, no.~6, p. e1004829, 2016.

\bibitem{szilagyi2020phenotypes}
A.~Szil{\'a}gyi, P.~Szab{\'o}, M.~Santos, and E.~Szathm{\'a}ry, ``Phenotypes to remember: Evolutionary developmental memory capacity and robustness,'' \emph{PLoS computational biology}, vol.~16, no.~11, p. e1008425, 2020.

\bibitem{carroll1995homeotic}
S.~B. Carroll, ``Homeotic genes and the evolution of arthropods and chordates,'' \emph{Nature}, vol. 376, no. 6540, pp. 479--485, 1995.

\bibitem{wan2024towards}
Z.~Wan, C.-K. Liu, H.~Yang, C.~Li, H.~You, Y.~Fu, C.~Wan, T.~Krishna, Y.~Lin, and A.~Raychowdhury, ``Towards cognitive ai systems: a survey and prospective on neuro-symbolic ai,'' \emph{arXiv preprint arXiv:2401.01040}, 2024.

\bibitem{colelough2025neuro}
B.~C. Colelough and W.~Regli, ``Neuro-symbolic ai in 2024: A systematic review,'' \emph{arXiv preprint arXiv:2501.05435}, 2025.

\bibitem{bal2024rethinking}
M.~Bal and A.~Sengupta, ``Rethinking spiking neural networks as state space models,'' \emph{arXiv e-prints}, pp. arXiv--2406, 2024.

\bibitem{opencv_shape_analysis}
\BIBentryALTinterwordspacing
{OpenCV}, ``Shape analysis - opencv documentation,'' 2025, accessed: 2025-01-29. [Online]. Available: \url{https://docs.opencv.org/4.x/d3/dc0/group__imgproc__shape.html#ga0012a5fdaea70b8a9970165d98722b4c}
\BIBentrySTDinterwordspacing

\bibitem{erden2025continual}
Z.~D. Erden, D.~Gasmi, and B.~Faltings, ``Continual {R}einforcement {L}earning via {A}utoencoder{-}driven {T}ask and {N}ew {E}nvironment {R}ecognition,'' in \emph{AAMAS 2025}, ser. Adaptive and Learning Agents \& Autonomous Robots and Multirobot Systems Workshops, 2025.

\end{thebibliography}
